\documentclass{llncs}

\usepackage{amsmath}
\usepackage{amssymb}

\usepackage{latexsym}
\usepackage{picinpar}

\begin{document}

\title{Non Deterministic Logic Programs}

\author{Emad Saad}
\institute{emsaad@gmail.com}

\maketitle
\begin{abstract}
Non deterministic applications arise in many domains, including, stochastic optimization, multi-objectives optimization, stochastic planning, contingent stochastic planning, reinforcement learning, reinforcement learning in partially observable Markov decision processes, and conditional planning. We present a logic programming framework called {\em non deterministic logic programs}, along with a declarative semantics and fixpoint semantics, to allow representing and reasoning about inherently non deterministic real-world applications. The language of non deterministic logic programs framework is extended with non-monotonic negation, and two alternative semantics are defined: the stable non deterministic model semantics and the well-founded non deterministic model semantics as well as their relationship is studied. These semantics subsume the deterministic stable model semantics and the deterministic well-founded semantics of deterministic normal logic programs, and they reduce to the semantics of deterministic definite logic programs without negation. We show the application of the non deterministic logic programs framework to a conditional planning problem.
\end{abstract}

\section{Introduction}

Logic programming is a declarative programming paradigm that is based on logic where a substantial subset of first-order logic is used as the basis for the programming language. The basic language of logic programming is the language of definite logic programs. A definite logic program is a set of Horn clauses whose semantics is given declaratively by model theory and fixpoint theory, where a unique model, which is the least model generated from the definite logic program, is adopted to be the meaning of the definite logic program \cite{Lloyd}. In addition, several extensions to the language of definite logic programs were developed to strength the knowledge representation and reasoning capabilities of the basic language to be more suitable for real-world applications. These extensions include extending definite logic programs with non-monotonic negation to be capable of performing default reasoning and deriving negative conclusions in the absence of positive conclusions. Therefore, definite logic programs were extended to normal logic programs which are definite logic programs with non-monotonic negation. The meaning of normal logic programs is given by the stable model semantics \cite{Gelfond_A}, which is the most well understood semantics for normal logic programs. In stable model semantics of normal logic programs, a normal logic program can have zero, one, or multiple stable models as a meaning of the normal logic program. In addition, a well-founded model semantics for normal logic programs was developed to provide exactly one model as a meaning for normal logic programs \cite{Gelder}. Furthermore, the relationship between the well-founded semantics and the stable model semantics for normal logic programs were carefully studied in \cite{Gelder}.

Definite logic programs were extended to extended logic programs to allow both classical negation and non-monotonic negation \cite{Gelfond_B}. This extension is necessary to allow knowledge representation and reasoning in the presence of incomplete knowledge. In normal logic programs with stable model semantics an assertion is either true or false. However, in extended logic programs an assertion is either true or false or unknown to cope with incomplete knowledge. The meaning of extended logic programs is given by the answer set semantics \cite{Gelfond_B}, where an atom with respect to a given answer set of an extended logic program is either true, false, or undecidable with respect to that answer set.

Another extension to definite logic programs is extending definite logic programs to allow disjunctions in the head of logic rules and classical negation, and non-monotonic negation in the body of the logic rules presenting several forms of expressive logic programs \cite{Gelfond_B}. These forms of logic programs are; disjunctive logic programs which are definite logic program but with only disjunctions in the head of logic rules; normal disjunctive logic programs which are disjunctive logic programs but with non-monotonic negation in the body of the logic rules; and finally extended disjunctive logic programs which are disjunctive logic programs with classical negation and non-monotonic negation in the body of logic rules. The meaning of disjunctive logic programs and normal disjunctive logic programs are given by the stable model semantics and the meaning of extended disjunctive logic programs are given by the answer set semantics.

Stable model semantics for various expressive forms of logic programs including normal, disjunctive, and normal disjunctive logic programs have been shown effective and efficiently applicable to many real-world problems including but not limited to planning, diagnoses, and model checking, where the stable models generated from the stable model semantics of the logic program encoding of the problem representing one-to-one correspondence to the possible solutions of the problem. For example, in normal logic programs with stable model semantics solution to classical planning, each stable model generated from the normal logic program with stable model semantics representation of a classical planning problem corresponds to a valid plan.

However, in many interesting real-world problems, normal, disjunctive, and normal disjunctive logic programs with stable models semantics are not expressive enough to represent these problems and their stable model semantics do not establish one-to-one correspondence to the solutions of these problems. This is because the solution of these real-world problems are {\em trees} of solutions, i.e., each solution to a problem is a tree called a solution tree and stable models semantics works only for problems whose solutions correspond to paths in a tree but not the whole tree, where the whole tree in stable model semantics corresponds to {\em all} the possible solutions. But some real-world problems require a tree per a solution.

Consequently, these kind of problems require expressive forms of logic programs whose semantics is capable of producing solution trees with multiple paths, unlike normal, disjunctive, and normal disjunctive logic programs with stable model semantics that generates only paths in a tree as the possible solutions. These real-world problems that require trees as solutions arise in many domains. The most prominent of these domains are {\em stochastic optimization, multi-objectives optimization, stochastic planning, contingent stochastic planning, reinforcement learning, reinforcement learning in partially observable Markov decision processes, and conditional planning}. An important observation over these applications is that all these applications are {\em  non deterministic}, which intuitively require different kind of logic program representation rather than normal, disjunctive, and normal disjunctive logic programs with stable model semantics that are deterministic and work efficiently for deterministic applications. Consider for example the following conditional planning problem which is clearly non deterministic.

\begin{example} Consider an indoor security robot that checks on the windows lockage. If a window is opened then close the window by the robot gets the window not opened (closed). But, the robot needs to check whether the window is opened or not opened before starting to close the window. In addition, the robot needs to have the window locked as well. Therefore, the robot have to inspect if the window lock is locked or not locked. Then, if the window lock is not locked then flip the window lock by the robot gets the window lock locked. However, if a window lock is locked then flip the window lock by the robot gets the window lock not locked. Initially, some windows are opened and not locked and the target of the security robot is to get these windows not opened and locked. This robot planning problem can be represented as an action theory of the form

\begin{eqnarray}
{\bf initially} \{ opened, \neg locked \} \label{eq:action:initial}
\\
{\bf executable} \; close  \; {\bf if} \; \emptyset \label{eq:action:exec-close}
\\
{\bf executable} \; flip\_lock \; {\bf if} \; \emptyset \label{eq:action:exec-flip}
\\
{\bf executable} \; check \; {\bf if} \; \emptyset \label{eq:action:exec-check}
\\
{\bf executable} \; inspect \; {\bf if} \; \emptyset \label{eq:action:exec-inspect}
\\
close \quad {\bf causes}  \quad \neg opened   \quad  {\bf if}  \quad opened \label{eq:action:non-sensing-close}
\\
\begin{array}{l}
flip\_lock \; {\bf causes}
\left \{
\begin{array}{lcl}
locked   \quad  {\bf if}  \quad \neg locked ,  \\
\neg locked \quad  {\bf if}  \quad locked
\end{array}
\right\}
\end{array}
\label{eq:action:non-sensing-flip}
\\
check \; {\bf determines}
\left \{
\begin{array}{lcl}
opened,  \\
\neg opened
\end{array}
\right\}
\label{eq:action:sensing-check}
\\
inspect \; {\bf determines}
\left \{
\begin{array}{lcl}
locked,  \\
\neg locked
\end{array}
\right\}
\label{eq:action:sensing-inspect}
\end{eqnarray}
The initial situation in this security robot planning problem is presented by the proposition (\ref{eq:action:initial}). Proposition (\ref{eq:action:initial}) states that the possible initial state $\{ opened, \neg locked \}$ holds, which means that initially a window is $opened$ and $not \; locked$ where $\neg$ is the classical negation. Executability conditions of  the various actions in this planning problem is represented by propositions (\ref{eq:action:exec-close}), (\ref{eq:action:exec-flip}), (\ref{eq:action:exec-check}), and (\ref{eq:action:exec-inspect}), which state that actions $close$, $flip\_lock$, $check$, and $inspect$ are executable in any state of the world without conditions, where $\emptyset$ means no conditions are needed for the executability of the actions.

Propositions (\ref{eq:action:non-sensing-close}) and (\ref{eq:action:non-sensing-flip}) represent the conditional effects of the non-sensing actions $close$ and $flip\_lock$. Proposition (\ref{eq:action:non-sensing-close}) says that the action $close$ causes a window to be $not \; opened$ to hold  in a successor state to a state in which the action $close$ is executed and the property $opened $ holds. Proposition (\ref{eq:action:non-sensing-flip}) says that the action $flip\_lock$ causes the window lock to be $locked$ to hold in a successor state to a state in which the action $flip\_lock$ is executed and the property $not\; locked$ holds. Or the action $flip\_lock$ causes the window lock to be $not \; locked$ to hold in a successor state to a state in which the action $flip\_lock$ is executed and the property $locked$ holds. The properties $locked$ and $not \; locked$ must be mutually exclusive and exhaustive.

Propositions (\ref{eq:action:sensing-check}) and (\ref{eq:action:sensing-inspect}) describe the sensing actions $check$ and $inspect$ with their conditional outcomes. Proposition (\ref{eq:action:sensing-check}) states that executing the sensing action $check$ in a state causes the property $opened$ or the property $not \; opened$ to be {\em known} true in a successor state to a state in which $check$ is executed. The properties $opened$ and $not \; opened$ must be mutually exclusive and exhaustive. Proposition (\ref{eq:action:sensing-inspect}) states that executing the sensing action $inspect$ in a state causes the property $locked$ or the property $not \; locked$ to be {\em known} true in a successor state to a state in which $inspect$ is executed. The properties $locked$ and $not \; locked$ must be mutually exclusive and exhaustive.
\label{ex:secure-robot}
\end{example}

The effects of the actions $flip\_lock$, $check$, and $inspect$ are non deterministically determined if their preconditions occur. This implies that for any representation to these actions into a logic program, the logic rules representing these actions and their effects should represent the non deterministic effects of the actions as well as their preconditions. In addition, the semantics of this logic program, representing these actions and their non deterministic effects and preconditions, should be capable of deriving the whole non deterministic effects of these actions whenever their preconditions hold.

Unlikely, no current logic programming language syntax and semantics including definite logic programs \cite{Lloyd}, normal, disjunctive, and normal disjunctive logic programs with stable model semantics, extended and extended disjunctive logic programs with answer set semantics \cite{Gelfond_A,Gelfond_B}, neither capable of representing nor reasoning in the presence of non deterministic knowledge, like the non deterministic knowledge that arise extensively in many critical applications including stochastic optimization, multi-objectives optimization, stochastic planning, contingent stochastic planning, reinforcement learning, reinforcement learning in partially observable Markov decision processes, and conditional planning.

Therefore, we introduce the notion of {\em non deterministic logic programs} to allow representing and reasoning in the presence of non deterministic knowledge. The building blocks of the language of non deterministic logic programs is the notion of {\em non deterministic atoms} to allow appropriately representing and reasoning bout inherently non deterministic real-world applications. The introduction of the notion of non deterministic atoms requires changes at the syntactical and semantical level to the exiting logic programming languages. The new framework provides more intuitive and accurate representation and reasoning about non deterministic knowledge. We show that problems such as the one described in Example (\ref{ex:secure-robot}) are properly addressed in the non deterministic logic programs framework. Furthermore, we show that the non deterministic logic programs framework subsume the deterministic definite logic programs framework \cite{Lloyd} for representing and reasoning about deterministic knowledge.

However, providing non deterministic logic programs to be more sophisticated for representing and reasoning about non deterministic knowledge is still not enough for strengthening the reasoning capabilities of the new logical language. For this reason, the non deterministic logic programs syntax and semantics need to be extended to cope with the non-monotonic negation. This is because non-monotonic negation is important to be able to perform default reasoning and derive negative conclusions in the absence of positive conclusions. As a consequence, enhancing the non deterministic logic programs framework with this capability makes it more suitable for real-world applications.

Therefore, we extend the non deterministic logic programs framework to cope with non-monotonic negation.
This is achieved by developing the stable non deterministic model semantics and the well-founded non deterministic model semantics for non deterministic logic programs with non-monotonic negation.

In this paper we are concerned with developing a proper syntax and semantics for logic programs to cope with the non deterministic knowledge, where every atom appear in a logic rule in a logic program is non deterministic.

\section{Non Deterministic Logic Programs}

In this section we present the syntax and semantics of the language of non deterministic logic programs. The semantics is based on the set-inclusion order and the notion of non deterministic atoms to appropriately represent and reason about inherently non deterministic real-world applications. The introduction of the notion of non deterministic atoms requires changes at the syntactical and semantical level to the exiting logic programming languages.

We start by defining the syntax of the language of non deterministic logic programs that allows the ability to represent non deterministic knowledge. Then we define a declarative semantics and a fixpoint semantics for non deterministic logic programs. The declarative semantics is based on the notion of satisfactions and non deterministic models in which every rule in a non deterministic logic program is satisfied. The fixpoint semantics is developed by defining the notion of the immediate consequence operator of non deterministic logic programs. In addition, we show that the declarative semantics coincides with the fixpoint semantics. The new framework provides more intuitive and easy way to capture non deterministic knowledge. Furthermore, we show that the syntax and semantics of non deterministic logic programs framework is a natural generalization and subsume the original syntax and semantics of definite logic programs

\subsection{Syntax}

In this section, we provide  the syntax of non deterministic logic programs. Let $\cal L$ be a first-order language with many predicate symbols, function symbols, constants, and infinitely many variables. A term is a constant, a variable, or a function $f(x_1,\ldots, x_n)$ where $f$ is an n-ary function symbol and $x_1, \ldots , x_n $ are terms. An atom, $p(x_1,\ldots, x_n)$, is an n-ary predicate symbol, $p$, and $x_1, \ldots , x_n$ are terms.

\begin{definition}
A {\em non deterministic atom} is a set of atoms of the form $\{A_1, \ldots, A_n\}$,  denoted by $\{A_i\}_{i =1}^{n}$, i.e., a set of predicates.
\end{definition}
Intuitively, a non deterministic atom, $\{A_1, \ldots, A_n\}$, is a new construct in the languages of logic programming in general to allow a set of atoms to {\em non deterministically} occurs. This means that if the non deterministic atom, $\{A_1, \ldots, A_n\}$, is occurred to be true is some interpretation, this implies that in any of the worlds one and only one of $A_i \in \{A_1, \ldots, A_n\}$ must be true in that world. This also means that all of the atoms in $\{A_1, \ldots, A_n\}$ are mutually true. In other words, if $A_i \in \{A_1, \ldots, A_n\}$ is true in one of the worlds, $w$, this excludes all the other $A_j \in \{A_1, \ldots, A_n\}$ such that $i \neq j$ from becoming true in that same world $w$.

An atom, $A$, is a non deterministic atom of the form $\{A\}$. The Herbrand universe $U_{\cal L}$ of ${\cal L}$ is the set of all ground terms which can be formed from constants and functions from ${\cal L}$. The Herbrand base ${\cal B_L}$ of ${\cal L}$ is the set of all ground atoms which can be formed using predicate symbols from ${\cal L}$ and ground terms from $U_{\cal L}$. The {\em non deterministic base}, ${\cal N_L}$, of ${\cal L}$ is the power set of ${\cal B_L}$, i.e., ${\cal N_L} = 2^{\cal B_L}$. Non-monotonic negation or the negation as failure is denoted by $not$.

\begin{definition}
A non deterministic logic rule is an expression of the form
\[
\{ A_i \}_{i = 1}^{n} \leftarrow \{ B_{i_1} \}_{i_1 = 1}^{n_1} , \ldots, \{ B_{i_m} \}_{i_m = 1}^{n_m}
\]
where $\{ A_i \}_{i = 1}^{n}, \{ B_{i_1} \}_{i_1 = 1}^{n_1}, \ldots, \{ B_{i_m} \}_{i_m = 1}^{n_m}$ are non deterministic atoms. $\{ A_i \}_{i = 1}^{n}$ is called the head of the non deterministic logic rule and $\{ B_{i_1} \}_{i_1 = 1}^{n_1} , \ldots, \{ B_{i_m} \}_{i_m = 1}^{n_m}$ is its body. If $m = 0$, the non deterministic logic rule is called a non deterministic fact, i.e., $\{ A_i \}_{i = 1}^{n} \leftarrow$.
\end {definition}

\begin{example} The non deterministic logic rule that represents the non deterministic conditional effects of the non-sensing action $flip\_lock$ described in Example (\ref{ex:secure-robot}) is given as

\begin{eqnarray}
\left \{ \begin{array}{r} holds(locked, T+1), \\  holds(\neg locked, T+1) \end{array} \right\} \leftarrow
occ(flip\_lock,T), exec(flip\_lock, T), \notag \\
\left \{\begin{array}{r} holds(\neg locked, T), \\  holds(locked, T) \end{array}  \right\}.
\end{eqnarray}
The above non deterministic logic rule says that if the action $flip\_lock$ occurs at time moment $T$ and the property $locked$ or the property $\neg locked$ non deterministically hold at the same time moment, $T$, then the property $\neg locked$ or the property $locked$ non deterministically hold at time moment $T+1$.
\end{example}

\begin{definition} A non deterministic logic program, $\Pi$, is a set of non deterministic logic rules.
\end{definition}
A term is ground if it does not contain any variables. A ground atom is an atom that does not contain any variables. A ground non deterministic atom is a non deterministic atom that does not contain any variables. A non deterministic logic rule, $r$, is ground if every non deterministic atom appearing in $r$ is ground. A non deterministic logic program, $\Pi$, is ground if every non deterministic logic rule in $\Pi$ is ground.

\begin{example} Fred is gone for his lunch at a restaurant but he is undeterminate about what to eat for the lunch. His preferences for today's lunch is either soup or salad but with either meat or fish as a main dish. Fred checked the menu and narrowed his choices to beef or buffalo soup, salmon or seafood salad, beef or buffalo meat for the main dish, and  salmon or seafood as a fish main dish. The possible lunch choices that Fred can make can be represented as a non deterministic logic program, $\Pi$, that consists of the following non deterministic logic rules, where $lunch(X, Y)$ predicate means that Fred chooses to eat $soup(X)$ with $meat(X)$ or $salad(Y)$ with $fish(Y)$.
\[
\begin{array}{lcl}
\{\;
lunch(X, Y)
\; \}
& \leftarrow &
\left\{
\begin{array}{r}
soup(X), \\
salad(Y)
\end{array}
\right\}
,
\left\{
\begin{array}{r}
meat(X), \\
fish(Y)
\end{array}
\right\}.
\end{array}
\]

\[
\begin{array}{lcl} 
\left\{
\begin{array}{r}
soup(beef), \\
salad(salmon)
\end{array}
\right\}
& \leftarrow &
\end{array}
\]

\[
\begin{array}{lcl} 
\left\{
\begin{array}{r}
soup(beef), \\
salad(seafood)
\end{array}
\right\}
& \leftarrow &
\end{array}
\]

\[
\begin{array}{lcl}
\left\{
\begin{array}{r}
soup(buffalo), \\
salad(salmon)
\end{array}
\right\}
& \leftarrow &
\end{array}
\]

\[
\begin{array}{lcl}
\left\{
\begin{array}{r}
soup(buffalo), \\
salad(seafood)
\end{array}
\right\}
& \leftarrow &
\end{array}
\]

\[
\begin{array}{lcl}
\left\{
\begin{array}{r}
meat(beef), \\
fish(salmon)
\end{array}
\right\}
& \leftarrow &
\end{array}
\]

\[
\begin{array}{lcl}
\left\{
\begin{array}{r}
meat(beef), \\
fish(seafood)
\end{array}
\right\}
& \leftarrow &
\end{array}
\]

\[
\begin{array}{lcl}
\left\{
\begin{array}{r}
meat(buffalo), \\
fish(salmon)
\end{array}
\right\}
& \leftarrow &
\end{array}
\]

\[
\begin{array}{lcl}
\left\{
\begin{array}{r}
meat(buffalo), \\
fish(seafood)
\end{array}
\right\}
& \leftarrow &
\end{array}
\]
\label{ex:Fred}
\end{example}

\begin{example} Fred is planning for his vacation to Europe, however, Fred is undecided about spending his vacation either in Paris or in London. Therefore, Fred wants to plan his vacation such that he travels on the same route from home to either Paris or London so that he would be able to accommodate to any last minute change to his vacation plan. Fred's vacation planning problem can be represented as a non deterministic logic program, $\Pi$, that consists of the following non deterministic logic rules where the predicate $connection_1(X,Y)$ means that there is a connection route from city $X$ to city $Y$ all the way to London, and the predicate $connection_2(X,Y)$ means that there is a connection route from city $X$ to city $Y$ all the way to Paris. The predicate $reachable(X,Y)$ means that city $Y$ is reachable from city $X$.

\[
\begin{array}{lcl}
\{\;
reachable(X,Y)
\; \}
& \leftarrow &
\left\{
\begin{array}{r}
connection_1(X,Y), \\
connection_2(X,Y)
\end{array}
\right\}
.
\\
\\
\{\;
reachable(X,Y)
\; \}
& \leftarrow &
\left\{
\begin{array}{r}
connection_1(X,Z), \\
connection_2(X,Z)
\end{array}
\right\}
,
\{\;
reachable(Z,Y)
\; \}.
\end{array}
\]
\[
\begin{array}{lcl}
\left\{
\begin{array}{r}
connection_1(home,rome), \\
connection_2(home,rome)
\end{array}
\right\}
\leftarrow.
\qquad \qquad
\left\{
\begin{array}{r}
connection_1(home,rome), \\
connection_2(rome,london)
\end{array}
\right\}
\leftarrow.
\\
\\
\left\{
\begin{array}{r}
connection_1(home,rome), \\
connection_2(rome,berlin)
\end{array}
\right\}
\leftarrow.
\qquad \qquad
\left\{
\begin{array}{r}
connection_1(home,rome), \\
connection_2(london,paris)
\end{array}
\right\}
\leftarrow.
\\
\\
\left\{
\begin{array}{r}
connection_1(home,rome), \\
connection_2(berlin,paris)
\end{array}
\right\}
\leftarrow.
\\
\\
\end{array}
\]
\[
\begin{array}{lcl}
\left\{
\begin{array}{r}
connection_1(rome,paris), \\
connection_2(home,rome)
\end{array}
\right\}
\leftarrow.
\qquad \qquad
\left\{
\begin{array}{r}
connection_1(rome,paris), \\
connection_2(rome,london)
\end{array}
\right\}
\leftarrow.
\\
\\
\left\{
\begin{array}{r}
connection_1(rome,paris), \\
connection_2(rome,berlin)
\end{array}
\right\}
\leftarrow.
\qquad \qquad
\left\{
\begin{array}{r}
connection_1(rome,paris), \\
connection_2(london,paris)
\end{array}
\right\}
\leftarrow.
\\
\\
\left\{
\begin{array}{r}
connection_1(rome,paris), \\
connection_2(berlin,paris)
\end{array}
\right\}
\leftarrow.
\\
\\
\end{array}
\]
\[
\begin{array}{lcl}
\left\{
\begin{array}{r}
connection_1(rome,berlin), \\
connection_2(home,rome)
\end{array}
\right\}
\leftarrow.
\qquad \qquad
\left\{
\begin{array}{r}
connection_1(rome,berlin), \\
connection_2(rome,london)
\end{array}
\right\}
\leftarrow.
\\
\\
\left\{
\begin{array}{r}
connection_1(rome,berlin), \\
connection_2(rome,berlin)
\end{array}
\right\}
\leftarrow.
\qquad \qquad
\left\{
\begin{array}{r}
connection_1(rome,berlin), \\
connection_2(london,paris)
\end{array}
\right\}
\leftarrow.
\\
\\
\left\{
\begin{array}{r}
connection_1(rome,berlin), \\
connection_2(berlin,paris)
\end{array}
\right\}
\leftarrow.
\\
\\
\end{array}
\]
\[
\begin{array}{lcl}
\left\{
\begin{array}{r}
connection_1(paris,london), \\
connection_2(home,rome)
\end{array}
\right\}
\leftarrow.
\qquad \qquad
\left\{
\begin{array}{r}
connection_1(paris,london), \\
connection_2(rome,london)
\end{array}
\right\}
\leftarrow.
\\
\\
\left\{
\begin{array}{r}
connection_1(paris,london), \\
connection_2(rome,berlin)
\end{array}
\right\}
\leftarrow.
\qquad \qquad
\left\{
\begin{array}{r}
connection_1(paris,london), \\
connection_2(london,paris)
\end{array}
\right\}
\leftarrow.
\\
\\
\left\{
\begin{array}{r}
connection_1(paris,london), \\
connection_2(berlin,paris)
\end{array}
\right\}
\leftarrow.
\\
\\
\end{array}
\]
\[
\begin{array}{lcl}
\left\{
\begin{array}{r}
connection_1(berlin,london), \\
connection_2(home,rome)
\end{array}
\right\}
\leftarrow.
\qquad \qquad
\left\{
\begin{array}{r}
connection_1(berlin,london), \\
connection_2(rome,london)
\end{array}
\right\}
\leftarrow.
\\
\\
\left\{
\begin{array}{r}
connection_1(berlin,london), \\
connection_2(rome,berlin)
\end{array}
\right\}
\leftarrow.
\qquad \qquad
\left\{
\begin{array}{r}
connection_1(berlin,london), \\
connection_2(london,paris)
\end{array}
\right\}
\leftarrow.
\\
\\
\left\{
\begin{array}{r}
connection_1(berlin,london), \\
connection_2(berlin,paris)
\end{array}
\right\}
\leftarrow.
\end{array}
\]
\label{ex:Connection}
\end{example}

\subsection{Declarative Semantics}

In this section, we define the declarative semantics, model-theoretic semantics, of non deterministic logic programs. We provide definitions for the notions of non deterministic interpretations, satisfaction, and non deterministic models of non deterministic logic programs.

\begin{definition} Let ${\cal L}$ be a first-order language. A non deterministic interpretation, $I$,
for ${\cal L}$ consists of:

\begin{enumerate}

\item The domain of $I$ is the Herbrand universe $U_{\cal L}$.

\item Each constant in ${\cal L}$ is an assignment of an element in $U_{\cal L}$.

\item Each n-ary function symbol in ${\cal L}$ is an assignment of a mapping $U_{\cal L}^n \rightarrow U_{\cal L}$.

\item Each atom, $\{A \}$, an n-ary predicate symbol in ${\cal L}$, is a mapping $U_{\cal L}^n \rightarrow \{true, false\}$.

\item For each non deterministic atom, $\{A_i \}_{i = 1}^n$, $\forall \; A_i \in \{A_i \}_{i = 1}^n$, an n-ary predicate symbol in ${\cal L}$, is a mapping $U_{\cal L}^n \rightarrow \{true\}$ or $\forall \; A_i \in \{A_i \}_{i = 1}^n$, an n-ary predicate symbol in ${\cal L}$, is a mapping $U_{\cal L}^n \rightarrow \{false\}$.

\end{enumerate}
\end{definition}
For easiness we adopt the following. We consider a non deterministic interpretation, $I$, for the first-order language, ${\cal L}$, as a subset of the non deterministic base, ${\cal N_L}$, where all non deterministic atoms that belong to $I$ are the {\em true} non deterministic atoms with respect to $I$ and all non deterministic atoms that do not belong to $I$ and belong to ${\cal N_L}$ are the {\em false} non deterministic atoms with respect to $I$.

\begin {definition} Let $\Pi$ be a non deterministic logic program. A non deterministic interpretation for $\Pi$ is a subset of the non deterministic base ${\cal N_L}$.
\end{definition}
The satisfaction of non deterministic logic programs with respect to non deterministic interpretations and non deterministic models of non deterministic logic programs are given by the following definitions.

\begin {definition} Let $\Pi$ be a ground non deterministic logic program and $I$ be a non deterministic interpretation. Then the satisfaction, denoted by $\models$, of a non deterministic atom and a non deterministic logic rule, by $I$, is defined as follows:

\begin{itemize}

\item $I \models \{ A_i \}_{i = 1}^{n}$ iff $\{ A_i \}_{i = 1}^{n} \in I$.

\item $I \models \{ B_{i_1} \}_{i_1 = 1}^{n_1} , \ldots, \{ B_{i_m} \}_{i_m = 1}^{n_m}$ iff for all  $ 1 \leq j
\leq m$, $I \models \{ B_{i_j} \}_{i_j = 1}^{n_j}$.

\item $I \models \{ A_i \}_{i = 1}^{n} \leftarrow \{ B_{i_1} \}_{i_1 = 1}^{n_1} , \ldots, \{ B_{i_m} \}_{i_m = 1}^{n_m}$ iff $I \models \{ A_i \}_{i = 1}^{n}$ whenever $I \models \{ B_{i_1} \}_{i_1 = 1}^{n_1} , \ldots, \{ B_{i_m} \}_{i_m = 1}^{n_m}$ or $I$ does not satisfy $\{ B_{i_1} \}_{i_1 = 1}^{n_1} , \ldots, \{ B_{i_m} \}_{i_m = 1}^{n_m}$.
\end{itemize}
\end {definition}

\begin {definition} Let $\Pi$ be a non deterministic logic program and $I$ be a non deterministic interpretation. Then, $I$ is a non deterministic model for $\Pi$ if $I$ satisfies every non deterministic logic rule in $\Pi$.
\end {definition}
We say a non deterministic atom is a logical consequence of a non deterministic logic program if this non deterministic atom is true in every non deterministic model of that non deterministic logic program. This is captured by the following definition.

\begin {definition} Let $\Pi$ be a non deterministic logic program and $\{ A_i \}_{i = 1}^{n} \in {\cal N_L}$. Then, $\{ A_i \}_{i = 1}^{n}$ is a logical consequence of $\Pi$ iff $\{ A_i \}_{i = 1}^{n}$ is true (satisfied) in every non deterministic model of $\Pi$.
\end {definition}
The set of all non deterministic interpretations of ${\cal L}$, denoted by ${\cal I_L}$ which is the set of all subsets of the non deterministic base ${\cal N_L}$, i.e., ${\cal I_L} = 2^{\cal N_L}$, along with the set-inclusion, $\subseteq$, forms a complete lattice $\langle 2^{\cal N_L},  \subseteq \rangle$. The top element of the lattice,$\langle 2^{\cal N_L},  \subseteq \rangle$, is the non deterministic base ${\cal N_L}$ and the bottom element is the empty set $\emptyset$.

\begin{lemma} The set of all non deterministic interpretations $2^{\cal N_L}$ and the set inclusion relation $\subseteq$ form a complete lattice $\langle 2^{\cal N_L},  \subseteq \rangle$. The join operation over $\langle 2^{\cal N_L},  \subseteq \rangle$ is the union operation $\cup$ and the the meet operation is the intersection operation $\cap$.
\label{lemma:lattice}
\end{lemma}
Every non deterministic logic program has several non deterministic models that satisfy each non deterministic logic rule in the non deterministic logic program. To provide a meaning for a non deterministic logic program, we use the least non deterministic model for the non deterministic logic program to be the meaning of the program and hence its declarative semantics. The following results show how to declaratively construct the least non deterministic model for non deterministic logic programs.

\begin {proposition} Let $\Pi$ be a non deterministic logic program and $I_1, I_2$ be non deterministic models of $\Pi$. Then $I_1 \cap I_2$ is also a non deterministic model of $\Pi$.
\label{prop:models}
\end{proposition}

\begin{theorem} Let $\Pi$ be a non deterministic logic program and let ${\cal I}_\Pi$ be the set of all non deterministic models of $\Pi$. Then, $I_\Pi = \bigcap_{I \in {\cal I}_\Pi} \; I $ is the least non deterministic model of $\Pi$.
\label{thm:least_pmodel}
\end{theorem}
Intuitively, the least non deterministic model, $I_\Pi$, of a non deterministic logic program, $\Pi$, is the smallest non deterministic model, with respect to the set inclusion $\subseteq$, that satisfies $\Pi$ which is unique.

\begin{definition}
The least non deterministic model, $I_\Pi$, of the non deterministic logic program, $\Pi$ is the intersection of all non deterministic models of $\Pi$   .
\end{definition}

\begin{lemma} Let $\Pi$ be a non deterministic logic program. The least non deterministic model $I_\Pi$ of $\Pi$ is unique.
\label{lemma:unique_least_pmodel}
\end{lemma}

\begin {proposition} Let $\Pi$ be a non deterministic logic program. Then $\Pi$ is unsatisfiable iff $\Pi$
has no non deterministic models.
\end {proposition}
The language of non deterministic logic programs syntax and semantics are designed to represent and reason about inherently non deterministic real-world applications whose solutions are described by trees. This means that the semantics of a non deterministic logic program representation of a non deterministic problem, described by the least non deterministic model of that non deterministic logic program, represents the solution tree of the represented problem. Therefore, to construct the solution tree represented in the least non deterministic model of a non deterministic logic program representation of a non deterministic problem we introduce the following definition.

\begin{definition}
Let $I_\Pi = \{\; \{ A_{i_1} \}_{i_1 = 1}^{n_1} , \{ A_{i_2} \}_{i_2 = 1}^{n_2}, \ldots, \{ A_{i_m} \}_{i_m = 1}^{n_m} \; \}$ be the least non deterministic model of the non deterministic logic program $\Pi$. Let $X_j$, for $1 \leq j \leq m$, be a variable ranging over the elements of $\{A_{i_j} \}_{i_j = 1}^{n_j} \in I_\Pi$ . Then, the set of answer sets, $S_\Pi$, corresponding to $I_\Pi$ is the set of all {\em minimal} sets formed from the elements of $I_\Pi$ such that
\[
S_\Pi = \{ \;  \{X_1, X_2, \ldots, X_m \} \:|\: \forall X_1 \; \forall X_2 \; \ldots \forall X_m \; \}.
\]
\end{definition}
The {\em set of answer sets} represents the solution tree of the represented non deterministic problem by a non deterministic logic program, where every {\em answer set} in the set of answer sets corresponds to a {\em branch} in the solution tree.

Observe that an answer set, $S$, in the set of answer sets, $S_\Pi$, that corresponds to the least non deterministic model, $I_\Pi$, of a non deterministic logic program, $\Pi$, is a subset of the Herbrand base ${\cal B_L}$. Intuitively, the meaning of an answer set is that every atom belongs to $S$ is true with respect to $S$ and every atom that does not belong to $S$ but belongs to ${\cal B_L}$ is false with respect to $S$.

\begin{example} The non deterministic logic program, $\Pi$, described in Example (\ref{ex:Fred}), has a least non deterministic model, $I_\Pi$. Including the relevant non deterministic atoms, $I_\Pi$ contains
\[
\left\{ \begin{array}{c}
\left\{
\begin{array}{r}
soup(beef), \\
salad(salmon)
\end{array}
\right\}
,
\left\{
\begin{array}{r}
soup(beef), \\
salad(seafood)
\end{array}
\right\}
,
\left\{
\begin{array}{r}
soup(buffalo), \\
salad(salmon)
\end{array}
\right\}
,
\left\{
\begin{array}{r}
soup(buffalo), \\
salad(seafood)
\end{array}
\right\}
,
\\
\\
\left\{
\begin{array}{r}
meat(beef), \\
fish(salmon)
\end{array}
\right\}
,
\left\{
\begin{array}{r}
meat(beef), \\
fish(seafood)
\end{array}
\right\}
,
\left\{
\begin{array}{r}
meat(buffalo), \\
fish(salmon)
\end{array}
\right\}
,
\left\{
\begin{array}{r}
meat(buffalo), \\
fish(seafood)
\end{array}
\right\},\\
\\
\{\; lunch(beef, salmon) \;\}, \;  \{\; lunch(beef, seafood) \;\}, \\ \;\{\; lunch(buffalo, salmon) \;\},
\{\;lunch(buffalo, seafood) \;\}
\end{array} \right\}
\]
The set of answer sets, $S_\Pi$, corresponding to the least non deterministic model, $I_\Pi$ of $\Pi$, that represents the solution tree of the non deterministic problem of Example (\ref{ex:Fred}) is given as follows, where each $S_i \in S_\Pi$, as described below, represents a branch of the solution tree of the problem in Example (\ref{ex:Fred}). We list below some of the answer sets from $S_\Pi$, since it is straight forward to construct the whole set of answer sets.

\[
\begin{array}{r}
S_1 = \{\;
lunch(beef, salmon),
lunch(beef, seafood),
lunch(buffalo, salmon), \\
lunch(buffalo, seafood),
soup(beef),
soup(buffalo), \\
meat(beef),
meat(buffalo)
\;\}
\end{array}
\]

\[
\begin{array}{r}
S_2 = \{ \;
lunch(beef, salmon),
lunch(beef, seafood),
lunch(buffalo, salmon), \\
lunch(buffalo, seafood),
soup(beef),
soup(buffalo), \\
meat(beef),
fish(seafood)
\; \}
\end{array}
\]

\[
\begin{array}{r}
S_3 = \{\;
lunch(beef, salmon),
lunch(beef, seafood),
lunch(buffalo, salmon), \\
lunch(buffalo, seafood),
soup(beef),
soup(buffalo), \\
fish(salmon),
meat(buffalo)
\;\}
\end{array}
\]

\[
\begin{array}{r}
S_4 = \{\;
lunch(beef, salmon),
lunch(beef, seafood),
lunch(buffalo, salmon), \\
lunch(buffalo, seafood),
soup(beef),
soup(buffalo), \\
fish(salmon),
fish(seafood)
\; \}
\end{array}
\]

\[
\begin{array}{r}
S_5 = \{\;
lunch(beef, salmon),
lunch(beef, seafood),
lunch(buffalo, salmon), \\
lunch(buffalo, seafood),
soup(beef),
soup(buffalo), \\
meat(beef),
meat(buffalo)
\;\}
\end{array}
\]

\[
\begin{array}{r}
S_6 = \{ \;
lunch(beef, salmon),
lunch(beef, seafood),
lunch(buffalo, salmon), \\
lunch(buffalo, seafood),
soup(beef),
soup(buffalo), \\
meat(beef),
fish(seafood)
\; \}
\end{array}
\]

\[
\begin{array}{r}
S_7 = \{\;
lunch(beef, salmon),
lunch(beef, seafood),
lunch(buffalo, salmon), \\
lunch(buffalo, seafood),
soup(beef),
soup(buffalo), \\
fish(salmon),
meat(beef),
meat(buffalo)
\;\}
\end{array}
\]

\[
\begin{array}{r}
S_8 = \{\;
lunch(beef, salmon),
lunch(beef, seafood),
lunch(buffalo, salmon), \\
lunch(buffalo, seafood),
soup(beef),
soup(buffalo), \\
meat(beef),
fish(salmon),
fish(seafood)
\; \}
\end{array}
\]
\end{example}

\begin{example} The non deterministic logic program, $\Pi$, presented in Example (\ref{ex:Connection}), has a least non deterministic model, $I_\Pi$. Including the relevant non deterministic atoms, in addition to every non deterministic fact appearing in $\Pi$, $I_\Pi$ contains
\[
\begin{array}{c}
\{
\\
\{\; reachable(home,rome) \; \},
\{\; reachable(rome,berlin) \; \},
\{\; reachable(rome,berlin) \; \}, \\
\{\; reachable(rome,london) \; \},
\{\; reachable(home,rome) \; \},
\{\; reachable(home,london) \; \}, \\
\{\; reachable(home,berlin) \; \},
\{\; reachable(rome,paris) \; \},
\{\; reachable(home,paris) \; \}
\\
\}
\end{array}
\]
The set of answer sets, $S_\Pi$, corresponding to the least non deterministic model, $I_\Pi$ of $\Pi$, that represents the solution tree of the non deterministic problem of Example (\ref{ex:Connection}) can be constructed in a straightforward way.
\end{example}

\subsection{Fixpoint Semantics}

In this section, we present the fixpoint semantics of non deterministic logic programs. The fixpoint semantics is based on the immediate consequence operator or the fixpoint operator of non deterministic logic programs which is used to compute the least non deterministic model of non deterministic logic programs inductively. Associated with each non deterministic logic program, $\Pi$, is an operator, $T_\Pi$, called the \emph{fixpoint operator}, which takes a non deterministic interpretation as an argument and returns a non deterministic interpretation. In this section we define the fixpoint operator of non deterministic logic programs
and show that every non deterministic model is a fixpoint of this operator. In addition, we show that the least fixpoint of the fixpoint operator coincides with the least non deterministic model of non deterministic logic programs.

The fixpoint semantics of non deterministic logic programs is considered as the operational counterpart of the non deterministic logic programs declarative semantics. The idea is based on the lattice theory. Let $I_1$ and $I_2$ be two non deterministic interpretations of a non deterministic logic program $\Pi$. Therefore, the non deterministic interpretations $I_1$ and $I_2$ are partially ordered under the subset inclusion $\subseteq$ iff $I_1 \subseteq I_2$. Consequently, the set of all non deterministic interpretations $2^{\cal N_L}$ and the set-inclusion forms a complete lattice $\langle 2^{\cal N_L},  \subseteq \rangle$. The bottom element of this lattice is the empty set $\emptyset$ and ${\cal N_L}$ is its top element. The meet (the lower bound) and join (the upper bound) operations associated with this lattice are the set intersection $\cap$ and the set union $\cup$ respectively.

\begin {definition} Let $\Pi$ be a ground non deterministic logic program and $I$ be a non deterministic interpretation. The immediate consequence operator $T_\Pi$ of $\Pi$ is the mapping $T_\Pi : 2^{\cal N_L} \rightarrow 2^{\cal N_L}$ which is defined as
\[
T_\Pi(I)= \{ \; \{ A_i \}_{i = 1}^{n} \; | \;  \{ A_i \}_{i = 1}^{n} \leftarrow \{ B_{i_1} \}_{i_1 = 1}^{n_1} , \ldots, \{ B_{i_m} \}_{i_m = 1}^{n_m} \in \Pi,
\]
and $\forall \: (1 \leq j \leq m), \{B_{i_j} \}_{i_j = 1}^{n_j} \in I \; \}$.
\end {definition}

\begin{lemma} Let $\Pi$ be a non deterministic logic program. Then $T_\Pi$ operator is monotonic
and continuous.
\label{lemma:Tp_mono_continue}
\end{lemma}

\begin{theorem} Let $\Pi$ be a non deterministic logic program and $I$ be a non deterministic interpretation. Then $I$ is a non deterministic model for $\Pi$ iff $T_\Pi(I) \subseteq I$.
\label{thm:Tp_subset_I}
\end{theorem}
The least non deterministic model, $I_\Pi$, of a non deterministic logic program, $\Pi$, can be constructed by the repeated iterations of the fixpoint operator, $T_\Pi$, as described by the following result.

\begin {definition} Let $\Pi$ be a non deterministic logic program and $T_\Pi$ be the immediate consequence operator of $\Pi$. Then

\begin{enumerate}
\item $T_\Pi \uparrow 0 = \emptyset$.

\item $T_\Pi \uparrow \alpha = T_\Pi (T_\Pi \uparrow (\alpha - 1 ) )$
where $\alpha$ is the successor ordinal of $(\alpha - 1)$.

\item $T_\Pi \uparrow \omega = \cup \{T_P \uparrow \alpha | \alpha <
\omega\}$ where $\omega$ is a limit ordinal.

\end{enumerate}
\end {definition}

\begin {theorem} Let $\Pi$ be a non deterministic logic program. Then $I_\Pi = lfp(T_\Pi)= T_\Pi \uparrow \omega$.
\label{thm:least_model_least_fixpoint}
\end {theorem}

\begin{example} It is easy to see that the least non deterministic models of the non deterministic logic programs described in Example (\ref{ex:Fred}) and Example (\ref{ex:Connection}) coincide with the least fixpoint of their corresponding immediate consequence operator $lfp(T_\Pi)$.

\end{example}

\subsection{Discussion}

In this section we show that the syntax and semantics of non deterministic logic programs subsume the syntax and semantics of the deterministic definite logic programs \cite{Lloyd}.

The model theoretic semantics and the fixpoint semantics of definite logic programs \cite{Lloyd} is deterministic in the sense that the definite logic programs and the model theoretic semantics and the fixpoint semantics defined for these definite logic programs  allow a single atom (deterministic atom) as the building block for the definite logic programs and model theoretic and the fixpoint semantics. However, non deterministic logic programs and their non deterministic model theoretic and fixpoint semantics allow atoms to be non deterministic for which a non deterministic atom is represented by a set of atoms of the form $\{A_i\}_{i = 1}^n$. This implies that any deterministic atom, $A$, representation in the language of deterministic definite logic programs \cite{Lloyd} can be represented as a non deterministic atom of the form, $\{A\}$, in the language of non deterministic logic programs. Consequently, it can be shown that the syntax and the model theoretic semantics and the fixpoint semantics of non deterministic logic programs naturally subsumes the syntax and the model theoretic semantics and the fixpoint semantics of deterministic definite logic programs \cite{Lloyd}.

Any deterministic definite logic program, $\Upsilon$, can be represented as a non deterministic logic program, $\Pi$, where each deterministic definite logic rule
\[
A \leftarrow B_1, \ldots, B_m \in \Upsilon
\]
can be represented as a non deterministic logic rule of the form
\[
\{A\} \leftarrow \{B_1\}, \ldots, \{B_m\} \in \Pi
\]
where $A, B_1, \ldots, B_m$ are atoms.

Observe that a Herbrand interpretation of a deterministic definite logic program, $\Upsilon$, is a subset of the Herbrand base ${\cal B_L}$, which is a set element in the non deterministic base ${\cal N_L}$. This means that a Herbrand interpretation, $I \subseteq {\cal B_L}$, of a deterministic definite logic program is a set element $I \in {\cal N_L}$ and not a subset of ${\cal N_L}$, i.e.,  $I \nsubseteq {\cal N_L}$. Therefore, Herbrand interpretations and Herbrand models for deterministic definite logic programs are deterministic Herbrand interpretations and deterministic Herbrand models.

\begin{theorem}
Let $\Upsilon$ be a deterministic definite logic program and $\Pi$ be the non deterministic logic program representation of $\Upsilon$. Then $I$ is a deterministic Herbrand model for $\Upsilon$ if and only if $\forall \: A \in I$, $\{A\} \in J$ is a non deterministic model for $\Pi$.
\label{thm:relate_to_definite}
\end{theorem}

\begin{theorem}
Let $\Upsilon$ be a deterministic definite logic program and $\Pi$ be the non deterministic logic program representation of $\Upsilon$. Then $I_\Upsilon$ is a least deterministic Herbrand model for $\Upsilon$ if and only if $\forall \: A \in I_\Upsilon$, $\{A\} \in J_\Pi$ is the least non deterministic model for $\Pi$.
\label{thm:equiv2definite}
\end{theorem}
The following example shows how the semantics of the non deterministic logic programs captures the semantics of deterministic definite logic programs.

\begin{example} Consider the following deterministic definite logic program, $\Upsilon$, that consists of the following deterministic definite logic rules
\[
\begin{array}{lcl}
a & \leftarrow& b \\
a& \leftarrow& c \\
a& \leftarrow& d, f \\
b & \leftarrow &\\
c & \leftarrow &
\end{array}
\]
The least deterministic Herbrand model, $I_\Upsilon$, of $\Upsilon$ is $I_\Upsilon = \{a, b, c\}$ which coincides with $lfp(T_\Upsilon)$. On the other hand, the non deterministic logic program, $\Pi$, equivalent to the deterministic definite logic program, $\Upsilon$, is given by
\[
\begin{array}{lcl}
\{\; a \; \}& \leftarrow & \{\; b \; \} \\
\{\; a \; \} & \leftarrow & \{\; c \; \} \\
\{\; a \; \} & \leftarrow & \{\; d \; \}, \{\; f \; \}  \\
\{\; b \; \} & \leftarrow & \\
\{\; c \; \} & \leftarrow &
\end{array}
\]
The least non deterministic model, $I_\Pi$, of $\Pi$ is $I_\Pi = \{\; \{a\}, \{b\}, \{c\} \; \}$ which coincides with $lfp(T_\Pi)$.
\end{example}

\section{Negation in Non Deterministic Logic Programs}

We want to extend the syntax and semantics of non deterministic logic programs to cope with non-monotonic negation. This is to enhance the semantics of non deterministic logic programs with the capabilities of performing default reasoning, which is an important feature in automated deduction systems based on logic. This is because non-monotonic negation is important to be able to perform default reasoning and deriving negative conclusions in the absence of positive conclusions. In addition, non-monotonic negation is very important because of its ability to support non-monotonic reasoning which has an essential role in capturing the fundamental aspects of commonsense reasoning. As a consequence, extending the language of non deterministic logic programs with this feature makes it more suitable for wider real-world applications.

Non-monotonic negation has been studied in deterministic logic programming by developing the notion of deterministic stable model semantics for deterministic normal logic programs \cite{Gelfond_A}, the notion of deterministic answer set semantics for deterministic extended and disjunctive logic programs \cite{Gelfond_B}, and the notion of deterministic well-founded semantics for deterministic normal logic programs \cite{Gelder}.

Therefore, in the rest of this paper, we extend the syntax and semantics of non deterministic logic programs to cope with \emph{non-monotonic negation}. We present the stable non deterministic model semantics and the well-founded non deterministic model semantics for the non deterministic logic programs with non-monotonic negation, namely \emph{normal non deterministic logic programs}.

The stable non deterministic model semantics is a generalization of the deterministic stable model semantics of deterministic normal logic programs. Analogous to the deterministic stable model semantics for deterministic normal logic programs, a non deterministic model is guessed and then verified whether it is a stable non deterministic model.

The well-founded non deterministic model semantics is a generalization of the deterministic well-founded semantics for deterministic normal logic programs. The definition of the well-founded non deterministic model semantics is developed in two steps. The first step is to derive the set of negative conclusions from a non deterministic logic program with non-monotonic negation by introducing the notion of the unfounded non deterministic set. The second step is to derive the set of positive conclusions by defining the notion of immediate consequence operator for non deterministic logic program with non-monotonic negation. Finally, the notion of well-founded non deterministic model is defined inductively in the well-founded non deterministic model semantics by combining the set of negative conclusions with the set of positive conclusions.

In addition, we show that the relationship between the stable non deterministic model semantics and the well-founded non deterministic model semantics of normal non deterministic logic programs preserves the relationship between the deterministic stable model semantics and the deterministic well-founded semantics for deterministic normal logic programs.

\section{Stable Non Deterministic Model Semantics}

In this section, we extend non deterministic logic programs to the notion of {\em normal non deterministic logic programs} to allow non-monotonic negation in the body of the non deterministic logic rules. The syntax of the language of normal non deterministic logic programs is the same as the syntax of non deterministic logic programs language but with the non-monotonic negation, $not$, added to the language. Stable non deterministic model semantics is defined to provide a meaning for normal non deterministic logic programs that have no unique minimal non deterministic model. In this semantics zero, one, or multiple minimal non deterministic models are the possible meaning for normal non deterministic logic program. The intuition behind the stable non deterministic model semantics is that, if a set of non deterministic atoms is a stable non deterministic model of a normal non deterministic logic program, then this set of non deterministic atoms must be able to derive itself from that normal non deterministic logic program.

\begin{definition}
A normal non deterministic logic rule is an expression of the form
\[
\{ A_i \}_{i = 1}^{n} \leftarrow \{ B_{i_1} \}_{i_1 = 1}^{n_1} , \ldots, \{ B_{i_l} \}_{i_l = 1}^{n_l}, not \: \{ B_{i_{l+1}} \}_{i_{l+1} = 1}^{n_{l+1}} , \ldots, not \: \{ B_{i_m} \}_{i_m = 1}^{n_m}.
\]
where $\{ A_i \}_{i = 1}^{n}, \{ B_{i_1} \}_{i_1 = 1}^{n_1} , \ldots, \{ B_{i_l} \}_{i_l = 1}^{n_l}, \{ B_{i_{l+1}} \}_{i_{l+1} = 1}^{n_{l+1}} , \ldots, \{ B_{i_m} \}_{i_m = 1}^{n_m}$ are non deterministic atoms. $\{ A_i \}_{i = 1}^{n}$ is called the head of the normal non deterministic logic rule and $\{ B_{i_1} \}_{i_1 = 1}^{n_1} , \ldots, \{ B_{i_l} \}_{i_l = 1}^{n_l}, not \: \{ B_{i_{l+1}} \}_{i_{l+1} = 1}^{n_{l+1}} , \ldots, not \: \{ B_{i_m} \}_{i_m = 1}^{n_m}$ is its body. If $m = 0$, the normal non deterministic logic rule is a non deterministic logic rule.
\end {definition}
The intuitive meaning of a normal non deterministic logic rule is that if for each $1 \leq j \leq l$ it is {\em believable} that $\{ B_{i_j} \}_{i_j = 1}^{n_j}$ is true (satisfied) and for every $l+1 \leq k \leq m$ it is {\em not believable} that $\{ B_{i_k} \}_{i_k = 1}^{n_k}$ is true, then  $\{ A_i \}_{i = 1}^{n}$ is true.

\begin {definition} A normal non deterministic logic program, $\Pi$, is a set of normal non deterministic logic rules.
\end {definition}
A normal non deterministic logic rule, $r$, is ground if every non deterministic atom appearing in $r$ is ground. A normal non deterministic logic program, $\Pi$, is ground if every normal non deterministic logic rule in $\Pi$ is ground.

Observe that the immediate consequence operator, $T_\Pi$, of non deterministic logic programs, $\Pi$, is monotonic and continuous and hence any non deterministic logic program has least fixpoint, $lfp(T_\Pi)$, non deterministic models. Moreover, the least fixpoint of the operator, $T_\Pi$, of non deterministic logic programs, $\Pi$, is also the least non deterministic model for $\Pi$. However, in dealing with non deterministic logic programs with negation,  {\em normal non deterministic
logic programs}, in general, the $T_\Pi$ operator is no longer monotonic and may have no fixpoints. For example, consider the following normal non deterministic logic program, $\Pi$, assuming that non deterministic atoms that does not appear in a non deterministic interpretation is false.

\begin{equation*}
\left\{
\begin{array}{r}
a_1, \\
a_2
\end{array}
\right\}
\leftarrow
not\;
\left\{
\begin{array}{r}
a_1, \\
a_2
\end{array}
\right\}
,
not\;
\left\{
\begin{array}{r}
b_1, \\
b_2
\end{array}
\right\}.
\end{equation*}
Consider also the non deterministic interpretation $\emptyset$. Applying the immediate consequence operator for non deterministic logic programs, $T_\Pi$, on the normal non deterministic logic program, $\Pi$, yields
$
T_\Pi(\emptyset) =
\left\{ \; \left\{
\begin{array}{r}
a_1, \\
a_2
\end{array}
\right\}\;
\right\}
$
but \\
$
T_\Pi(
\left\{ \; \left\{
\begin{array}{r}
a_1, \\
a_2
\end{array}
\right\}\;
\right\}
) = \emptyset
$,
and hence, there is neither a fixpoint nor least non deterministic model for $\Pi$ with respect to the operator $T_\Pi$. On the other hand, observe that the normal non deterministic logic program, $\Pi$, has two minimal non deterministic models which are
$
\left\{ \; \left\{
\begin{array}{r}
a_1, \\
a_2
\end{array}
\right\}\;
\right\}
$
and
$\left\{ \; \left\{
\begin{array}{r}
b_1, \\
b_2
\end{array}
\right\}\;
\right\}
$.
In addition, their intersection, which is the empty set $\emptyset$, is not a non deterministic model for $\Pi$.

As another example, consider the following normal non deterministic logic program, $\Pi$, that consists of the following normal non deterministic logic rules

\[
\begin{array}{lcl}
\left\{
\begin{array}{r}
a_1, \\
a_2
\end{array}
\right\}
&\leftarrow&
\\
\\
\left\{
\begin{array}{r}
b_1, \\
b_2
\end{array}
\right\}
&\leftarrow&
not\;
\left\{
\begin{array}{r}
c_1, \\
c_2
\end{array}
\right\}.
\\ \\
\left\{
\begin{array}{r}
c_1, \\
c_2
\end{array}
\right\}
&\leftarrow&
\left\{
\begin{array}{r}
c_1, \\
c_2
\end{array}
\right\}
,
not\;
\left\{
\begin{array}{r}
a_1, \\
a_2
\end{array}
\right\}.
\end{array}
\]
Although this normal non deterministic logic program, $\Pi$, contains negated non deterministic atoms, it has a unique minimal non deterministic model, which is
$\left\{ \; \left\{
\begin{array}{r}
a_1, \\
a_2
\end{array}
\right\}
,
\left\{
\begin{array}{r}
b_1, \\
b_2
\end{array}
\right\}
\;\right\}
$, that coincides with the least fixpoint of the immediate consequence operator, $T_\Pi$, of non deterministic logic programs, which is equivalent to
$
lfp(T_\Pi) =
\left\{ \; \left\{
\begin{array}{r}
a_1, \\
a_2
\end{array}
\right\}
,
\left\{
\begin{array}{r}
b_1, \\
b_2
\end{array}
\right\}
\;\right\}
$.

The following definitions describe the notions of non deterministic interpretations, satisfaction, and non deterministic models for normal non deterministic logic programs.

\begin {definition} Let $\Pi$ be a normal non deterministic logic program. A non deterministic interpretation for $\Pi$ is a subset of the non deterministic base ${\cal N_L}$.
\end {definition}

\begin {definition} Let $\Pi$ be a ground normal non deterministic logic program and $I$ be a non deterministic interpretation. Then the notion of satisfaction, denoted by $\models$, of a non deterministic atom and a normal non deterministic logic rule, by $I$, is defined as follows:

\begin{itemize}

\item $I \models \{ B_{i_j} \}_{i_j = 1}^{n_j}$ iff $\{ B_{i_j} \}_{i_j = 1}^{n_j} \in I$.

\item $I \models not \; \{ B_{i_k} \}_{i_k = 1}^{n_k}$ iff $\{ B_{i_k} \}_{i_k = 1}^{n_k} \notin I$.

\item $I \models \{ B_{i_1} \}_{i_1 = 1}^{n_1} , \ldots, \{ B_{i_l} \}_{i_l = 1}^{n_l}, not \: \{ B_{i_{l+1}} \}_{i_{l+1} = 1}^{n_{l+1}} , \ldots, not \: \{ B_{i_m} \}_{i_m = 1}^{n_m}$
    iff $\forall (1 \leq j \leq l)$ $I \models \{ B_{i_j} \}_{i_j = 1}^{n_j}$ and $\forall (l+1 \leq k \leq m$) $I \models not \; \{ B_{i_k} \}_{i_k = 1}^{n_k}$

\item $I \models \{ A_i \}_{i = 1}^{n} \leftarrow \{ B_{i_1} \}_{i_1 = 1}^{n_1} , \ldots, \{ B_{i_l} \}_{i_l = 1}^{n_l}, not \: \{ B_{i_{l+1}} \}_{i_{l+1} = 1}^{n_{l+1}} , \ldots, not \: \{ B_{i_m} \}_{i_m = 1}^{n_m}$
    iff $I \models \{ A_i \}_{i = 1}^{n}$ whenever 
    
    $I \models \{ B_{i_1} \}_{i_1 = 1}^{n_1} , \ldots, \{ B_{i_l} \}_{i_l = 1}^{n_l}, not \: \{ B_{i_{l+1}} \}_{i_{l+1} = 1}^{n_{l+1}} , \ldots, not \: \{ B_{i_m} \}_{i_m = 1}^{n_m}$ or $I$ does not satisfy $\{ B_{i_1} \}_{i_1 = 1}^{n_1} , \ldots, \{ B_{i_l} \}_{i_l = 1}^{n_l}, not \: \{ B_{i_{l+1}} \}_{i_{l+1} = 1}^{n_{l+1}} , \ldots, not \: \{ B_{i_m} \}_{i_m = 1}^{n_m}$.
\end{itemize}
\end {definition}

\begin {definition} Let $\Pi$ be a normal non deterministic logic program and $I$ be a non deterministic interpretation. Then, $I$ is a non deterministic model for $\Pi$ if $I$ satisfies every normal non deterministic logic rule in $\Pi$.
\end {definition}
In order to develop stable non deterministic model semantics for normal non deterministic logic programs, we define the notion of stable non deterministic models. A stable non deterministic model is given in two steps. The first step is to guess a non deterministic model, $I$, for a certain normal non deterministic logic program $\Pi$. Then, we define the non deterministic reduct of $\Pi$ with respect to $I$, denoted by $\Pi^I$, which is a non deterministic logic program, and then apply the fixpoint operator, $T_{\Pi^I}$ on the non deterministic reduct, $\Pi^I$, to verify whether $I$ is the least non deterministic model of the reduct, which in turn is the stable non deterministic model of $\Pi$. Intuitively, a stable non deterministic model is the set of {\em non deterministic beliefs} that a rational agent believes true.

\begin {definition} Let $\Pi$ be a ground normal non deterministic logic program and $I$ be a non deterministic interpretation. The non deterministic reduct $\Pi^I$ of $\Pi$ w.r.t. $I$ is the non deterministic logic program $\Pi^I$ such that
\[
\Pi^I = \left\{\begin{array}{l} \{ A_i \}_{i = 1}^{n} \leftarrow \{ B_{i_1} \}_{i_1 = 1}^{n_1} , \ldots, \{ B_{i_l} \}_{i_l = 1}^{n_l} \qquad  |
\\ \\
\{ A_i \}_{i = 1}^{n} \leftarrow \{ B_{i_1} \}_{i_1 = 1}^{n_1} , \ldots, \{ B_{i_l} \}_{i_l = 1}^{n_l}, not \: \{ B_{i_{l+1}} \}_{i_{l+1} = 1}^{n_{l+1}} , \ldots, not \: \{ B_{i_m} \}_{i_m = 1}^{n_m} \in \Pi,
\\ \\
and \; \forall (l+1 \leq k \leq m$) $\{ B_{i_k} \}_{i_k = 1}^{n_k} \notin I \end{array}
\right\}.
\]
\label{def:prob_reduct}
\end {definition}

\begin {definition} A non deterministic interpretation, $I$, is a stable non deterministic model for the normal non deterministic logic program, $\Pi$, if $I$ is the least non deterministic model of $\Pi^I$.
\end {definition}
Similar to non deterministic models, non deterministic atoms that belong to a stable non deterministic model, $I$, are true with respect to $I$, but non deterministic atoms that do not belong to $I$ but belong to the non deterministic base, ${\cal N_L}$, are false with respect to $I$. The following theorem establishes the relationship between stable non deterministic models and minimal non deterministic models for normal non deterministic logic programs.

\begin {theorem} Any stable non deterministic model for a normal non deterministic logic program, $\Pi$, is a minimal non deterministic model for $\Pi$.
\label{thm:stable_model_minimal}
\end{theorem}
Consequently, each non deterministic logic program has only one stable non deterministic model, which is its only least non deterministic model.

\begin{theorem} Every non deterministic logic program, $\Pi$, has a unique stable non deterministic model, $I$, iff $I$ is the least non deterministic model for $\Pi$.
\label{thm:sp-non-deterministic-program}
\end {theorem}
The syntax and semantics of normal non deterministic logic programs are developed to represent and reason about non deterministic real-world applications whose solutions are zero, one, or multiple trees. This means that every stable non deterministic model of a normal non deterministic logic program representation of a non deterministic problem, represents one solution tree of the represented problem. This also implies that all the stable non deterministic models of a normal non deterministic logic program representation of a non deterministic problem represent all the solution trees of the represented problem. Therefore, to construct a solution tree per a stable non deterministic model of a normal non deterministic logic program representation of a non deterministic problem we introduce the following definition.

\begin{definition}
Let $I = \{\; \{ A_{i_1} \}_{i_1 = 1}^{n_1} , \{ A_{i_2} \}_{i_2 = 1}^{n_2}, \ldots, \{ A_{i_m} \}_{i_m = 1}^{n_m} \; \}$ be a stable non deterministic model of a normal non deterministic logic program $\Pi$. Let $X_j$, for $1 \leq j \leq m$, be a variable ranging over the elements of $\{A_{i_j} \}_{i_j = 1}^{n_j} \in I$ . Then, the set of answer sets, $S_I$, corresponding to $I$ is the set of all {\em minimal} sets formed from the elements of $I$ such that
\[
S_I = \{ \;  \{X_1, X_2, \ldots, X_m \} \:|\: \forall X_1 \; \forall X_2 \; \ldots \forall X_m \; \}.
\]
\end{definition}
The {\em set of answer sets} represents one solution tree of the represented non deterministic problem by a normal non deterministic logic program and corresponds to one of its stable non deterministic models, where every {\em answer set} in the set of answer sets corresponds to a {\em branch} in the solution tree.

Observe that an answer set, $S$, in the set of answer sets, $S_I$, that corresponds to a stable non deterministic model, $I$, of a normal non deterministic logic program, $\Pi$, is a subset of the Herbrand base ${\cal B_L}$. Intuitively, the meaning of an answer set, $S$, is that every atom belongs to $S$ is true with respect to $S$ and every atom that does not belong to $S$ but belongs to ${\cal B_L}$ is false with respect to $S$.

\section{Examples}

\begin {example} Consider an instructor who decides on which course to teach in a given semester. The instructor's options are math 101, math 102, stat 101, and stat 102. However, due to the Math department constraints, the instructor has to make his choices according to the following. The instructor can choose either math 101 or math 102 if he decides not to choose stat 101 or stat 102. Otherwise, the instructor can choose either stat 101 or stat 102 if he decides not to choose math 101 or math 102. Then the head of the Math department decides which course the instructor would teach after the instructor makes his decision. This decision problem can be represented as a normal non deterministic logic program, $\Pi$, that consists of the normal non deterministic logic rules:
\[
\begin{array}{lcl}
\left\{
\begin{array}{r}
math(101), \\
math(102)
\end{array}
\right\}
&\leftarrow&
not \;
\left\{
\begin{array}{r}
stat(101), \\
stat(102)
\end{array}
\right\}.
\\ \\
\left\{
\begin{array}{r}
stat(101), \\
stat(102)
\end{array}
\right\}
&\leftarrow&
not\;
\left\{
\begin{array}{r}
math(101), \\
math(102)
\end{array}
\right\}.
\end{array}
\]
This normal non deterministic logic program, $\Pi$, has two stable non deterministic models which are
$
I_1 = \left\{ \left\{
\begin{array}{r}
math(101), \\
math(102)
\end{array}
\right\} \right\}
$
and
$
I_2 = \left\{ \left\{
\begin{array}{r}
stat(101), \\
stat(102)
\end{array}
\right\} \right\}
$.
This is because the non deterministic reduct, $\Pi^{I_1}$, of $\Pi$ with respect to $I_1$ is given by
\[
\begin{array}{lcl}
\left\{
\begin{array}{r}
math(101), \\
math(102)
\end{array}
\right\}
&\leftarrow&
\end{array}
\]
and $lfp(T_{\Pi^{I_1}}) = \left\{ \left\{
\begin{array}{r}
math(101), \\
math(102)
\end{array}
\right\} \right\}$.
Similarly, the non deterministic reduct, $\Pi^{I_2}$, of $\Pi$ with respect to $I_2$ is given by
\[
\begin{array}{lcl}
\left\{
\begin{array}{r}
stat(101), \\
stat(102)
\end{array}
\right\}
&\leftarrow&
\end{array}
\]
and $lfp(T_{\Pi^{I_2}}) = \left\{ \left\{
\begin{array}{r}
stat(101), \\
stat(102)
\end{array}
\right\} \right\}$. The non deterministic interpretation
$
I = \left\{ \left\{
\begin{array}{r}
math(101), \\
math(102)
\end{array}
\right\}
,
\left\{
\begin{array}{r}
stat(101), \\
stat(102)
\end{array}
\right\} \right\}
$
is not a stable non deterministic model for $\Pi$ because the non deterministic reduct, $\Pi^I$, of $\Pi$ with respect to $I$ is the empty set, $\emptyset$, and hence, $lfp(T_{\Pi^{I}}) = \emptyset \neq I$.

The set of answer sets, $S_{I_1}$, corresponding to the stable non deterministic model, $I_1$, consists of the following answer sets:

\[
\begin{array}{lcl}
S_1 = \{\;  math(101) \; \}
\\
S_2 = \{\;  math(102) \; \}
\end{array}
\]
In addition, the set of answer sets, $S_{I_2}$, corresponding to the stable non deterministic model, $I_2$, consists of the following answer sets:

\[
\begin{array}{lcl}
S_1 = \{\;  stat(101) \; \}
\\
S_2 = \{\;  stat(102) \; \}
\end{array}
\]

\label{ex:teaching}
\end {example}

\begin{example} Consider that the head of the Math department decides on which course the instructor teaches in the given semester according to the following. If the instructor chooses either math 101 or math 102, then the department head assigns math 102 to the instructor. But, if the instructor chooses either stat 101 or stat 102, then the department head assigns stat 101 to the instructor. This decision problem can be represented as a normal non deterministic logic program, $\Pi$, that consists of the normal non deterministic logic rules:

\[
\begin{array}{lcl}
\left\{
\begin{array}{r}
math(101), \\
math(102)
\end{array}
\right\}
&\leftarrow&
not \;
\left\{
\begin{array}{r}
stat(101), \\
stat(102)
\end{array}
\right\}.
\\ \\
\left\{
\begin{array}{r}
stat(101), \\
stat(102)
\end{array}
\right\}
&\leftarrow&
not\;
\left\{
\begin{array}{r}
math(101), \\
math(102)
\end{array}
\right\}.
\\
\\
\{\;
math(102)
\; \}
&\leftarrow&
\left\{
\begin{array}{r}
math(101), \\
math(102)
\end{array}
\right\}.
\\ \\
\{\;
stat(101)
\;\}
&\leftarrow&
\left\{
\begin{array}{r}
stat(101), \\
stat(102)
\end{array}
\right\}.

\end{array}
\]
This normal non deterministic logic program, $\Pi$, has two stable non deterministic models which are
$
I_1 = \left\{ \left\{
\begin{array}{r}
math(101), \\
math(102)
\end{array}
\right\},
\{\;
math(102)
\;\}
\right\}
$
and
$
I_2 = \left\{ \left\{
\begin{array}{r}
stat(101), \\
stat(102)
\end{array}
\right\},
\{\;
stat(101)
\;\}
\right\}
$.
This is because the non deterministic reduct, $\Pi^{I_1}$, of $\Pi$ with respect to $I_1$ is given by
\[
\begin{array}{lcl}
\left\{
\begin{array}{r}
math(101), \\
math(102)
\end{array}
\right\}
&\leftarrow&
\\
\\
\{\;
math(102)
\; \}
&\leftarrow&
\left\{
\begin{array}{r}
math(101), \\
math(102)
\end{array}
\right\}.
\\ \\
\{\;
stat(101)
\;\}
&\leftarrow&
\left\{
\begin{array}{r}
stat(101), \\
stat(102)
\end{array}
\right\}.
\end{array}
\]
and $lfp(T_{\Pi^{I_1}}) = \left\{ \left\{
\begin{array}{r}
math(101), \\
math(102)
\end{array}
\right\}
,
\{\;
math(102)
\; \}
\right
\}$.

The set of answer sets, $S_{I_1}$, corresponding to the stable non deterministic model, $I_1$, consists of the following answer sets:

\[
\begin{array}{lcl}
S_1 = \{\;  math(101), math(102) \; \}
\\
S_2 = \{\;  math(102) \; \}
\end{array}
\]
Similarly, the non deterministic reduct, $\Pi^{I_2}$, of $\Pi$ with respect to $I_2$ is given by
\[
\begin{array}{lcl}
\left\{
\begin{array}{r}
stat(101), \\
stat(102)
\end{array}
\right\}
&\leftarrow&
\\ \\
\{\;
math(102)
\; \}
&\leftarrow&
\left\{
\begin{array}{r}
math(101), \\
math(102)
\end{array}
\right\}.
\\ \\
\{\;
stat(101)
\;\}
&\leftarrow&
\left\{
\begin{array}{r}
stat(101), \\
stat(102)
\end{array}
\right\}.
\end{array}
\]
and $lfp(T_{\Pi^{I_2}}) = \left\{ \left\{
\begin{array}{r}
stat(101), \\
stat(102)
\end{array}
\right\}
,
\{\;
stat(101)
\;\}
\right\}$.

The set of answer sets, $S_{I_2}$, corresponding to the stable non deterministic model, $I_2$, consists of the following answer sets:

\[
\begin{array}{lcl}
S_1 = \{\;  stat(101) \; \}
\\
S_2 = \{\;  stat(101), stat(102) \; \}
\end{array}
\]
\end{example}

\begin{example} Consider the following normal non deterministic logic program, $\Pi$, that consists of the normal non deterministic logic rules:
\[
\begin{array}{lcl}
\left\{
\begin{array}{r}
a_1, \\
a_2
\end{array}
\right\}
&\leftarrow&
\\
\\
\left\{
\begin{array}{r}
b_1, \\
b_2
\end{array}
\right\}
&\leftarrow&
\left\{
\begin{array}{r}
a_1, \\
a_2
\end{array}
\right\}.
\\
\\
\left\{
\begin{array}{r}
c_1, \\
c_2
\end{array}
\right\}
&\leftarrow&
not\;
\left\{
\begin{array}{r}
c_1, \\
c_2
\end{array}
\right\}.
\end{array}
\]
The normal non deterministic logic program, $\Pi$, has no stable non deterministic models. This is because the non deterministic atom $\left\{
\begin{array}{r}
c_1, \\
c_2
\end{array}
\right\}$
appearing in the above normal non deterministic logic program, $\Pi$, has two choices in any stable non deterministic model for $\Pi$. If $\left\{
\begin{array}{r}
c_1, \\
c_2
\end{array}
\right\}$
is assumed to be in any  stable non deterministic model, $I$, for $\Pi$, then the non deterministic reduct, $\Pi^I$, for $\Pi$ with respect to $I$ excludes the the normal non deterministic logic rule
\[
\left\{
\begin{array}{r}
c_1, \\
c_2
\end{array}
\right\}
\leftarrow
not\;
\left\{
\begin{array}{r}
c_1, \\
c_2
\end{array}
\right\}
\]
from $\Pi$ and hence $\Pi^I$ consists of the normal non deterministic logic rules
\[
\begin{array}{lcl}
\left\{
\begin{array}{r}
a_1, \\
a_2
\end{array}
\right\}
&\leftarrow&
\\
\\
\left\{
\begin{array}{r}
b_1, \\
b_2
\end{array}
\right\}
&\leftarrow&
\left\{
\begin{array}{r}
a_1, \\
a_2
\end{array}
\right\}.
\end{array}
\]
Consequently,
$\left\{
\begin{array}{r}
c_1, \\
c_2
\end{array}
\right\} \notin lfp(T_{\Pi^I})$.
However, since $I$ is a stable non deterministic model for $\Pi$ and
$\left\{
\begin{array}{r}
c_1, \\
c_2
\end{array}
\right\} \in I$
but also $\left\{
\begin{array}{r}
c_1, \\
c_2
\end{array}
\right\} \notin I = lfp(T_{\Pi^I})$, a contradiction. Thus, $\left\{
\begin{array}{r}
c_1, \\
c_2
\end{array}
\right\}$
must not be in any stable non deterministic model for $\Pi$. On the other hand, assume that
$\left\{
\begin{array}{r}
c_1, \\
c_2
\end{array}
\right\}$
does not belong to any stable non deterministic model, $I$, for $\Pi$, then the non deterministic reduct, $\Pi^I$, for $\Pi$ with respect to $I$ consists of the normal non deterministic logic rules
\[
\begin{array}{lcl}
\left\{
\begin{array}{r}
a_1, \\
a_2
\end{array}
\right\}
&\leftarrow&
\\
\\
\left\{
\begin{array}{r}
b_1, \\
b_2
\end{array}
\right\}
&\leftarrow&
\left\{
\begin{array}{r}
a_1, \\
a_2
\end{array}
\right\}.
\\
\\
\left\{
\begin{array}{r}
c_1, \\
c_2
\end{array}
\right\}
&\leftarrow&
\end{array}
\]
Consequently,
$\left\{
\begin{array}{r}
c_1, \\
c_2
\end{array}
\right\} \in lfp(T_{\Pi^I})$.
However, since $I$ is a stable non deterministic model for $\Pi$ and
$\left\{
\begin{array}{r}
c_1, \\
c_2
\end{array}
\right\} \notin I$
but also $\left\{
\begin{array}{r}
c_1, \\
c_2
\end{array}
\right\} \in I = lfp(T_{\Pi^I})$, a contradiction. Thus, $\left\{
\begin{array}{r}
c_1, \\
c_2
\end{array}
\right\}$ must be in any stable non deterministic model for $\Pi$. This implies that the normal non deterministic logic program, $\Pi$, cannot have any stable non deterministic model.

However, removing the last normal non deterministic logic rule from $\Pi$ yields a stable non deterministic model for $\Pi$ which is
\[
\left\{ \left\{
\begin{array}{r}
a_1, \\
a_2
\end{array}
\right\}
,
\left\{
\begin{array}{r}
b_1, \\
b_2
\end{array}
\right\}
\right\}.
\]
\end{example}

\section{The Security Robot Example}

In this section, we show that the security robot planning problem described in Example (\ref{ex:secure-robot}) can be intuitively represented as a normal non deterministic logic program, $\Pi$, as follows.

Each action in the security robot planning problem, described in Example (\ref{ex:secure-robot}), is represented as a non deterministic fact as
\begin{eqnarray}
\{\; action(close) \; \}  \leftarrow \\
\{ \; action(flip\_lock) \; \}  \leftarrow \\
\{ \; action(check) \; \} \leftarrow \\
\{ \; action(inspect) \; \}  \leftarrow
\end{eqnarray}
The various properties of the security robot planning domain are represented as the non deterministic facts

\begin{eqnarray}
\{ \; atom(opened) \; \}  \leftarrow
\\
\{ \; atom(\neg opened) \; \}  \leftarrow
\\
\left\{
\begin{array}{r}
atom(opened), \\
atom(\neg opened)
\end{array}
\right\}
\leftarrow
\\
\left\{
\begin{array}{r}
atom(locked), \\
atom(\neg locked)
\end{array}
\right\}
\leftarrow
\end{eqnarray}
The following normal non deterministic logic rules specify that each atom, $A$, in a non deterministic atom, and its negation, $\neg A$, are contrary forming the atom $contrary(A, \neg A)$ in a non deterministic atom.
\begin{eqnarray}
\{ \; contrary(A_i, \neg A_i) \; \}_{i = 1}^{n} & \leftarrow & \{ \; atom(A_i)\; \}_{i = 1}^{n}  \label{eq:contrary1} \\
\{ \; contrary(\neg A_i, A_i) \; \}_{i = 1}^{n} & \leftarrow & \{ \; atom(A_i) \; \}_{i = 1}^{n} \label{eq:contrary2}
\end{eqnarray}
The possible initial state of the security robot planning domain is represented by the non deterministic facts
\begin{eqnarray}
\{ \; holds(opened, 0) \; \} \leftarrow \\
\{ \; holds(\neg locked, 0) \; \} \leftarrow
\end{eqnarray}
The executability conditions of the actions in the security robot planning domain are represented by the following normal non deterministic logic rules, where $T$ represents the time moment in which the action is executed.

\begin{eqnarray}
\{ \; exec(close, T) \; \}  \leftarrow \\
\{ \; exec(flip\_lock, T) \; \}  \leftarrow \\
\{ \; exec(check, T) \; \} \leftarrow \\
\{ \; exec(inspect, T) \; \}  \leftarrow
\end{eqnarray}
The effect of the non-sensing action $close$ is represented by the following normal non deterministic logic rule, which states that if the action $close$ occurs at time moment $T$ and the property $opened$ holds at the same time moment, then the property $\neg opened$ holds at time moment $T+1$.
\begin{equation}
\{ \; holds(\neg opened,T+1) \; \}  \leftarrow  \{ \; occ(close,T) \; \}, \{ \; exec(close, T) \; \}, \{ \; holds(opened,T) \; \}
\end{equation}
The effects of the non-sensing action $flip\_lock$ are represented by the following normal non deterministic logic rule, which states that if the action $flip\_lock$ occurs at time moment $T$ and the property $locked$ or the property $\neg locked$ holds at the same time moment, $T$, then the property $\neg locked$ or the property $locked$ holds at time moment $T+1$.
\begin{eqnarray}
\left \{ \begin{array}{r} holds(locked, T+1), \\  holds(\neg locked, T+1) \end{array} \right\} \leftarrow
\{ \; occ(flip\_lock,T) \; \}, \{ \; exec(flip\_lock, T) \; \}, \notag \\
\left \{\begin{array}{r} holds(\neg locked, T), \\  holds(locked, T) \end{array} \right\}
\end{eqnarray}
The effects of the sensing action $check$ are represented by the following normal non deterministic logic rule, which states that if the sensing action $check$ occurs at time moment, $T$, then the sensed property $opened$ is sensed to be known true or the sensed property $\neg opened$ is sensed to be known true at time moment $T+1$.

\begin{equation}
\left \{ \begin{array}{r} holds(opened, T+1), \\  holds(\neg opened, T+1) \end{array} \right\} \leftarrow
\{ \; occ(check,T) \; \}, \{ \;  exec(check, T) \; \}.
\end{equation}
The effects of the sensing action $inspect$ are represented by the following normal non deterministic logic rule, which states that if the sensing action $inspect$ occurs at time moment, $T$, then the sensed property $locked$ is sensed to be known true or the sensed property $\neg locked$ is sensed to be known true at time moment $T+1$.

\begin{equation}
\left \{ \begin{array}{r} holds(locked, T+1), \\  holds(\neg locked, T+1) \end{array} \right\} \leftarrow
\{ \; occ(inspect,T) \; \} , \{ \; exec(inspect, T) \; \}.
\end{equation}
The following normal non deterministic logic rule says that the non deterministic atom, $\{A_i\}_{i = 1}^n$, continues to hold at the time moment, $T+1$, if $\{A_i\}_{i = 1}^n$ holds at the time moment, $T$, and $\{A_i\}_{i = 1}^n$ is not a {\em subset} of any non deterministic atom at the time moment, $T+1$, and its contrary, $\{A_i'\}_{i = 1}^n$, does not hold at the time moment, $T+1$, (the frame axiom).
\begin{eqnarray}
\{ \; holds(A_i,T+1) \; \}_{i = 1}^n \leftarrow \{ \; holds(A_i,T) \; \}_{i = 1}^n,  not \; \{ \; holds(A_i',T+1) \; \}_{i = 1}^n,
\notag \\
not  \; \{ \; holds(A_i,T+1),  holds(B_i,T+1) \; \}_{i = 1}^n,
\notag\\
\{ \; contrary(A_i,A_i') \; \}_{i = 1}^n \label{eq:frame}
\end{eqnarray}
The following normal non deterministic logic rule represents that every atom, $A$, in a non deterministic atom, and its negation, $\neg A$, in another non deterministic atom cannot hold at the same time, where $\Gamma$ is a special non deterministic atom that does not appear in $\Pi$.

\begin{eqnarray}
\{ \; \Gamma \; \} \leftarrow not \; \{ \; \Gamma \; \}, \{ \; holds(A_i,T) \; \}_{i = 1}^n, \{ \; holds(\neg A_i,T) \; \}_{i = 1}^n \label{eq:inconsistent}
\end{eqnarray}
Actions are generated by the following normal non deterministic logic rules which generate action occurrences once at a time, where $C$ and $C'$ are variables representing actions.
\begin{eqnarray}
\{ \; occ(C,T) \; \}   \leftarrow  \{ \; action(C) \; \}, not \; \{ \; abocc(C,T) \; \} \label{eq:action1} \\
\{ \; abocc(C,T) \; \}   \leftarrow  \{ \; occ(C',T) \; \}, \{ \; C \neq C' \; \}
\label{eq:action2}
\end{eqnarray}
The goal of the security robot is to have a window $not \; opened$ and $locked$, which is represented by the following normal non deterministic logic rules.

\begin{eqnarray}
\{ \; goal(T) \; \}   \leftarrow  \{ \; holds(\neg opened, T) \;
\},
\left \{ \begin{array}{r} holds(locked, T), \\  holds(\neg locked, T) \end{array} \right\}
.
\\ \notag \\
\{ \; goal(T) \; \}   \leftarrow
\left \{ \begin{array}{r} holds(opened, T), \\  holds(\neg opened, T) \end{array} \right\},
\left \{ \begin{array}{r} holds(locked, T), \\  holds(\neg locked, T) \end{array} \right\}.
\end{eqnarray}

Considering a three steps plans, i.e., $T = 0, 1, 2.$, the security robot planning problem represented by the normal non deterministic logic program, $\Pi$, described above, has $64$ stable non deterministic models only $30$ stable non deterministic models of them correspond to valid {\em conditional plans} for this security robot planning problem.

We list below three different stable non deterministic models of $\Pi$ that represent three different valid conditional plans of the security robot planning problem described in Example (\ref{ex:secure-robot}). In addition, we list the set of answer sets of each of these stable non deterministic models which corresponds to a solution tree, a conditional plan in this case,  where each of these answer sets represents a {\em trajectory} in the conditional plan for the security robot planning problem described in Example (\ref{ex:secure-robot}). These stable non deterministic models of $\Pi$ that represent the three different conditional plans of the security robot planning problem are denoted by $I_1$, $I_2$, and $I_3$, where
\[
\begin{array}{c}
I_1 = \{
\\
\\
occ(close,0), occ(check,1), occ(inspect,2),  \\
holds(opened,0), \\
holds(\neg locked,0) \\
holds(\neg opened,1), \\
holds( \neg locked,1), \\
holds( \neg locked,2), \\
\\
\left\{ \begin{array}{r} holds(opened, 2), \\ holds(\neg opened, 2 ) \end{array} \right \}
,
\left\{ \begin{array}{r} holds(opened, 3), \\ holds(\neg opened, 3 ) \end{array} \right \}
,
\left\{ \begin{array}{r} holds(locked, 3), \\ holds(\neg locked, 3 ) \end{array} \right \},
\\
goal
\\
\\
\}
\end{array}
\]
The set of answer sets, $S_{I_1}$, that corresponds to the stable non deterministic model, $I_1$, is listed by the following answer sets.
\begin{eqnarray*}
S_1 = \{ \; occ(close,0), occ(check,1), occ(inspect,2), \\
holds(opened,0),
holds(\neg locked,0) , \\
holds(\neg opened,1),
holds( \neg locked,1),
holds( \neg locked,2),
holds(opened, 2), \\
holds(opened, 3),
holds(locked, 3), goal
\; \}
\end{eqnarray*}
\begin{eqnarray*}
S_2 = \{ \; occ(close,0), occ(check,1), occ(inspect,2), \\
holds(opened,0),
holds(\neg locked,0) , \\
holds(\neg opened,1),
holds( \neg locked,1),
holds( \neg locked,2),
holds(opened, 2), \\
holds(opened, 3),
holds(\neg locked, 3), goal
\; \}
\end{eqnarray*}
\begin{eqnarray*}
S_3  = \{ \; occ(close,0), occ(check,1), occ(inspect,2), \\
holds(opened,0),
holds(\neg locked,0) , \\
holds(\neg opened,1),
holds( \neg locked,1),
holds( \neg locked,2),
holds(opened, 2), \\
holds(\neg opened, 3),
holds(locked, 3), goal
\; \}
\end{eqnarray*}
\begin{eqnarray*}
S_4  = \{ \; occ(close,0), occ(check,1), occ(inspect,2), \\
holds(opened,0),
holds(\neg locked,0) , \\
holds(\neg opened,1),
holds( \neg locked,1),
holds( \neg locked,2),
holds(opened, 2), \\
holds(\neg opened, 3),
holds(\neg locked, 3), goal
\; \}
\end{eqnarray*}
\begin{eqnarray*}
S_5 = \{ \; occ(close,0), occ(check,1), occ(inspect,2), \\
holds(opened,0),
holds(\neg locked,0) , \\
holds(\neg opened,1),
holds( \neg locked,1),
holds( \neg locked,2),
holds(\neg opened, 2), \\
holds(opened, 3),
holds(locked, 3), goal
\; \}
\end{eqnarray*}
\begin{eqnarray*}
S_6 = \{ \; occ(close,0), occ(check,1), occ(inspect,2), \\
holds(opened,0),
holds(\neg locked,0) , \\
holds(\neg opened,1),
holds( \neg locked,1),
holds( \neg locked,2),
holds(\neg opened, 2), \\
holds(opened, 3),
holds(\neg locked, 3), goal
\; \}
\end{eqnarray*}
\begin{eqnarray*}
S_7  = \{ \; occ(close,0), occ(check,1), occ(inspect,2), \\
holds(opened,0),
holds(\neg locked,0) , \\
holds(\neg opened,1),
holds( \neg locked,1),
holds( \neg locked,2),
holds(\neg opened, 2), \\
holds(\neg opened, 3),
holds(locked, 3), goal
\; \}
\end{eqnarray*}
\begin{eqnarray*}
S_8  = \{ \; occ(close,0), occ(check,1), occ(inspect,2), \\
holds(opened,0),
holds(\neg locked,0) , \\
holds(\neg opened,1),
holds( \neg locked,1),
holds( \neg locked,2),
holds(\neg opened, 2), \\
holds(\neg opened, 3),
holds(\neg locked, 3), goal
\; \}
\end{eqnarray*}
Observe that the answer sets $S_3$ and $S_7$ are the answer sets that satisfy the goal which in turn correspond to the conditional plan trajectories that satisfy the security robot goal. The stable non deterministic model, $I_2$, is described as follows.
\[
\begin{array}{c}
I_2 = \{
\\
\\
occ(check,0), occ(flip\_lock,1), occ(inspect,2),
\\
holds(opened, 0),
\\
holds(\neg locked, 0),
\\
holds(\neg locked, 1),
\\
holds(\neg locked, 2),
\\
\\
\left\{ \begin{array}{r}  holds(opened, 1), \\ holds(\neg opened, 1) \end{array} \right \},
\left\{ \begin{array}{r} holds(opened, 2), \\ holds(\neg opened, 2) \end{array} \right \},
\\
\\
\left\{ \begin{array}{r} holds(locked, 3), \\ holds(\neg locked,3) \end{array} \right \},
\left\{ \begin{array}{r} holds(opened, 3), \\ holds(\neg opened,3) \end{array} \right \},
\\
goal
\\
\\
\}
\end{array}
\]
The set of answer sets, $S_{I_2}$, that corresponds to the stable non deterministic model, $I_2$, is listed by the following answer sets.

\begin{eqnarray*}
S_1 = \{ \; occ(check,0), occ(flip\_lock,1), occ(inspect,2), \\
holds(opened, 0),
holds(\neg locked, 0),
holds(\neg locked, 1),
holds(\neg locked, 2), \\
holds(opened, 1), holds(opened, 2), holds(locked, 3), holds(opened, 3),
goal
\; \}
\end{eqnarray*}

\begin{eqnarray*}
S_2 = \{ \; occ(check,0), occ(flip\_lock,1), occ(inspect,2), \\
holds(opened, 0),
holds(\neg locked, 0),
holds(\neg locked, 1),
holds(\neg locked, 2), \\
holds(opened, 1), holds(opened, 2), holds(locked, 3), holds(\neg opened, 3),
goal
\; \}
\end{eqnarray*}

\begin{eqnarray*}
S_3 = \{ \; occ(check,0), occ(flip\_lock,1), occ(inspect,2), \\
holds(opened, 0),
holds(\neg locked, 0),
holds(\neg locked, 1),
holds(\neg locked, 2), \\
holds(opened, 1), holds(opened, 2), holds(\neg locked, 3), holds(opened, 3),
goal
\; \}
\end{eqnarray*}

\begin{eqnarray*}
S_4 = \{ \; occ(check,0), occ(flip\_lock,1), occ(inspect,2), \\
holds(opened, 0),
holds(\neg locked, 0),
holds(\neg locked, 1),
holds(\neg locked, 2), \\
holds(opened, 1), holds(opened, 2), holds(\neg locked, 3), holds(\neg opened, 3),
goal
\; \}
\end{eqnarray*}

\begin{eqnarray*}
S_5 = \{ \; occ(check,0), occ(flip\_lock,1), occ(inspect,2), \\
holds(opened, 0),
holds(\neg locked, 0),
holds(\neg locked, 1),
holds(\neg locked, 2), \\
holds(opened, 1), holds(\neg opened, 2), holds(locked, 3), holds(opened, 3),
goal
\; \}
\end{eqnarray*}

\begin{eqnarray*}
S_6 = \{ \; occ(check,0), occ(flip\_lock,1), occ(inspect,2), \\
holds(opened, 0),
holds(\neg locked, 0),
holds(\neg locked, 1),
holds(\neg locked, 2), \\
holds(opened, 1), holds(\neg opened, 2), holds(locked, 3), holds(\neg opened, 3),
goal
\; \}
\end{eqnarray*}

\begin{eqnarray*}
S_7 = \{ \; occ(check,0), occ(flip\_lock,1), occ(inspect,2), \\
holds(opened, 0),
holds(\neg locked, 0),
holds(\neg locked, 1),
holds(\neg locked, 2), \\
holds(opened, 1), holds(\neg opened, 2), holds(\neg locked, 3), holds(opened, 3),
goal
\; \}
\end{eqnarray*}

\begin{eqnarray*}
S_8 = \{ \; occ(check,0), occ(flip\_lock,1), occ(inspect,2), \\
holds(opened, 0),
holds(\neg locked, 0),
holds(\neg locked, 1),
holds(\neg locked, 2), \\
holds(opened, 1), holds(\neg opened, 2), holds(\neg locked, 3), holds(\neg opened, 3),
goal
\; \}
\end{eqnarray*}

\begin{eqnarray*}
S_9 = \{ \; occ(check,0), occ(flip\_lock,1), occ(inspect,2), \\
holds(opened, 0),
holds(\neg locked, 0),
holds(\neg locked, 1),
holds(\neg locked, 2), \\
holds(\neg opened, 1), holds(opened, 2), holds(locked, 3), holds(opened, 3),
goal
\; \}
\end{eqnarray*}

\begin{eqnarray*}
S_{10} = \{ \; occ(check,0), occ(flip\_lock,1), occ(inspect,2), \\
holds(opened, 0),
holds(\neg locked, 0),
holds(\neg locked, 1),
holds(\neg locked, 2), \\
holds(\neg opened, 1), holds(opened, 2), holds(locked, 3), holds(\neg opened, 3),
goal
\; \}
\end{eqnarray*}

\begin{eqnarray*}
S_{11} = \{ \; occ(check,0), occ(flip\_lock,1), occ(inspect,2), \\
holds(opened, 0),
holds(\neg locked, 0),
holds(\neg locked, 1),
holds(\neg locked, 2), \\
holds(\neg opened, 1), holds(opened, 2), holds(\neg locked, 3), holds(opened, 3),
goal
\; \}
\end{eqnarray*}

\begin{eqnarray*}
S_{12} = \{ \; occ(check,0), occ(flip\_lock,1), occ(inspect,2), \\
holds(opened, 0),
holds(\neg locked, 0),
holds(\neg locked, 1),
holds(\neg locked, 2), \\
holds(\neg opened, 1), holds(opened, 2), holds(\neg locked, 3), holds(\neg opened, 3),
goal
\; \}
\end{eqnarray*}

\begin{eqnarray*}
S_{13} = \{ \; occ(check,0), occ(flip\_lock,1), occ(inspect,2), \\
holds(opened, 0),
holds(\neg locked, 0),
holds(\neg locked, 1),
holds(\neg locked, 2), \\
holds(\neg opened, 1), holds(\neg opened, 2), holds(locked, 3), holds(opened, 3),
goal
\; \}
\end{eqnarray*}

\begin{eqnarray*}
S_{14} = \{ \; occ(check,0), occ(flip\_lock,1), occ(inspect,2), \\
holds(opened, 0),
holds(\neg locked, 0),
holds(\neg locked, 1),
holds(\neg locked, 2), \\
holds(\neg opened, 1), holds(\neg opened, 2), holds(locked, 3), holds(\neg opened, 3),
goal
\; \}
\end{eqnarray*}

\begin{eqnarray*}
S_{15} = \{ \; occ(check,0), occ(flip\_lock,1), occ(inspect,2), \\
holds(opened, 0),
holds(\neg locked, 0),
holds(\neg locked, 1),
holds(\neg locked, 2), \\
holds(\neg opened, 1), holds(\neg opened, 2), holds(\neg locked, 3), holds(opened, 3),
goal
\; \}
\end{eqnarray*}

\begin{eqnarray*}
S_{16} = \{ \; occ(check,0), occ(flip\_lock,1), occ(inspect,2), \\
holds(opened, 0),
holds(\neg locked, 0),
holds(\neg locked, 1),
holds(\neg locked, 2), \\
holds(\neg opened, 1), holds(\neg opened, 2), holds(\neg locked, 3), holds(\neg opened, 3),
goal
\; \}
\end{eqnarray*}
Observe that the answer sets $S_2$, $S_6$, $S_{10}$, and $S_{14}$ are the answer sets that satisfy the goal which in turn correspond to the conditional plan trajectories that satisfy the security robot goal. The stable non deterministic model, $I_3$, is described as follows.

\[
\begin{array}{c}
I_3 = \{
\\
\\
occ(check, 0), occ(check, 1),  occ(inspect, 2),
\\
holds(opened, 0), \\
holds(\neg locked, 0), \\
holds(\neg locked, 1), \\
holds(\neg locked, 2), \\
\\
\left \{ \begin{array}{r} holds(opened, 1), \\ holds(\neg opened, 1) \end{array} \right \},
\left \{ \begin{array}{r} holds(opened, 2), \\ holds(\neg opened, 2) \end{array} \right \},
\\
\\
\left \{ \begin{array}{r} holds(opened, 3), \\ holds(\neg opened, 3) \end{array} \right \},
\left \{ \begin{array}{r} holds(locked, 3), \\ holds(\neg locked, 3) \end{array} \right \},
\\
goal
\\
\\
\}
\end{array}
\]
The set of answer sets, $S_{I_3}$, that corresponds to the stable non deterministic model, $I_3$, is listed by the following answer sets.

\begin{eqnarray*}
S_1 = \{ \; occ(check, 0), occ(check, 1),  occ(inspect, 2), \\
holds(opened, 0),
holds(\neg locked, 0),
holds(\neg locked, 1),
holds(\neg locked, 2),
\\
holds(opened, 1), holds(opened, 2), holds(opened, 3), holds(locked, 3),
goal
\; \}
\end{eqnarray*}

\begin{eqnarray*}
S_2 = \{ \; occ(check, 0), occ(check, 1),  occ(inspect, 2), \\
holds(opened, 0),
holds(\neg locked, 0),
holds(\neg locked, 1),
holds(\neg locked, 2),
\\
holds(opened, 1), holds(opened, 2), holds(opened, 3), holds(\neg locked, 3),
goal
\; \}
\end{eqnarray*}

\begin{eqnarray*}
S_3 = \{ \; occ(check, 0), occ(check, 1),  occ(inspect, 2), \\
holds(opened, 0),
holds(\neg locked, 0),
holds(\neg locked, 1),
holds(\neg locked, 2),
\\
holds(opened, 1), holds(opened, 2), holds(\neg opened, 3), holds(locked, 3),
goal
\; \}
\end{eqnarray*}

\begin{eqnarray*}
S_4 = \{ \; occ(check, 0), occ(check, 1),  occ(inspect, 2), \\
holds(opened, 0),
holds(\neg locked, 0),
holds(\neg locked, 1),
holds(\neg locked, 2),
\\
holds(opened, 1), holds(opened, 2), holds(\neg opened, 3), holds(\neg locked, 3),
goal
\; \}
\end{eqnarray*}

\begin{eqnarray*}
S_5 = \{ \; occ(check, 0), occ(check, 1),  occ(inspect, 2), \\
holds(opened, 0),
holds(\neg locked, 0),
holds(\neg locked, 1),
holds(\neg locked, 2),
\\
holds(opened, 1), holds(\neg opened, 2), holds(opened, 3), holds(locked, 3),
goal
\; \}
\end{eqnarray*}

\begin{eqnarray*}
S_6 = \{ \; occ(check, 0), occ(check, 1),  occ(inspect, 2), \\
holds(opened, 0),
holds(\neg locked, 0),
holds(\neg locked, 1),
holds(\neg locked, 2),
\\
holds(opened, 1), holds(\neg opened, 2), holds(opened, 3), holds(\neg locked, 3),
goal
\; \}
\end{eqnarray*}

\begin{eqnarray*}
S_7 = \{ \; occ(check, 0), occ(check, 1),  occ(inspect, 2), \\
holds(opened, 0),
holds(\neg locked, 0),
holds(\neg locked, 1),
holds(\neg locked, 2),
\\
holds(opened, 1), holds(\neg opened, 2), holds(\neg opened, 3), holds(locked, 3),
goal
\; \}
\end{eqnarray*}

\begin{eqnarray*}
S_8 = \{ \; occ(check, 0), occ(check, 1),  occ(inspect, 2), \\
holds(opened, 0),
holds(\neg locked, 0),
holds(\neg locked, 1),
holds(\neg locked, 2),
\\
holds(opened, 1), holds(\neg opened, 2), holds(\neg opened, 3), holds(\neg locked, 3),
goal
\; \}
\end{eqnarray*}

\begin{eqnarray*}
S_9 = \{ \; occ(check, 0), occ(check, 1),  occ(inspect, 2), \\
holds(opened, 0),
holds(\neg locked, 0),
holds(\neg locked, 1),
holds(\neg locked, 2),
\\
holds(\neg opened, 1), holds(opened, 2), holds(opened, 3), holds(locked, 3),
goal
\; \}
\end{eqnarray*}

\begin{eqnarray*}
S_{10} = \{ \; occ(check, 0), occ(check, 1),  occ(inspect, 2), \\
holds(opened, 0),
holds(\neg locked, 0),
holds(\neg locked, 1),
holds(\neg locked, 2),
\\
holds(\neg opened, 1), holds(opened, 2), holds(opened, 3), holds(\neg locked, 3),
goal
\; \}
\end{eqnarray*}

\begin{eqnarray*}
S_{11} = \{ \; occ(check, 0), occ(check, 1),  occ(inspect, 2), \\
holds(opened, 0),
holds(\neg locked, 0),
holds(\neg locked, 1),
holds(\neg locked, 2),
\\
holds(\neg opened, 1), holds(opened, 2), holds(\neg opened, 3), holds(locked, 3),
goal
\; \}
\end{eqnarray*}

\begin{eqnarray*}
S_{12} = \{ \; occ(check, 0), occ(check, 1),  occ(inspect, 2), \\
holds(opened, 0),
holds(\neg locked, 0),
holds(\neg locked, 1),
holds(\neg locked, 2),
\\
holds(\neg opened, 1), holds(opened, 2), holds(\neg opened, 3), holds(\neg locked, 3),
goal
\; \}
\end{eqnarray*}

\begin{eqnarray*}
S_{13} = \{ \; occ(check, 0), occ(check, 1),  occ(inspect, 2), \\
holds(opened, 0),
holds(\neg locked, 0),
holds(\neg locked, 1),
holds(\neg locked, 2),
\\
holds(\neg opened, 1), holds(\neg opened, 2), holds(opened, 3), holds(locked, 3),
goal
\; \}
\end{eqnarray*}

\begin{eqnarray*}
S_{14} = \{ \; occ(check, 0), occ(check, 1),  occ(inspect, 2), \\
holds(opened, 0),
holds(\neg locked, 0),
holds(\neg locked, 1),
holds(\neg locked, 2),
\\
holds(\neg opened, 1), holds(\neg opened, 2), holds(opened, 3), holds(\neg locked, 3),
goal
\; \}
\end{eqnarray*}

\begin{eqnarray*}
S_{15} = \{ \; occ(check, 0), occ(check, 1),  occ(inspect, 2), \\
holds(opened, 0),
holds(\neg locked, 0),
holds(\neg locked, 1),
holds(\neg locked, 2),
\\
holds(\neg opened, 1), holds(\neg opened, 2), holds(\neg opened, 3), holds(locked, 3),
goal
\; \}
\end{eqnarray*}

\begin{eqnarray*}
S_{16} = \{ \; occ(check, 0), occ(check, 1),  occ(inspect, 2), \\
holds(opened, 0),
holds(\neg locked, 0),
holds(\neg locked, 1),
holds(\neg locked, 2),
\\
holds(\neg opened, 1), holds(\neg opened, 2), holds(\neg opened, 3), holds(\neg locked, 3),
goal
\; \}
\end{eqnarray*}
Observe that the answer sets $S_3$, $S_7$, $S_{11}$, and $S_{15}$ are the answer sets that satisfy the goal which in turn correspond to the conditional plan trajectories that satisfy the security robot goal.

\section{The Fixpoint Operator of Normal Non Deterministic Logic Programs}

In this section we define the fixpoint operator of normal non deterministic logic programs. We show that this operator is non-monotonic as well as every stable non deterministic model is a minimal fixpoint of this fixpoint operator with respect to the set inclusion order $\subseteq$. The following definition formulates the notion of the fixpoint operator, denoted by $T^\prime_\Pi$, associated to a normal non deterministic logic program $\Pi$.

\begin {definition} Let $\Pi$ be a ground normal non deterministic logic program and $I$ be a non deterministic interpretation. The immediate consequence operator $T^\prime_\Pi$ of $\Pi$ is the mapping $T^\prime_\Pi : 2^{\cal N_L} \rightarrow 2^{\cal N_L}$ which is defined as
\[
\begin{array}{lcl}
T^\prime_\Pi(I)= \{ \qquad \{ A_i \}_{i = 1}^{n} \; | \;  \{ A_i \}_{i = 1}^{n} & \leftarrow &\{ B_{i_1} \}_{i_1 = 1}^{n_1} , \ldots, \{ B_{i_l} \}_{i_l = 1}^{n_l}, \\
&& not \: \{ B_{i_{l+1}} \}_{i_{l+1} = 1}^{n_{l+1}} , \ldots, not \: \{ B_{i_m} \}_{i_m = 1}^{n_m} \; \in \Pi
\end{array}
\]
and $\forall (1 \leq j \leq l)$ $\{ B_{i_j} \}_{i_j = 1}^{n_j} \in I$ and $\forall (l+1 \leq k \leq m$) $\{ B_{i_k} \}_{i_k = 1}^{n_k} \notin I \; \; \}$.
\end {definition}
It is easy to verify that $T^\prime_\Pi$ extends $T_\Pi$ to handle non deterministic logic rules with non-monotonic negation. The following theorem establishes the relationship between $T^\prime_\Pi$ and $T_\Pi$ operators.

\begin{theorem} Let $\Pi$ be a normal non deterministic logic program such that for every normal non deterministic logic rule in $\Pi$, $m = 0$. Then $T^\prime_\Pi = T_\Pi$.
\label{thm:Tp'-and-Tp}
\end{theorem}
The operator $T^\prime_\Pi$ is not monotonic with respect to the set-inclusion order $\subseteq$. To show that the $T^\prime_\Pi$ operator is not monotonic consider the following normal non deterministic logic program.

\begin{example} Consider the normal non deterministic logic programs, $\Pi$, that consists of the normal non deterministic logic rule
\[
\left\{
\begin{array}{r}
a_1, \\
a_2
\end{array}
\right\}
\leftarrow
not\;
\left\{
\begin{array}{r}
b_1, \\
b_2
\end{array}
\right\}
\]
Let $I_1 = \emptyset$ and
$
I_2 =
\left\{ \left\{
\begin{array}{r}
b_1, \\
b_2
\end{array}
\right\} \right\}
$
be two non deterministic interpretations for $\Pi$. It is clear that $I_1 \subseteq I_2$. However,
$
T^\prime_\Pi(I_1)=
\left\{ \left\{
\begin{array}{r}
a_1, \\
a_2
\end{array}
\right\} \right\}
$
and
$
T^\prime_\Pi(I_2) = \emptyset
$.
This implies that, $T^\prime_\Pi(I_1) \nsubseteq T^\prime_\Pi(I_2)$.
\end{example}
The following results establish the relationship between the fixpoint operator, $T^\prime_\Pi$, and the stable non deterministic models of normal non deterministic logic programs.

\begin{lemma} Let $\Pi$ be a normal non deterministic logic program and $I$ be a stable non deterministic model for $\Pi$. Then $T^\prime_\Pi(I) = I$, i.e., $I$ is a fixpoint of $T^\prime_\Pi$.
\label{lemma:rel_smodels}
\end{lemma}

\begin{theorem} Let $\Pi$ be a normal non deterministic logic program and $I$ be a stable non deterministic model for $\Pi$. Then $I$ is a minimal fixpoint of $T'_\Pi$.
\label{thm:rel_smodels}
\end{theorem}
It is worth noting that not every minimal fixpoint of the immediate consequence operator, $T^\prime_\Pi$, of a normal non deterministic logic program, $\Pi$, is a stable non deterministic model for $\Pi$. To show this consider the following normal non deterministic logic programs.

\begin{example} Let $\Pi$ be a normal non deterministic logic program that consists of the following normal non deterministic logic rules
\[
\begin{array}{lcl}
\left\{
\begin{array}{r}
a_1, \\
a_2
\end{array}
\right\}
& \leftarrow &
not\;
\left\{
\begin{array}{r}
a_1, \\
a_2
\end{array}
\right\}
\\
\\
\left\{
\begin{array}{r}
a_1, \\
a_2
\end{array}
\right\}
& \leftarrow &
\left\{
\begin{array}{r}
b_1, \\
b_2
\end{array}
\right\}
\end{array}
\]
The non deterministic interpretation
$
I =
\left\{ \left\{
\begin{array}{r}
a_1, \\
a_2
\end{array}
\right\}
,
\left\{
\begin{array}{r}
b_1, \\
b_2
\end{array}
\right\} \right\}
$
is a minimal fixpoint of the operator $T^\prime_\Pi$. However, the non deterministic reduct, $\Pi^I$, of $\Pi$ consists of the normal non deterministic logic rule
\[
\left\{
\begin{array}{r}
a_1, \\
a_2
\end{array}
\right\}
 \leftarrow
\left\{
\begin{array}{r}
b_1, \\
b_2
\end{array}
\right\}
\]
where $lfp(T_{\Pi^I}) = \emptyset$ which is not equal to $I$. Consequently, $I$ is not a stable non deterministic model for $\Pi$.
\end{example}

\section{Relationship Between the Stable Non Deterministic Models and the Deterministic Stable Models}

In this section we establish the relationship between the stable non deterministic model semantics of normal non deterministic logic programs and the deterministic stable model semantics of deterministic normal logic programs introduced in \cite{Gelfond_A}. The stable model semantics of normal logic programs presented in \cite{Gelfond_A} is deterministic in the sense that the normal logic programs considered in \cite{Gelfond_A} and the stable models defined for the normal logic programs in \cite{Gelfond_A} allow a single atom (deterministic atom) as the building block for both the normal logic programs and their stable models. However, normal non deterministic logic programs and their stable non deterministic model semantics allow atoms to be non deterministic for which a non deterministic atom is represented by a set of atoms of the form $\{A_i\}_{i = 1}^n$. This implies that any deterministic atom, $A$, representation in the language of deterministic normal logic programs described in \cite{Gelfond_A} can be represented as a non deterministic atom of the form, $\{A\}$, in the language of normal non deterministic logic programs. And hence, it can be shown that the syntax and the stable non deterministic model semantics of normal non deterministic logic programs naturally subsumes the syntax and the deterministic stable model semantics of the deterministic normal logic programs described in \cite{Gelfond_A}.

Any deterministic normal logic program, $\Upsilon$, can be represented as a normal non deterministic logic program, $\Pi$, where each deterministic normal logic rule of the form
\[
A \leftarrow B_1, \ldots, B_l, not \; B_{l+1},\ldots, not \; B_m \in \Upsilon
\]
can be represented as a normal non deterministic logic rule of the form
\[
\{A\} \leftarrow \{B_1\}, \ldots, \{B_l\}, not \; \{B_{l+1}\},\ldots, not \; \{B_m\} \in \Pi
\]
where $A, B_1, \ldots, B_l, B_{l+1},\ldots, B_m$ are atoms.
\\
\\
Observe that a Herbrand interpretation of a deterministic normal logic program, $\Upsilon$, is a subset of the Herbrand base ${\cal B_L}$, which is a set element in the non deterministic base ${\cal N_L}$. This means that a Herbrand interpretation, $I$, of a deterministic normal logic program is a set element in ${\cal N_L}$, i.e., $I \in {\cal N_L}$.

\begin{theorem}
Let $\Upsilon$ be a deterministic normal logic program and $\Pi$ be the normal non deterministic logic program representation of $\Upsilon$. Then $I$ is a deterministic stable model for $\Upsilon$ if and only if $\forall \: A \in I, \: \{ A \} \in J$ is a stable non deterministic model for $\Pi$.
\label{thm:non-deterministic-stable-to-deterministic-stable}
\end{theorem}
The following example shows how the stable non deterministic model semantics of normal non deterministic logic programs captures the deterministic stable model semantics of deterministic normal logic programs.

\begin{example} Consider the following deterministic normal logic program $\Upsilon$.

\[
\begin{array}{lcl}
a & \leftarrow & not \; b \\
b & \leftarrow & not \; a
\end{array}
\]
The deterministic stable models of the deterministic normal logic program $\Upsilon$ are $\{a\}$ and $\{b\}$. On the other hand, the normal non deterministic logic program representation of $\Upsilon$, denoted by $\Pi$, is given by
\[
\begin{array}{lcl}
\{\; a \;\} & \leftarrow  & not \; \{\; b \;\} \\
\{\; b \; \} & \leftarrow & not \; \{\; a \;\}
\end{array}
\]
Thus, it is easy to verify that the only two stable non deterministic models of $\Pi$ are $\{\; \{a\} \; \}$ and $\{\;\{ b\}\;\}$.
\end{example}

\section{Well-Founded Non Deterministic Model Semantics}

In this section we introduce the well-founded non deterministic model semantics for normal non deterministic logic program. Our main motivation is to provide an easy to compute semantics for normal non deterministic logic programs and to investigate its relationship to the stable non deterministic model semantics introduced earlier. This is because providing an easy to compute non deterministic models for normal non deterministic logic programs is an important issue in many applications.

The definition of the well-founded non deterministic model semantics is based on the notions of unfounded non deterministic set and the notion of immediate consequence operator of normal non deterministic logic program. Unfounded non deterministic set corresponds to the set of negative conclusions and the immediate consequence operator is used to derive the set of positive conclusions from the normal non deterministic logic programs. The well-founded non deterministic model produced from the well-founded non deterministic model semantics is defined inductively by combining the set of negative conclusions with the set of positive conclusions.

We show that the relationship between the well-founded non deterministic model semantics and the stable non deterministic model semantics for normal non deterministic logic programs preserves the relationship between the
deterministic well-founded semantics and the deterministic stable model semantics for deterministic normal logic programs. With a simple translation of deterministic normal logic programs into normal non deterministic logic programs, we show that the well-founded non deterministic model semantics for normal non deterministic logic programs naturally subsumes the deterministic well-founded semantics for deterministic normal logic programs, presented in \cite{Gelder},  as well as it reduces to the semantics of non deterministic logic programs in the absence of negation.

The well-founded non deterministic model semantics is another approach to provide meaning to normal non deterministic logic programs. In the well-founded non deterministic model semantics, if $I$ is a well-founded non deterministic model for a normal non deterministic logic program, $\Pi$, then for any non deterministic atom $\{A_i\}_{i = 1}^n \in {\cal N_L}$ either $\{A_i\}_{i = 1}^n$ is true in $I$ if $\{A_i\}_{i = 1}^n \in I$ or $\{A_i\}_{i = 1}^n$ is false in $I$ if $not \; \{A_i\}_{i = 1}^n \in I$, or $\{A_i\}_{i = 1}^n$ is \emph {undefined} in $I$ if neither $\{A_i\}_{i = 1}^n$ nor $not\; \{A_i\}_{i = 1}^n$ belongs to $I$. Unlike the stable non deterministic model semantics, the idea of the well-founded non deterministic model semantics is to have only one non deterministic model as the meaning of a normal non deterministic logic program. The semantics is defined as follows.

\begin{definition}
Let $\Pi$ be a normal non deterministic logic program and ${\cal N_L}$ be the non deterministic base. A partial non deterministic interpretation, $I$, for $\Pi$ is a subset from $ \{ \: \{A_i \}_{i = 1}^n \: | \: \{A_i \}_{i = 1}^n \in {\cal N_L} \: \} \cup \{ \: not \; \{A_i \}_{i = 1}^n \: | \: \{A_i \}_{i = 1}^n \in {\cal N_L} \: \}$ such that $ \forall \{A_i \}_{i = 1}^n \in {\cal N_L}, \{ \: \{A_i \}_{i = 1}^n , not \; \{A_i \}_{i = 1}^n \: \} \nsubseteq I$. We say $I$ is a total non deterministic interpretation for $\Pi$ if $ \forall \{A_i \}_{i = 1}^n \in {\cal N_L}$, either $\{A_i \}_{i = 1}^n$ or $not\; \{A_i \}_{i = 1}^n$ belongs to $I$.
\end {definition}

\begin{definition} Let $\Pi$ be a normal non deterministic logic program, ${\cal I}_\Pi$ be the set of all partial non deterministic interpretations of $\Pi$, and $I_1$ and $I_2$ be two partial non deterministic interpretations in ${\cal I}_\Pi$. Then the set inclusion $\subseteq$ is the natural partial order on the set of partial non deterministic interpretations ${\cal I}_\Pi$ of $\Pi$. In particular, the join operation of $I_1$ and $I_2$ is the union, $I_1 \cup I_2 $, of $I_1$ and $I_2$ and the meet operation of $I_1$ and $I_2$ is the intersection, $I_1 \cap I_2 $,  of $I_1$ and $I_2$.
\end {definition}

\begin{definition} Let $X$ be a set of partial non deterministic interpretations. Then:
\[
\bigcup_{S \in X} \; S
\]
is the join of all partial non deterministic interpretations in $X$, and
\[
\bigcap_{S \in X} \; S
\]
is the meet of all partial non deterministic interpretations in $X$.
\end {definition}

\begin{lemma} Let $I_1$ and $I_2$ be two partial or total non deterministic interpretations. If $I_1 \subseteq I_2$ and $I_2 \subseteq I_1$, then $I_1 = I_2$.
\label{lemma:interpretations-equality}
\end{lemma}
The set of all partial non deterministic interpretation, ${\cal I}_\Pi$, of a normal non deterministic logic program, $\Pi$, along with the partial order, $\subseteq$, $\langle {\cal I}_\Pi, \subseteq \rangle$, do \emph{not} form a lattice. Indeed if $I_1, I_2 \in {\cal I}_\Pi$ are partial or total non deterministic interpretations, then $I_1 \cup I_2$ may not exist. For example, consider ${\cal B_L} = \{a,b,c,d,e,f\}$ and $I_1$, $I_2$ be two partial non deterministic interpretations where
\[
I_1 = \{\; \{d)\}, \{e\}, \{f\}, not \; \{a\}, not \; \{b\}, not \; \{c\} \; \}.
\]

\[
I_2 = \{ \; \{c\}, \{d\}, \{e\}, \{f\}, not\; \{a\}, not\; \{b\} \; \}
\]
Therefore, the upper bound (the join) of $I_1$ and $I_2$ is given by
\[
I_1 \cup I_2 = \{ \; \{c\}, \{d\}, \{e\}, \{f\}, not\; \{a\}, not\; \{b\}, not \; \{c\} \; \}
\]
which does not exist because $I_1 \cup I_2$ is not a partial or total non deterministic interpretation (by the definition) since $ \{ \; \{c\}, not \; \{c\} \; \} \subseteq I_1 \cup I_2$.

However, $\langle {\cal I}_\Pi, \subseteq \rangle$ form \emph{a complete partial order (cpo)}, that is a partial order in which the limits of growing chains exit. This is sufficient for building well-founded non deterministic models inductively. The bottom element of the partially ordered set ${\cal I}_\Pi$ under $\subseteq$, $\langle {\cal I}_\Pi, \subseteq \rangle$, is the empty set, $\emptyset$, and the top element in $\langle {\cal I}_\Pi, \subseteq \rangle$ is a total non deterministic interpretation, $I$, such that $\forall \: \{A_i\}_{i = 1}^n  \in {\cal N_L}$, either $\{A_i\}_{i = 1}^n $ or $not \; \{A_i\}_{i = 1}^n $ belongs to $I$. The following results show that $\langle {\cal I}_\Pi, \subseteq \rangle$ is \emph{a complete partial order (cpo)}.

\begin{lemma} Let $\Pi$ be a normal non deterministic logic program, ${\cal I}_\Pi$ be the set of all partial non deterministic interpretation of $\Pi$, and $I_1, I_2 \in {\cal I}_\Pi$. If $lub\{I_1, I_2\}$ and $glb\{I_1, I_2\}$ exist, then $lub\{I_1, I_2\} = I_1 \cup I_2$ and $glb\{I_1, I_2\} = I_1 \cap I_2$.
\label{lemma:lub-and-glb-for-partial}
\end{lemma}

\begin{lemma} The set of all partial non deterministic interpretations, ${\cal I}_\Pi$, along with the partial order $\subseteq$ form a complete partial order.
\label{lemma:cpo-for-partial}
\end{lemma}
The definition of satisfaction in the well-founded non deterministic model semantics is similar to the definition of satisfaction in the stable non deterministic model semantics. Except that, in the well-founded non deterministic model semantics, a negative non deterministic atom $not\; \{A_i \}_{i = 1}^n$ is satisfied by a partial or total non deterministic interpretation $I$ if $not\; \{A_i \}_{i = 1}^n$ belongs to $I$.

\begin {definition} Let $\Pi$ be a ground normal non deterministic logic program and $I$ be a partial or total non deterministic interpretation. Then the notion of satisfaction, denoted by $\models$, of a non deterministic atom and a normal non deterministic logic rule, by $I$, is defined as follows:

\begin{itemize}

\item $I \models \{ B_{i_j} \}_{i_j = 1}^{n_j}$ iff $\{ B_{i_j} \}_{i_j = 1}^{n_j} \in I$.

\item $I \models not \; \{ B_{i_k} \}_{i_k = 1}^{n_k}$ iff $not \; \{ B_{i_k} \}_{i_k = 1}^{n_k} \in I$.

\item $I \models \{ B_{i_1} \}_{i_1 = 1}^{n_1} , \ldots, \{ B_{i_l} \}_{i_l = 1}^{n_l}, not \: \{ B_{i_{l+1}} \}_{i_{l+1} = 1}^{n_{l+1}} , \ldots, not \: \{ B_{i_m} \}_{i_m = 1}^{n_m}$
    iff $\forall (1 \leq j \leq l)$ $I \models \{ B_{i_j} \}_{i_j = 1}^{n_j}$ and $\forall (l+1 \leq k \leq m$) $I \models not \; \{ B_{i_k} \}_{i_k = 1}^{n_k}$

\item $I \models \{ A_i \}_{i = 1}^{n} \leftarrow \{ B_{i_1} \}_{i_1 = 1}^{n_1} , \ldots, \{ B_{i_l} \}_{i_l = 1}^{n_l}, not \: \{ B_{i_{l+1}} \}_{i_{l+1} = 1}^{n_{l+1}} , \ldots, not \: \{ B_{i_m} \}_{i_m = 1}^{n_m}$
    iff $I \models \{ A_i \}_{i = 1}^{n}$ whenever

    $I \models \{ B_{i_1} \}_{i_1 = 1}^{n_1} , \ldots, \{ B_{i_l} \}_{i_l = 1}^{n_l}, not \: \{ B_{i_{l+1}} \}_{i_{l+1} = 1}^{n_{l+1}} , \ldots, not \: \{ B_{i_m} \}_{i_m = 1}^{n_m}$

    or $I$ does not satisfy $\{ B_{i_1} \}_{i_1 = 1}^{n_1} , \ldots, \{ B_{i_l} \}_{i_l = 1}^{n_l}, not \: \{ B_{i_{l+1}} \}_{i_{l+1} = 1}^{n_{l+1}} , \ldots, not \: \{ B_{i_m} \}_{i_m = 1}^{n_m}$.
\end{itemize}
\end {definition}
The following definition describes partial and total non deterministic models in the well-founded non deterministic model semantics.

\begin{definition} Let $\Pi$ be a normal non deterministic logic program. A total non deterministic interpretation, $I$, is a total non deterministic model for $\Pi$ if $I$ satisfies every normal non deterministic logic rule in $\Pi$. A partial non deterministic interpretation, $I$, is a partial non deterministic model for $\Pi$ if $I$ can be extended to a total non deterministic model for $\Pi$.
\end {definition}
A partial non deterministic model for a normal non deterministic logic program, $\Pi$, is a non deterministic model, for $\Pi$, for which some normal non deterministic logic rules in $\Pi$ may not be satisfied.

The following definition formulates the notion of \emph{unfounded non deterministic set}. The idea is that the unfounded non deterministic set of a normal non deterministic logic program, $\Pi$, corresponds to the set of negative conclusions of the normal non deterministic logic program $\Pi$. Therefore, if a non deterministic atom $\{ A_i \}_{i = 1}^{n}$ is in an unfounded non deterministic set of $\Pi$, then $\{ A_i \}_{i = 1}^{n}$ should asserted to be false in the partial or total non deterministic model of $\Pi$.

\begin {definition} Let $\Pi$ be a ground normal non deterministic logic program, ${\cal N_L}$ be the non deterministic base, $\xi \subseteq {\cal N_L}$, and $I$ be a partial non deterministic interpretation. $\xi$ is said to be unfounded non deterministic set of $\Pi$ w.r.t. $I$ if for each $\{A_i \}_{i = 1}^n \in \xi$, we have
$\{A_i \}_{i = 1}^n$ does not appear as a head of any normal non deterministic logic rule, $r$, in $\Pi$. Or for each normal non deterministic logic rule, $r \in \Pi$, with a head $\{A_i \}_{i = 1}^n$, we have at least one of the following conditions holds:
\begin{enumerate}

\item Some non deterministic atom, $\{B_i \}_{i = 1}^n$, or the negation of a non deterministic atom, $not \; \{B_i \}_{i = 1}^n$, in the body of $r$ is false w.r.t. $I$.

\item Some non deterministic atom, $\{B_i \}_{i = 1}^n$, in the body of $r$ belongs to $\xi$.
\end{enumerate}
\label{def:unfounded}
\end {definition}

\begin{definition} Let $\Pi$ be a ground normal non deterministic logic program and $I$ be a non deterministic interpretation. The greatest unfounded non deterministic set, $U_\Pi(I)$, of $\Pi$ w.r.t. $I$ is the union of
all unfounded non deterministic sets of $\Pi$ w.r.t. $I$.
\end {definition}
The immediate consequence operator, $T_\Pi(I)$, in the well-founded non deterministic model semantics, defined below, is a one step deduction with respect to the partial non deterministic interpretation $I$.

\begin {definition} Let $\Pi$ be a ground normal non deterministic logic program and $I$ be a partial non deterministic interpretation. The immediate consequence operator, $T_\Pi(I)$, of $\Pi$ is defined as follows.

\[
\begin{array}{lcl}
T_\Pi(I)= \{ \qquad \{ A_i \}_{i = 1}^{n} \; | \;  \{ A_i \}_{i = 1}^{n} & \leftarrow &\{ B_{i_1} \}_{i_1 = 1}^{n_1} , \ldots, \{ B_{i_l} \}_{i_l = 1}^{n_l}, \\
&& not \: \{ B_{i_{l+1}} \}_{i_{l+1} = 1}^{n_{l+1}} , \ldots, not \: \{ B_{i_m} \}_{i_m = 1}^{n_m} \; \in \Pi
\end{array}
\]
and $\forall (1 \leq j \leq l)$ $\{ B_{i_j} \}_{i_j = 1}^{n_j} \in I$ and $\forall (l+1 \leq k \leq m$) $not \; \{ B_{i_k} \}_{i_k = 1}^{n_k} \in I \; \; \}$.
\label{def:Tp_wf}
\end {definition}
The well-founded non deterministic operator, $W_\Pi$, is defined in terms of the greatest unfounded non deterministic set, $U_\Pi$, and the immediate consequence operator, $T_\Pi$, to combine the set of negative conclusions with the set of positive conclusions derived from the normal non deterministic logic programs, $\Pi$. The definition of $W_\Pi$ is determined by combing both operators $U_\Pi$ and $T_\Pi$ as follows.

\begin{definition} Let $\Pi$ be a normal non deterministic logic program, $I$ be a non deterministic interpretation, $T_\Pi$ be the immediate consequence operator of $\Pi$, and $U_\Pi(I)$ be the greatest unfounded non deterministic set of $\Pi$ w.r.t. $I$. The well-founded partial non deterministic interpretation of $\Pi$ w.r.t. $I$ is given by
\[
W_\Pi(I) = T_\Pi(I) \cup not \; U_\Pi(I).
\]
where $W_\Pi$ is well-founded non deterministic operator of $\Pi$.
\label{def:W-P}
\end{definition}

The following results assert the monotonicity of the well-founded non deterministic operator, the greatest unfounded non deterministic set, and the immediate consequence operator in the well-founded non deterministic model semantics. In addition we show that the successive application of the well-founded non deterministic operator produces monotonic sequence of well-founded partial non deterministic interpretations. Finally, we give the definition of the well-founded partial non deterministic model in the well-founded non deterministic model semantics.

\begin{lemma}
The operators $W_\Pi, T_\Pi$, and $U_\Pi$ are monotonic with respect to $\subseteq$.
\label{lemma:Wp-Tp-Up-monotonic}
\end {lemma}

\begin{definition} Let $\Pi$ be a normal non deterministic logic program. The well-founded partial non deterministic interpretations, for $\Pi$, $I_\alpha$ and $I_\omega$ are defined inductively as follows:

\begin{enumerate}

\item $I_0 = \emptyset$.

\item $I_\alpha = W_\Pi(I_{\alpha- 1})$ where $\alpha$ is the successor ordinal of $(\alpha - 1)$.

\item $I_\omega = lub \: \{I_\alpha  \; | \; \alpha < \omega \}$ where $\omega$ is a limit ordinal.


\end {enumerate}
\label{def:part_inter}
\end {definition}

\begin{lemma} $I_0, I_1, I_2, \ldots$, as defined in Definition (\ref{def:part_inter}), is a monotonic sequence
of well-founded partial non deterministic interpretations.
\label{lemma:I0-I1-I2-monotonic}
\end{lemma}

\begin {lemma} $I_0, I_1, I_2, \ldots$, as defined in Definition (\ref{def:part_inter}) is a chain in $\langle {\cal I}_\Pi, \subseteq \rangle$ whose upper bound is $I_\omega$. In addition, $I_\omega$ is the least upper bound of $I_0, I_1, I_2, \ldots$.
\label{lemma:I0-I1-I2-chain}
\end {lemma}

\begin {definition} Let $\Pi$ be a normal non deterministic logic program. The well-founded partial non deterministic model of $\Pi$ is $I_\omega$.
\end {definition}
The well-founded non deterministic model semantics of normal non deterministic logic programs is developed to provide a single common solution tree for non deterministic real-world applications whose solution would be described by one or multiple trees. This means that the well-founded partial or total non deterministic model of a normal non deterministic logic program representation of a non deterministic problem, represents the single common solution tree of the represented problem. Therefore, to construct the solution tree represented in the well-founded partial or total non deterministic model of the normal non deterministic logic program representation of the non deterministic problem we introduce the following definition.

\begin{definition}
Let \[I =
\{\;
\{ A_{i_1} \}_{i_1 = 1}^{n_1}, \ldots, \{ A_{i_l} \}_{i_l = 1}^{n_l}, not \; \{ A_{i_{l+1}} \}_{i_{l+1} = 1}^{n_{l+1}}, \ldots, not \; \{ A_{i_m} \}_{i_m = 1}^{n_m}
\;\}
\]
be the well-founded partial or total non deterministic model of the normal non deterministic logic program $\Pi$. Let $X_j$, for $1 \leq j \leq m$, be a variable ranging over the elements, $\{A_{i_j} \}_{i_j = 1}^{n_j}$, appearing in $I$. Then, the set of answer sets, $S_I$, corresponding to $I$ is the set of all {\em minimal} sets formed from the elements of $I$ such that
\[
S_I = \{ \;  \{X_1, \ldots, X_l, not \; X_{l+1}, \ldots, not \; X_m \} \:|\: \forall X_1 \; \forall X_2 \; \ldots \forall X_m \; \}.
\]
\end{definition}
The {\em set of answer sets} represents the solution tree of the represented non deterministic problem by a normal non deterministic logic program, $\Pi$, and corresponds to the well-founded partial or total non deterministic model of $\Pi$, where every {\em answer set} in the set of answer sets corresponds to a {\em branch} in the solution tree of the non deterministic problem.

Observe that an answer set, $S$, in the set of answer sets, $S_I$, that corresponds to the well-founded partial or total non deterministic model, $I$, of a normal non deterministic logic program, $\Pi$, is a subset of $({\cal B_L} \cup not \; {\cal B_L})$. Intuitively, the meaning of an answer set, $S$, in the well-founded non deterministic model semantics, is that if an atom, $A$, belongs to $S$ then we say that $A$ is true with respect to $S$, and if the negation of an atom, $not \; A$, belongs to $S$ then we say that $A$ is false with respect to $S$, however, if neither an atom, $A$, or the negation of an atom, $not \; A$, belongs to $S$ then we say that $A$ is {\em undefined} in $S$.

\begin {example} Consider the following normal non deterministic logic program, $\Pi$, that consists of the following normal non deterministic logic rules.
\[
\begin{array}{lcl}
\left\{
\begin{array}{r}
c_1, \\
c_2
\end{array}
\right\}
& \leftarrow &
\\
\\
\left\{
\begin{array}{r}
a_1, \\
a_2
\end{array}
\right\}
&  \leftarrow &
not \;
\left\{
\begin{array}{r}
b_1, \\
b_2
\end{array}
\right\}
\\
\\
\left\{
\begin{array}{r}
b_1, \\
b_2
\end{array}
\right\}
&  \leftarrow &
not \;
\left\{
\begin{array}{r}
c_1, \\
c_2
\end{array}
\right\}
\end{array}
\]
This normal non deterministic logic program, $\Pi$, has a well-founded total non deterministic model which is
$
\left\{ \; \left\{
\begin{array}{r}
a_1, \\
a_2
\end{array}
\right\}
, \; not \;
\left\{
\begin{array}{r}
b_1, \\
b_2
\end{array}
\right\}
,
\left\{
\begin{array}{r}
c_1, \\
c_2
\end{array}
\right\} \; \right\}
$.
Starting from the initial non deterministic interpretation, $I_0 = \emptyset$, then $U_\Pi(I_0) = \emptyset$
and
\[
T_\Pi(I_0)= \left\{ \; \left\{
\begin{array}{r}
c_1, \\
c_2
\end{array}
\right\} \;
\right\}.
\]
Thus,
\[
I_1 = W_\Pi(I_0) =
\left\{ \;
\left\{
\begin{array}{r}
c_1, \\
c_2
\end{array}
\right\}
\right\}.
\]
Furthermore,
\\
\\
$U_\Pi(I_1) =
\left\{ \; \left\{
\begin{array}{r}
b_1, \\
b_2
\end{array}
\right\}
\;
\right\}
$
and $T_\Pi(I_1)  =
\left\{ \; \left\{
\begin{array}{r}
c_1, \\
c_2
\end{array}
\right\} \; \right\}
$.
Hence,
\[
I_2 = W_\Pi(I_1) =
\left\{ \;
\; not \;
\left\{
\begin{array}{r}
b_1, \\
b_2
\end{array}
\right\}
,
\left\{
\begin{array}{r}
c_1, \\
c_2
\end{array}
\right\} \; \right\}.
\]
In addition,  $U_\Pi(I_2) =
\left\{ \; \left\{
\begin{array}{r}
b_1, \\
b_2
\end{array}
\right\}
\;
\right\}
$
and $T_\Pi(I_2)  =
\left\{ \;
\left\{
\begin{array}{r}
a_1, \\
a_2
\end{array}
\right\}
,
\left\{
\begin{array}{r}
c_1, \\
c_2
\end{array}
\right\} \; \right\}
$. Therefore,
\[
I_3 = W_\Pi(I_2) =
\left\{ \;
\left\{
\begin{array}{r}
a_1, \\
a_2
\end{array}
\right\}
,
\; not \;
\left\{
\begin{array}{r}
b_1, \\
b_2
\end{array}
\right\}
,
\left\{
\begin{array}{r}
c_1, \\
c_2
\end{array}
\right\} \; \right\}
\]
which is the least upper bound of $I_0$, $I_1$, $I_2$, and $I_3$. Consequently, $I_3$ is the well-founded total non deterministic model for $\Pi$.

The set of answer sets, $S_{I_3}$, corresponding to the well-founded total non deterministic model, $I_3$, of $\Pi$ consists of the following answer sets:
\[
\begin{array}{c}
S_1 = \{\;  a_1, \: not\; b_1, \: c_1  \;\} \\
S_2 = \{\;  a_1, \: not\; b_1, \: c_2  \;\} \\
S_3 = \{\;  a_1, \: not\; b_2, \: c_1  \;\} \\
S_4 = \{\;  a_1, \: not\; b_2, \: c_2  \;\} \\
S_5 = \{\;  a_2, \: not\; b_1, \: c_1  \;\} \\
S_6 = \{\;  a_2, \: not\; b_1, \: c_2  \;\} \\
S_7 = \{\;  a_2, \: not\; b_2, \: c_1  \;\} \\
S_8 = \{\;  a_2, \: not\; b_2, \: c_2  \;\}
\end{array}
\]

\end{example}

\begin{example} Consider the following normal non deterministic logic program, $\Pi$, that consists of the normal non deterministic logic rules
\[
\begin{array}{lcl}
\left\{
\begin{array}{r}
a_1, \\
a_2
\end{array}
\right\}
&  \leftarrow &
not \;
\left\{
\begin{array}{r}
b_1, \\
b_2
\end{array}
\right\}
\\
\\
\left\{
\begin{array}{r}
b_1, \\
b_2
\end{array}
\right\}
&  \leftarrow &
not \;
\left\{
\begin{array}{r}
a_1, \\
a_2
\end{array}
\right\}
\end{array}
\]
The well-founded partial non deterministic model of the normal non deterministic logic program, $\Pi$, is the empty set $\emptyset$. This is because $I_\omega = W_\Pi(\emptyset) = \emptyset$, since $T_\Pi(\emptyset) = \emptyset$ and $U_\Pi(\emptyset) = \emptyset$.
\label{ex:a-not-b}
\end{example}

\begin{example} Consider the following normal non deterministic logic program, $\Pi$, that consists of the normal non deterministic logic rules
\[
\begin{array}{lcl}
\left\{
\begin{array}{r}
a_1, \\
a_2
\end{array}
\right\}
&  \leftarrow &
not \;
\left\{
\begin{array}{r}
b_1, \\
b_2
\end{array}
\right\}
\\
\\
\left\{
\begin{array}{r}
b_1, \\
b_2
\end{array}
\right\}
&  \leftarrow &
not \;
\left\{
\begin{array}{r}
a_1, \\
a_2
\end{array}
\right\}
\\
\\
\left\{
\begin{array}{r}
c_1, \\
c_2
\end{array}
\right\}
&  \leftarrow &
\left\{
\begin{array}{r}
a_1, \\
a_2
\end{array}
\right\}
\\
\\
\left\{
\begin{array}{r}
c_1, \\
c_2
\end{array}
\right\}
&  \leftarrow &
\left\{
\begin{array}{r}
b_1, \\
b_2
\end{array}
\right\}
\end{array}
\]
The well-founded partial non deterministic model of the normal non deterministic logic program, $\Pi$, is the empty set $\emptyset$. This is because $I_\omega = W_\Pi(\emptyset) = \emptyset$, since $T_\Pi(\emptyset) = \emptyset$ and $U_\Pi(\emptyset) = \emptyset$.
\end{example}

\begin{example} Consider the following normal non deterministic logic program, $\Pi$, that consists of the normal non deterministic logic rule
\[
\begin{array}{lcl}
\left\{
\begin{array}{r}
a_1, \\
a_2
\end{array}
\right\}
&  \leftarrow &
not \;
\left\{
\begin{array}{r}
b_1, \\
b_2
\end{array}
\right\}.
\end{array}
\]
This normal non deterministic logic program, $\Pi$, has a well-founded total non deterministic model which is
$
\left\{ \; \left\{
\begin{array}{r}
a_1, \\
a_2
\end{array}
\right\}
, \; not \;
\left\{
\begin{array}{r}
b_1, \\
b_2
\end{array}
\right\}
\; \right\}
$.
Starting from the initial non deterministic interpretation, $I_0 = \emptyset$, then
\[
U_\Pi(I_0)= \left\{ \; \left\{
\begin{array}{r}
b_1, \\
b_2
\end{array}
\right\} \;
\right\}.
\]
and $T_\Pi(I_0) = \emptyset$. Thus,
\[
I_1 = W_\Pi(I_0) =
\left\{ \;
not \;
\left\{
\begin{array}{r}
b_1, \\
b_2
\end{array}
\right\}
\right\}.
\]
Furthermore,
\\
\\
$U_\Pi(I_1) =
\left\{ \; \left\{
\begin{array}{r}
b_1, \\
b_2
\end{array}
\right\}
\;
\right\}
$
and $T_\Pi(I_1)  =
\left\{ \; \left\{
\begin{array}{r}
a_1, \\
a_2
\end{array}
\right\} \; \right\}
$.
Hence,
\[
I_2 = W_\Pi(I_1) =
\left\{ \;
\left\{
\begin{array}{r}
a_1, \\
a_2
\end{array}
\right\}
,
\; not \;
\left\{
\begin{array}{r}
b_1, \\
b_2
\end{array}
\right\}
\; \right\}.
\]
which is the least upper bound of $I_0$, $I_1$, and $I_2$. Consequently, $I_2$ is the well-founded total non deterministic model for $\Pi$.

The set of answer sets, $S_{I_2}$, corresponding to the well-founded total non deterministic model, $I_2$, of $\Pi$ consists of the following answer sets:
\[
\begin{array}{c}
S_1 = \{\;  a_1, \: not\; b_1  \;\} \\
S_2 = \{\;  a_1, \: not\; b_2  \;\} \\
S_3 = \{\;  a_2, \: not\; b_1  \;\} \\
S_4 = \{\;  a_2, \: not\; b_2  \;\}
\end{array}
\]
\end{example}

\begin{example} Consider the following normal non deterministic logic program, $\Pi$, that consists of the normal non deterministic logic rules

\[
\begin{array}{lcl}
\left\{
\begin{array}{r}
a_1, \\
a_2
\end{array}
\right\}
&  \leftarrow &
not \;
\left\{
\begin{array}{r}
b_1, \\
b_2
\end{array}
\right\}
\\
\\
\left\{
\begin{array}{r}
b_1, \\
b_2
\end{array}
\right\}
&  \leftarrow &
not \;
\left\{
\begin{array}{r}
a_1, \\
a_2
\end{array}
\right\}
\\
\\
\left\{
\begin{array}{r}
c_1, \\
c_2
\end{array}
\right\}
&  \leftarrow &
not \;
\left\{
\begin{array}{r}
d_1, \\
d_2
\end{array}
\right\}
\end{array}
\]
This normal non deterministic logic program has a well-founded partial non deterministic model which is
$
\left\{ \; \left\{
\begin{array}{r}
c_1, \\
c_2
\end{array}
\right\}
, \; not \;
\left\{
\begin{array}{r}
d_1, \\
d_2
\end{array}
\right\} \; \right\}
$.
Starting from the initial non deterministic interpretation, $I_0 = \emptyset$, then $U_\Pi(I_0) =
\left\{ \; \left\{
\begin{array}{r}
d_1, \\
d_2
\end{array}
\right\} \; \right\}
$
and $T_\Pi (I_0) = \emptyset$. Thus,
\[
I_1 = W_\Pi (I_0) =
\left\{ \; not \; \left\{
\begin{array}{r}
d_1, \\
d_2
\end{array}
\right\} \; \right\}.
\]
In addition, $U_\Pi(I_1) =
\left\{ \; \left\{
\begin{array}{r}
d_1, \\
d_2
\end{array}
\right\} \; \right\}
$
and $T_\Pi (I_1) =
\left\{ \; \left\{
\begin{array}{r}
c_1, \\
c_2
\end{array}
\right\} \; \right\}
$. Therefore,
\[
I_2 = W_\Pi (I_1) =
\left\{
\left\{
\begin{array}{r}
c_1, \\
c_2
\end{array}
\right\}
,
\; not \; \left\{
\begin{array}{r}
d_1, \\
d_2
\end{array}
\right\} \; \right\}
,
\]
which is the least upper bound of $I_0$, $I_1$, and $I_2$. Consequently, $I_2$ is the well-founded partial non deterministic model for $\Pi$.

The set of answer sets, $S_{I_2}$, corresponding to the well-founded partial non deterministic model, $I_2$, of $\Pi$ consists of the following answer sets:
\[
\begin{array}{c}
S_1 = \{\;  c_1, \: not\; d_1  \;\} \\
S_2 = \{\;  c_1, \: not\; d_2  \;\} \\
S_3 = \{\;  c_2, \: not\; d_1  \;\} \\
S_4 = \{\;  c_2, \: not\; d_2  \;\}
\end{array}
\]

\end{example}

\section{Relation to Stable Non Deterministic Model Semantics}

In this section we study the relationship between the well-founded non deterministic model semantics and the stable non deterministic model semantics. We show that the relationship between the well-founded non deterministic model semantics and the stable non deterministic model semantics preserves the relationship between the deterministic well-founded semantics \cite{Gelder} and the deterministic stable model semantics \cite{Gelfond_A} for deterministic normal logic programs. We adopt the following terminology.

\begin{definition} Let $I$ be a partial or total non deterministic interpretation. Then $pos(I) = \{\: \{A_i\}_{i = 1}^n \: | \: \{A_i\}_{i = 1}^n \in I \}$ and $neg(I) = \{\: \{B_i\}_{i = 1}^n \: | \: not \; \{B_i\}_{i = 1}^n \in I \}$.
\end{definition}

However, in the context of the stable non deterministic model semantics we adopt the following definitions for $pos(I)$ and $neg(I)$. Let $I$ be a non deterministic interpretation, a non deterministic model, or a stable non deterministic model, in the sense of the stable non deterministic model semantics, then $pos(I) = \{\: \{A_i\}_{i = 1}^n \: | \: \{A_i\}_{i = 1}^n \in I \}$ and $neg(I) = \{\: \{B_i\}_{i = 1}^n \: | \{B_i\}_{i = 1}^n \notin I \; and \; \{B_i\}_{i = 1}^n \in {\cal N_L} \}$.

There is a close relationship between the well-founded non deterministic model and the stable non deterministic models for a normal non deterministic logic program.

To establish this relationship, we show that for a total non deterministic model, $I$, of a normal non deterministic logic program, $\Pi$, it is the case that $pos(lfp(T_{\Pi^I})) \subseteq T_\Pi(I)$, where $T_{\Pi^I}$ is the immediate consequence operator of the non deterministic logic programs, observe that $\Pi^I$ is a non deterministic logic program, and $T_\Pi(I)$ is the immediate consequence operator of the normal non deterministic logic programs defined in the context of the well-founded non deterministic model semantics. In addition, we show that the greatest unfounded non deterministic set of $\Pi$ with respect to $I$ is equivalent to $neg(lfp(T_{\Pi^I}))$, i.e., $neg(lfp(T_{\Pi^I})) = U_\Pi$. The following results show this close relationship.

\begin{theorem} Let $\Pi$ be a normal non deterministic logic program and $I$ be the well-founded partial or total non deterministic model of $\Pi$. Then $I$ is the only fixpoint of $W_\Pi$.
\label{thm:only_fix_Wp}
\end{theorem}

\begin{lemma} Let $I_1$ and $I_2$ be two non deterministic interpretations. If $I_1 \subseteq I_2$, then $pos(I_1) \subseteq pos(I_2)$.
\label{lemma:posI1-and-posI2}
\end{lemma}

\begin{lemma} Let $I$ be a total non deterministic model for the normal non deterministic logic program $\Pi$. Then $pos(lfp(T_{\Pi^I})) \subseteq pos(I)$.
\label{lemma:total111}
\end{lemma}

\begin{lemma} Let $\Pi$ be a normal non deterministic logic program and $I$ be a total non deterministic model for $\Pi$. Then $neg(lfp(T_{\Pi^I})) = U_\Pi(I)$.
\label{lemma:total2}
\end{lemma}

\begin{lemma} Let $\Pi$ be a normal non deterministic logic program and $I$ be a total non deterministic model for $\Pi$. Then $pos(lfp(T_{\Pi^I})) \subseteq T_\Pi(I)$.
\label{lemma:total3}
\end{lemma}
The following theorem establishes that given a normal non deterministic logic program, $\Pi$, and given a total non deterministic model $I$ of $\Pi$, then $I$ is a stable non deterministic model of $\Pi$ iff $I$ is a fixpoint of $W_\Pi$. It is possible that a stable non deterministic model of $\Pi$ is not the least fixpoint of $W_\Pi$. On the other hand, it is the case that if $I$ is a stable non deterministic model and the least fixpoint of $W_\Pi$, then $I$ is the unique stable non deterministic model of $\Pi$.

\begin{theorem} Let $\Pi$ be a normal non deterministic logic program and $I$ be a total non deterministic model for $\Pi$. Then $I$ is a stable non deterministic model for $\Pi$ iff $I$ is a fixpoint of $W_\Pi$.
\label{thm:sp-wp1}
\end{theorem}
The following corollary extends Theorem (\ref{thm:sp-wp1}) to total non deterministic interpretations.

\begin{corollary} Let $\Pi$ be a normal non deterministic logic program and $I$ be a total non deterministic interpretation for $\Pi$. Then $I$ is a stable non deterministic model for $\Pi$ iff $I$ is a fixpoint of $W_\Pi$. \label{coro:sp-wp1}
\end{corollary}

\begin{corollary} Let $\Pi$ be a normal non deterministic logic program, $I$ be the well-founded partial non deterministic model of $\Pi$, and $I'$ be a stable non deterministic model for $\Pi$. Then $I \subseteq I'$ for every stable non deterministic model $I'$ for $\Pi$.
\label{coro:sp-partial}
\end{corollary}

\begin{corollary} Let $\Pi$ be a normal non deterministic logic program and $I$ be the well-founded total non deterministic model of $\Pi$. Then $I$ is the unique stable non deterministic model of $\Pi$.
\label{coro:sp-wp2}
\end{corollary}
Hence, the following theorem immediately follows.

\begin{theorem} Every non deterministic logic program, $\Pi$, has a well-founded total non deterministic model, $I$, iff $I$ is the least non deterministic model for $\Pi$.
\label{thm:wf-model-for-nd-program}
\end {theorem}

\section{Relation to Deterministic Well-Founded Semantics}

In this section we establish the relationship between the well-founded non deterministic model semantics of the normal non deterministic logic programs and the deterministic well-founded semantics of the deterministic normal logic programs introduced in \cite{Gelder}. The deterministic well-founded semantics of the deterministic normal logic programs presented in \cite{Gelder} is deterministic in the sense that the normal logic programs considered in \cite{Gelder} and well-founded models defined for the normal logic programs in \cite{Gelder} allow a single atom (deterministic atom) as the building block for both the normal logic programs and their well-founded models. However, normal non deterministic logic programs and their well-founded non deterministic model semantics allow atoms to be non deterministic for which a non deterministic atom is represented by a set of atoms of the form $\{A_i\}_{i = 1}^n$. This implies that any deterministic atom, $A$, representation in the language of deterministic normal logic programs described in \cite{Gelder} can be represented as a non deterministic atom of the form, $\{A\}$, in the language of normal non deterministic logic programs. And hence, it can be shown that the syntax and the well-founded non deterministic model semantics of normal non deterministic logic programs naturally subsume the syntax and the deterministic well-founded semantics of the deterministic normal logic programs.

Any deterministic normal logic program, $\Upsilon$, can be represented as a normal non deterministic logic program, $\Pi$, where each deterministic normal logic rule of the form
\[
A \leftarrow B_1, \ldots, B_l, not \; B_{l+1},\ldots, not \; B_m \in \Upsilon
\]
can be represented as a normal non deterministic logic rule of the form
\[
\{A\} \leftarrow \{B_1\}, \ldots, \{B_l\}, not \; \{B_{l+1}\},\ldots, not \; \{B_m\} \in \Pi
\]
where $A, B_1, \ldots, B_l, B_{l+1},\ldots, B_m$ are atoms.

\begin{theorem} Let $\Upsilon$ be a deterministic normal logic program and $\Pi$ be the normal non deterministic logic program representation of $\Upsilon$. Then $I$ is a deterministic well-founded partial or total model for $\Upsilon$ iff $\;\forall \: A$ or  $not \; A \in I$, $\{A\}$ or  $not \; \{A\} \in J$ is a well-founded partial or total non deterministic model for $\Pi$.
\label{thm:wf-model-for-normal-program}
\end{theorem}

\section{Conclusions}

We presented the language of non deterministic logic programs, as well as, its declarative and fixpoint semantics. The language and semantics of non deterministic logic programs are necessary in order to logically enable real-world non deterministic applications, such as those in stochastic optimization, multi-objectives optimization, stochastic planning, contingent stochastic planning, reinforcement learning, reinforcement learning in partially observable Markov decision processes, and conditional planning.

We presented an extension of the language of non deterministic logic programs framework, called normal non deterministic logic programs, to cope with non-monotonic negation. The extension is a necessary requirement in many real-world non deterministic applications. We developed the well-founded non deterministic model semantics and the stable non deterministic model semantics for the normal non deterministic logic programs. We showed that the well-founded non deterministic model semantics and the stable non deterministic model semantics naturally subsume the deterministic well-founded semantics and the deterministic stable model semantics for deterministic normal logic programs. Furthermore, we showed that they naturally reduce to the semantics of non deterministic logic programs. Moreover, we showed that the relationship between the well-founded non deterministic model semantics and the stable non deterministic model semantics for normal non deterministic logic programs preserves the relationship between the deterministic well-founded semantics and the deterministic stable model semantics for deterministic normal logic programs. In addition, we showed the applicability of the normal non deterministic logic programs framework to a conditional planning problem, an inherently a non deterministic problem.

\section{Appendix: Proofs}

\textbf{Proof of Lemma (\ref{lemma:lattice}).} We show that for every subset $X$ of $2^{\cal N_L}$, there exist $lub(X)$ and $glb(X)$ defined as
\[
lub(X) = \bigcup_{S \in X} \; S
\]
and
\[glb(X) = \bigcap_{S \in X} \; S
\]
We show that $\bigcup_{S \in X} \; S$ and $\bigcap_{S \in X} \; S$ are an upper bound and a lower bound for $X$ respectively and for any $U \in 2^{\cal N_L}$, an upper bound of $X$, $\bigcup_{S \in X} \; S \subseteq U$ and for any $L \in 2^{\cal N_L}$, a lower bound of $X$, $L \subseteq \bigcap_{S \in X} \; S$.
\\
\\
For any $T \in X$, we have
\[
T \subseteq  \bigcup_{S \in X} \; S
\]
Similarly,
\[
\bigcap_{S \in X} \; S \subseteq  T
\]
Let $U \in 2^{\cal N_L}$ be an upper bound for $X$, then for all $x \in X$, $x \subseteq U$. However, for all $x \in X$, $x \subseteq \bigcup_{S \in X} \; S  = lub(X)$. Hence, $lub(X) = \bigcup_{S \in X} \; S  \subseteq U$. Similarly, Let $L \in 2^{\cal N_L}$ be a lower bound for $X$, then for all $x \in X$, $L \subseteq x$. However, for all $x \in X$, $ \bigcap_{S \in X} \; S = glb(X) \subseteq x$. Hence, $L \subseteq \bigcap_{S \in X} \; S = glb(X)$. $\blacksquare$
\\
\\
\textbf{Proof of Proposition (\ref{prop:models}).} Let $\Pi$ be a ground non deterministic logic program. We prove the proposition by showing that for any non deterministic logic rule, $r \in \Pi$, of the form $\{ A_i \}_{i = 1}^{n} \leftarrow \{ B_{i_1} \}_{i_1 = 1}^{n_1} , \ldots, \{ B_{i_m} \}_{i_m = 1}^{n_m}$, whenever $I_1$ and $I_2$ satisfy $r \in \Pi$, so does $I_1 \cap I_2$. Let $body(r)$ denotes $\{ B_{i_1} \}_{i_1 = 1}^{n_1} , \ldots, \{ B_{i_m} \}_{i_m = 1}^{n_m}$. Thus, two cases are possible.

\begin{enumerate}

\item Let us assume that $I_1$ or $I_2$ (or both) do not satisfy the body of $r$, i.e, there exists $1 \leq j \leq m$, such that $\{ B_{i_j} \}_{i_j = 1}^{n_j} \notin I_1$ or $\{ B_{i_j} \}_{i_j = 1}^{n_j} \notin I_2$. Then $\{ B_{i_j} \}_{i_j = 1}^{n_j} \notin I_1 \cap I_2$, which implies that $I_1 \cap I_2$ also does not satisfy the body of $r$. This means that $I_1 \cap I_2$ satisfies $r$.

\item Let us assume that $I_1 \models body(r)$ and $I_2 \models body(r)$. Therefore, $\{ A_i \}_{i = 1}^{n} \in I_1$ whenever $\forall (1 \leq j \leq m)$, $\{ B_{i_j} \}_{i_j = 1}^{n_j} \in I_1$. In addition, $\{ A_i \}_{i = 1}^{n} \in I_2$ whenever $\forall (1 \leq j \leq m)$, $\{ B_{i_j} \}_{i_j = 1}^{n_j} \in I_2$. Therefore, $\{ A_i \}_{i = 1}^{n} \in I_1 \cap I_2$ whenever $\forall (1 \leq j \leq m)$, $\{ B_{i_j} \}_{i_j = 1}^{n_j} \in I_1 \cap I_2$.  This implies that $I_1 \cap I_2$ satisfies $r$. $\blacksquare$
\end{enumerate}
\textbf{Proof of Theorem (\ref{thm:least_pmodel}).} The proof follows directly from Proposition (\ref{prop:models}).
$\blacksquare$
\\
\\
\textbf{Proof of Lemma (\ref{lemma:unique_least_pmodel}).} The proof follows directly from the fact that the least non deterministic model, $I_\Pi$, of a non deterministic logic program, $\Pi$, is the smallest subset of the non deterministic base, ${\cal N_L}$, that satisfies $\Pi$ which is a unique subset. $\blacksquare$
\\
\\
\textbf{Proof of Lemma (\ref{lemma:Tp_mono_continue})}. Let $I_1, I_2$ be two non deterministic interpretations for the ground non deterministic logic program $\Pi$. To prove that $T_\Pi$ is monotonic, we show that if $I_1 \subseteq I_2$, then $T_\Pi(I_1) \subseteq T_\Pi(I_2)$.
By the definition of $T_\Pi$, we have
\[
\begin{array}r
T_\Pi(I_1)= \{ \; \{ A_i \}_{i = 1}^{n} \; | \;  \{ A_i \}_{i = 1}^{n} \leftarrow \{ B_{i_1} \}_{i_1 = 1}^{n_1} , \ldots, \{ B_{i_m} \}_{i_m = 1}^{n_m} \in \Pi \\  and\; \forall \: (1 \leq j \leq m), \{B_{i_j} \}_{i_j = 1}^{n_j} \in I_1 \}.
\end{array}
\]
In addition, we have
\[
\begin{array}r
T_\Pi(I_2)= \{ \; \{ A_i \}_{i = 1}^{n} \; | \;  \{ A_i \}_{i = 1}^{n} \leftarrow \{ B_{i_1} \}_{i_1 = 1}^{n_1} , \ldots, \{ B_{i_m} \}_{i_m = 1}^{n_m} \in \Pi \\  and\; \forall \: (1 \leq j \leq m), \{B_{i_j} \}_{i_j = 1}^{n_j} \in I_2 \}.
\end{array}
\]
However, since $I_1 \subseteq I_2$, we have for each $ \{ A_i \}_{i = 1}^{n} \in T_\Pi(I_1)$, we also have $ \{ A_i \}_{i = 1}^{n} \in T_\Pi(I_2)$, which implies that $T_\Pi(I_1) \subseteq T_\Pi(I_2)$. This means that $T_\Pi$ is monotonic.

Now, to prove that $T_\Pi$ is continuous, we show that for any set of non deterministic interpretations, $X$, it is the case that
\[
T_\Pi(\:lub \: \{\; I \; | \; I \in X \; \}\:) = lub \: \{ \; T_\Pi(I) \; | \; I \in X \; \}
\]
I.e., we want to show that
\[
T_\Pi(\: \cup \: \{\; I \; | \; I \in X \; \}\:) = \cup \: \{ \; T_\Pi(I) \; | \; I \in X \; \}
\]
We proceed by induction on the cardinality of $X$, denoted by $| X |$, as follows.

\begin{itemize}

\item Base case: $|X| = 0$, which implies that $X = \emptyset$. This means that
\[
T_\Pi(\: \cup \: \emptyset  \:) = \cup \: \{ \; T_\Pi(\emptyset)  \; \}
\]
which is obviously,
\[
T_\Pi(\emptyset) =  T_\Pi(\emptyset)
\]
Similarly, if $|X| = 1$, which implies that $X = I$.
This means that
\[
T_\Pi(\: \cup \: I  \:) = \cup \: \{ \; T_\Pi(I)  \; \}
\]
which implies,
\[
T_\Pi(I) =  T_\Pi(I)
\]
\item Inductive hypothesis: Let for $| X | = k$ it is true that
\[
T_\Pi(\: \cup \: \{\; I \; | \; I \in X \; \}\:) = \cup \: \{ \; T_\Pi(I) \; | \; I \in X \; \}
\]

\item Induction: Let $J$ be a non deterministic interpretation then
\[
\begin{array}{lcl}
T_\Pi(\: \cup \: \{\; I \; | \; I \in X \cup \{J\} \; \}\:) &  = & T_\Pi(\: \cup \: \{\; I \; | \; I \in X \; \}\:)
\cup
T_\Pi(\: \cup \: \{\; J \; \}\:)
\\
\\
& = & \cup \: \{ \; T_\Pi(I) \; | \; I \in X \; \} \cup T_\Pi(\: \cup \: \{\; J \; \}\:)
\\
\\
& = & \cup \: \{ \; T_\Pi(I) \; | \; I \in X \; \} \cup T_\Pi(\: \{\; J \; \}\:)
\\
\\
& = & \cup \: \{ \; T_\Pi(I) \; | \; I \in X \cup \{\; J \; \} \; \}.
\end{array}
\]
\end{itemize}
This implies that for any set of non deterministic interpretations, $X$, it is the case that
\[
T_\Pi(\: \cup \: \{\; I \; | \; I \in X \; \}\:) = \cup \: \{ \; T_\Pi(I) \; | \; I \in X \; \}
\]
$\blacksquare$
\\
\\
\textbf{Proof of Theorem (\ref{thm:Tp_subset_I}).} Let $\Pi$ be a ground non deterministic logic program. The proof proceeds as follows.

\begin{itemize}

\item First we prove that if $T_\Pi(I) \subseteq I$, then $I$ is a non deterministic model for $\Pi$. To prove that $I$ is a non deterministic model for $\Pi$, for any non deterministic logic rule $r \in \Pi$ of the form
\[
\{ A_i \}_{i = 1}^{n} \leftarrow \{ B_{i_1} \}_{i_1 = 1}^{n_1} , \ldots, \{ B_{i_m} \}_{i_m = 1}^{n_m}
\]
we want to show that $I$ satisfies $r$ as $T_\Pi(I) \subseteq I$. Since $T_\Pi(I) \subseteq I$, then

\begin{itemize}

\item  if $\exists (1 \leq j \leq m)$ such that $\{ B_{i_j} \}_{i_j = 1}^{n_j} \notin I$, then
$I$ does not satisfy $\{ B_{i_j} \}_{i_j = 1}^{n_j}$, and hence $I$ does not satisfy the body of $r$, and therefore $I$ satisfies $r$.

\item if $\forall (1 \leq j \leq m)$ such that $\{ B_{i_j} \}_{i_j = 1}^{n_j} \in I$, then $I$ satisfies the body of $r$. By the definition of the $T_\Pi$ we have

\[
\begin{array}r
T_\Pi(I)= \{ \; \{ A_i \}_{i = 1}^{n} \; | \;  \{ A_i \}_{i = 1}^{n} \leftarrow \{ B_{i_1} \}_{i_1 = 1}^{n_1} , \ldots, \{ B_{i_m} \}_{i_m = 1}^{n_m} \in \Pi \\  and\; \forall \: (1 \leq j \leq m), \{B_{i_j} \}_{i_j = 1}^{n_j} \in I \}.
\end{array}
\]
This implies that $\forall (1 \leq j \leq m)$, $\{ B_{i_j} \}_{i_j = 1}^{n_j} \in T_\Pi(I)$, which in turn implies that $\{ A_i \}_{i = 1}^{n} \in T_\Pi(I)$. Since $\forall (1 \leq j \leq m)$, $\{ B_{i_j} \}_{i_j = 1}^{n_j} \in T_\Pi(I)$, $\{ A_i \}_{i = 1}^{n} \in T_\Pi(I)$ and $T_\Pi(I) \subseteq I$, then $\{ A_i \}_{i = 1}^{n} \in I$. This means that $I$ satisfies $r$.

\end{itemize}

\item Second we prove that if $I$ is a non deterministic model for $\Pi$, then $T_\Pi(I) \subseteq I$.

\begin{itemize}

\item  From the definition of $T_\Pi$, we have that

\[
\begin{array}r
T_\Pi(I)= \{ \; \{ A_i \}_{i = 1}^{n} \; | \;  \{ A_i \}_{i = 1}^{n} \leftarrow \{ B_{i_1} \}_{i_1 = 1}^{n_1} , \ldots, \{ B_{i_m} \}_{i_m = 1}^{n_m} \in \Pi \\  and\; \forall \: (1 \leq j \leq m), \{B_{i_j} \}_{i_j = 1}^{n_j} \in I \}.
\end{array}
\]
Since $I$ satisfies the body of each non deterministic logic rule $r \in \Pi$ and consequently satisfies
$\Pi$, then $I$ must satisfy the head $\{ A_i \}_{i = 1}^{n}$ of each non deterministic logic rule $r \in \Pi$, i.e., $\{ A_i \}_{i = 1}^{n} \in I$. This means that $T_\Pi(I) \subseteq I$. $\blacksquare$

\end{itemize}
\end{itemize}
\textbf{Proof of Theorem (\ref{thm:least_model_least_fixpoint}).} Follows from Theorem (\ref{thm:least_pmodel}), Lemma (\ref{lemma:Tp_mono_continue}) and Theorem (\ref{thm:Tp_subset_I}) as follows. Since the $T_\Pi$ operator is monotonic (from Lemma (\ref{lemma:Tp_mono_continue})) and $\langle 2^{\cal N_L}, \subseteq \rangle$ forms a complete lattice (from Lemma (\ref{lemma:lattice})), then $T_\Pi$ has a least fixpoint, $lfp(T_\Pi)$, which is defined as $lfp(T_\Pi) = \bigcap_{ T_\Pi(I) \subseteq I } \; I $. (This is true since for any complete lattice $L$ and a monotonic mapping $T:L \rightarrow L$, $lfp(T) = glb\{x | T(x) \leq x\}$ \cite{Lloyd}). We have $I_\Pi = \bigcap_{ I \in {\cal I}_\Pi } \; I$, by Theorem (\ref{thm:least_pmodel}), where ${\cal I}_\Pi$ is the set of all non deterministic models of $\Pi$. But $I$ is a non deterministic model of $\Pi$ iff $T_\Pi(I) \subseteq I$ by Theorem (\ref{thm:Tp_subset_I}). Therefore, $I_\Pi = \bigcap_{ T_\Pi(I) \subseteq I } \; I = lfp(T_P)$. This implies $I_\Pi = lfp(T_\Pi)$. $\blacksquare$
\\
\\
\textbf{Proof of Theorem (\ref{thm:relate_to_definite}).} The proof follows directly from the definition of deterministic Herbrand models for deterministic definite logic programs \cite{Lloyd} and the definition of the non deterministic models of non deterministic logic programs.  $\blacksquare$
\\
\\
\textbf{Proof of Theorem (\ref{thm:equiv2definite}).} Let $\Upsilon$ be a ground deterministic definite logic program, $T_\Upsilon$ be the immediate consequence operator of $\Upsilon$ as defined in \cite{Lloyd}, and $\Pi$ be the ground non deterministic logic program representation of $\Upsilon$.

Since the least deterministic Herbrand model, $I_\Upsilon$, of $\Upsilon$ coincides with the least fixpoint of the immediate consequence operator, $T_\Upsilon$, of $\Upsilon$, and the least non deterministic model, $J_\Pi$, of $\Pi$ coincides with the least fixpoint of the immediate consequence operator, $T_\Pi$, of $\Pi$, it is sufficient to show that
\[
\forall \: A  \in lfp(T_\Upsilon) \leftrightarrow  \{A\} \in lfp(T_\Pi).
\]
where $(\leftrightarrow)$ means if and only if. However, it is known that for both semantics it is the case that $lfp(T_\Upsilon)= T_\Upsilon \uparrow \omega$ and $lfp(T_\Pi)= T_\Pi \uparrow \omega$, where $\omega$ is a limit ordinal.
\\
\\
So that it is sufficient to prove this theorem to show that
\[
\forall \: A  \in T_\Upsilon \uparrow \omega  \leftrightarrow  \{A\} \in T_\Pi \uparrow \omega.
\]
This is achieved by showing that for each fixpoint iteration, $i$, it is the case that
\[
\forall \: A  \in T_\Upsilon \uparrow i \leftrightarrow  \{A\} \in T_\Pi \uparrow i.
\]
We proceed by induction on $i$ as follows.

\begin{itemize}
\item Base case: $i = 0$

$T_\Upsilon \uparrow 0 = \emptyset$ \cite{Lloyd}.

$T_\Pi \uparrow 0 = \emptyset$.

Hence, $\forall \: A  \in T_\Upsilon \uparrow 0 \leftrightarrow  \{A\} \in T_\Pi \uparrow 0.$

\item Inductive hypothesis: Let for $k < i$, it is true that
\[
\forall \: A  \in T_\Upsilon \uparrow k  \leftrightarrow  \{A\} \in T_\Pi \uparrow k.
\]

\item Induction:

From the inductive hypothesis we know that
\[
\forall \: A  \in T_\Upsilon \uparrow (i - 1) \leftrightarrow  \{A\} \in T_\Pi \uparrow (i - 1).
\]
Let for any $A \in {\cal B_L}$,
\[
\begin{array}{lcl}
A & \leftarrow & B_{1,1}, B_{1,2}, \ldots, B_{1,m_1} \\
A & \leftarrow & B_{2,1}, B_{2,2}, \ldots, B_{1,m_2} \\ \\
& \ldots & \\ \\
A & \leftarrow & B_{n,1}, B_{n,2}, \ldots, B_{n,m_n}
\end{array}
\]
be the list of $n$ deterministic definite logic rules in $\Upsilon$ whose head is, $A$, and their bodies are
satisfied by $T_\Upsilon \uparrow (i - 1)$. In addition, let

\[
\begin{array}{lcl}
\{ A \} & \leftarrow & \{ B_{1,1} \}, \{ B_{1,2} \} , \ldots, \{ B_{1,m_1} \} \\
\{ A \} & \leftarrow & \{ B_{2,1} \}, \{ B_{2,2} \}, \ldots, \{ B_{1,m_2} \}\\ \\
& \ldots & \\ \\
\{ A \} & \leftarrow & \{ B_{n,1} \} , \{ B_{n,2} \}, \ldots, \{ B_{n,m_n} \}
\\ \\
\end{array}
\]
be the list of $n$ non deterministic logic rules in $\Pi$ whose head is, $\{ A \}$, and their bodies are
satisfied by $T_\Pi \uparrow (i - 1)$. Consequently, by the definition of $T_\Upsilon$, we must have that
\[
A \in T_\Upsilon(T_\Upsilon \uparrow (i - 1)).
\]
which implies that
\[
A \in T_\Upsilon \uparrow i.
\]
In addition, we must have that
\[
\{ A \} \in T_\Pi(T_\Pi \uparrow (i - 1)).
\]
which implies that
\[
\{ A \} \in T_\Pi \uparrow i.
\]
However, by the induction hypothesis we have that
\[
\forall \: A  \in T_\Upsilon \uparrow (i - 1) \leftrightarrow  \{A\} \in T_\Pi \uparrow (i - 1).
\]
In addition, by induction we have shown that for any $A \in {\cal B_L}$,
\[
A  \in T_\Upsilon(T_\Upsilon \uparrow (i - 1)) \leftrightarrow   \{ A \} \in T_\Pi(T_\Pi \uparrow (i - 1)).
\]

Therefore,
\[
\forall \; A  \in T_\Upsilon \uparrow i  \leftrightarrow   \{ A \} \in T_\Pi \uparrow i.
\]
This implies that
\[
\forall \: A  \in T_\Upsilon \uparrow \omega  \leftrightarrow  \{A\} \in T_\Pi \uparrow \omega.
\]
Consequently implies
\[
\forall \: A  \in lfp(T_\Upsilon) \leftrightarrow  \{A\} \in lfp(T_\Pi).
\]
 $\blacksquare$
\end{itemize}
\textbf{Proof of Theorem (\ref{thm:stable_model_minimal})}. We prove this theorem by contradiction. Assume that $I_1$ is a non deterministic model for $\Pi$ and $I$ is a stable non deterministic model for $\Pi$ and $I_1$ is a subset of $I$, i.e., $I_1 \subset I$. We show that $I_1$ is not a non deterministic model for $\Pi$ that contradicting our assumption that $I_1$ is a non deterministic model for $\Pi$.

Since $I$ is a stable non deterministic model for $\Pi$, it is a subset-minimal non deterministic model for $\Pi$ by definition. Therefore, $I_1$ is not a non deterministic model of the reduct, $\Pi^I$ of $\Pi$, otherwise $I$ cannot be a subset-minimal non deterministic model of $\Pi^I$. Consequently, there must be some non deterministic logic rules $r$ in $\Pi^I$ that are not satisfied by $I_1$. This is true because $I_1 \subset I$ which means the number of non deterministic atoms in $I_1$ is less than the number of non deterministic atoms in $I$ which makes some non deterministic logic rules in $\Pi^I$ that are satisfied by $I$ are no longer satisfied by $I_1$. And hence, $I_1$ does not satisfy $\Pi^I$ and cannot be a non deterministic model for $\Pi^I$. Consequently, $I_1$ does not satisfy $\Pi$ either and cannot be a non deterministic model for $\Pi$, which contradicting our assumption that $I_1$ is a non deterministic model for $\Pi$. $\blacksquare$
\\
\\
\textbf{Proof of Theorem (\ref{thm:sp-non-deterministic-program}).} Follows directly from the fact that the non deterministic reduct of the non deterministic logic program $\Pi$ with respect to any non deterministic interpretation is $\Pi$ itself. $\blacksquare$
\\
\\
\textbf{Proof of Theorem (\ref{thm:Tp'-and-Tp}).} Follows directly from the definitions of $T^\prime_\Pi$ and $T_\Pi$. $\blacksquare$
\\
\\
\textbf{Proof of Lemma (\ref{lemma:rel_smodels}).} Let $\Pi$ be a ground normal non deterministic logic program and $\Pi^I$ be the non deterministic reduct of $\Pi$ with respect to $I$. To prove this lemma, it is sufficient to prove that for each non deterministic atom $\{ A \}_{i = 1}^n \in I$, it holds that $\{ A \}_{i = 1}^n \in T'_\Pi(I)$ if and only if $\{ A \}_{i = 1}^n \in T_{\Pi^I}(I)$, where
\[
\begin{array}{lcl}
T^\prime_\Pi(I)= \{ \quad \{ A_i \}_{i = 1}^{n} \; | \;  \{ A_i \}_{i = 1}^{n} & \leftarrow &\{ B_{i_1} \}_{i_1 = 1}^{n_1} , \ldots, \{ B_{i_l} \}_{i_l = 1}^{n_l}, \\
&& not \: \{ B_{i_{l+1}} \}_{i_{l+1} = 1}^{n_{l+1}} , \ldots, not \: \{ B_{i_m} \}_{i_m = 1}^{n_m} \; \in \Pi
\end{array}
\]
and $\forall (1 \leq j \leq l)$ $\{ B_{i_j} \}_{i_j = 1}^{n_j} \in I$ and $\forall (l+1 \leq k \leq m$) $\{ B_{i_k} \}_{i_k = 1}^{n_k} \notin I \; \; \}$, and
\[
T_{\Pi^I}(I)= \{ \; \{ A_i \}_{i = 1}^{n} \; | \;  \{ A_i \}_{i = 1}^{n} \leftarrow \{ B_{i_1} \}_{i_1 = 1}^{n_1} , \ldots, \{ B_{i_l} \}_{i_l = 1}^{n_l} \in \Pi^I
\]
and $\forall \: (1 \leq j \leq l), \{B_{i_j} \}_{i_j = 1}^{n_j} \in I  \; \}$.
\\
\\
This is because $I = lfp(T_{\Pi^I})$. Let $r'$ be a normal non deterministic logic rule in $\Pi$ of the form
\begin{equation}
\{ A_i \}_{i = 1}^{n} \leftarrow \{ B_{i_1} \}_{i_1 = 1}^{n_1} , \ldots, \{ B_{i_l} \}_{i_l = 1}^{n_l},  not \: \{ B_{i_{l+1}} \}_{i_{l+1} = 1}^{n_{l+1}} , \ldots, not \: \{ B_{i_m} \}_{i_m = 1}^{n_m} \label{eq:normal_rule}
\end{equation}
In addition, let $r$ be a non deterministic logic rule in $\Pi^I$ of the form
\begin{equation}
\{ A_i \}_{i = 1}^{n} \leftarrow \{ B_{i_1} \}_{i_1 = 1}^{n_1} , \ldots, \{ B_{i_l} \}_{i_l = 1}^{n_l} \label{eq:rule}
\end{equation}

\begin{enumerate}

\item Case 1: $T'_\Pi(I) \neq \emptyset$. For any normal non deterministic logic rule, $r' \in \Pi$, of the form (\ref{eq:normal_rule}) such that $\forall (1 \leq j \leq l)$ $\{ B_{i_j} \}_{i_j = 1}^{n_j} \in I$ and $\forall (l+1 \leq k \leq m$) $\{ B_{i_k} \}_{i_k = 1}^{n_k} \notin I$, we have $\{ A_i \}_{i = 1}^{n} \in  T'_\Pi(I)$ {\em if and only if} $\forall (l+1 \leq k \leq m$) $\{ B_{i_k} \}_{i_k = 1}^{n_k} \notin I$, $r \in \Pi^I$, by the definition of the non deterministic reduct, and $\forall (1 \leq j \leq l)$ $\{ B_{i_j} \}_{i_j = 1}^{n_j} \in I$, we have $\{ A_i \}_{i = 1}^{n} \in T_{\Pi^I}(I)$. Hence it must be the case that $\{ A \}_{i = 1}^n \in T'_\Pi(I)$ {\em if and only if} $\{ A \}_{i = 1}^n \in T_{\Pi^I}(I)$, and obviously $T_{\Pi^I}(I) \neq \emptyset$.

\item Case 2: $T'_\Pi(I) = \emptyset$. This means that there is no $r' \in \Pi$, of the form (\ref{eq:normal_rule}) such that $\forall (1 \leq j \leq l)$ $\{ B_{i_j} \}_{i_j = 1}^{n_j} \in I$ and $\forall (l+1 \leq k \leq m$) $\{ B_{i_k} \}_{i_k = 1}^{n_k} \notin I$ {\em if and only if} there is no $r \in \Pi^I$ such that $\forall (l+1 \leq k \leq m$) $\{ B_{i_k} \}_{i_k = 1}^{n_k} \notin I$, by the definition of the non deterministic reduct, and $\forall (1 \leq j \leq l)$ $\{ B_{i_j} \}_{i_j = 1}^{n_j} \in I$. Hence $T'_\Pi(I) = T_{\Pi^I}(I) = \emptyset$.

\end{enumerate}
Therefore, it follows that $T'_\Pi(I) = T_{\Pi^I}(I)$. But since $I$ is a stable non deterministic model for $\Pi$, then $I = T_{\Pi^I}(I)$. This means that $I$ is a fixpoint of $T_{\Pi^I}$, i.e., $I = T_{\Pi^I}(I)$. Consequently, it must be that $I = T'_\Pi(I)$. $\blacksquare$
\\
\\
\textbf{Proof of Theorem (\ref{thm:rel_smodels}).} Let $\Pi$ be a ground normal non deterministic logic program and $\Pi^I$ be the non deterministic reduct of $\Pi$ with respect to $I$. By Lemma (\ref{lemma:rel_smodels}), $I$ is a fixpoint of $T'_\Pi$. Thus
it is sufficient to prove that $I$ is a minimal fixpoint of $T'_\Pi$.

We prove this theorem by contradiction. Suppose that there exists a non deterministic interpretation, $I_1$, such that $I_1$ is a fixpoint of $T'_\Pi$ and $I_1 \subset I$. Then there must exist some non deterministic atoms, $\{ A_i \}_{i = 1}^n \in {\cal N_L}$ such that $\{ A_i \}_{i = 1}^n \notin I_1$ and $\{ A_i \}_{i = 1}^n \in I$.

Let $\gamma = \min \{k \: | \: $ there exist $\{ A_i \}_{i = 1}^n \in {\cal N_L}$ such that $k$ is the smallest ordinal such that $\{ A_i \}_{i = 1}^n \notin T_{\Pi^I} \uparrow k \}$. We proceed by induction on $\gamma$ and show that a contradiction arises.

\begin{itemize}

\item Base case: $\gamma = 1$. We show that there exists some non deterministic atom $\{ A_i \}_{i = 1}^n \in {\cal N_L}$ such that $\{ A_i \}_{i = 1}^n \in T_{\Pi^I} \uparrow 1$ and $\{ A_i \}_{i = 1}^n \notin I_1$.
    \\
    \\
    Let $g = T_{\Pi^I} \uparrow 0$. Then
    \[
    T_{\Pi^I}(g) = \{\: \{ A_i \}_{i = 1}^n \; | \; \{ A_i \}_{i = 1}^n \leftarrow \quad \in \Pi^I \}.
    \]

\begin{enumerate}

\item Case 1: $T_{\Pi^I}(g) \neq \emptyset$. For each non deterministic logic rule $r \in \Pi^I$ of the form
\[
\{ A_i \}_{i = 1}^n \leftarrow
\]
we have $\{ A_i \}_{i = 1}^n \in T_{\Pi^I}(g)$ {\em if and only if} there exists a normal non deterministic logic rule $r' \in \Pi$ of the form
\[
\{ A_i \}_{i = 1}^n \leftarrow not \: \{ B_{i_{l+1}} \}_{i_{l+1} = 1}^{n_{l+1}} , \ldots, not \: \{ B_{i_m} \}_{i_m = 1}^{n_m}
\]
such that $\forall (l+1 \leq k \leq m$), we have $\{ B_{i_k} \}_{i_k = 1}^{n_k} \notin I$.
\\
\\
Since $I_1 \subset I$, then it follows that $\forall (l+1 \leq k \leq m$), $\{ B_{i_k} \}_{i_k = 1}^{n_k} \notin I_1$.
\\
\\
Therefore, $\{ A_i \}_{i = 1}^n \in T'_\Pi(I_1)$. However, $T'_\Pi(I_1) = I_1$, which implies that $\{ A_i \}_{i = 1}^n \in I_1$, a contradiction.
\\

\item Case 2: $T_{\Pi^I}(g) = \emptyset$. Then, $T_{\Pi^I}(g) = \emptyset$ {\em if and only if} $T'_\Pi(I_1) = \emptyset$, since $I_1 \subset I$.  However, $T'_\Pi(I_1) = I_1$, which implies that $\forall \; \{ A_i \}_{i = 1}^n \in T_{\Pi^I}(g)$, $\{ A_i \}_{i = 1}^n \in I_1$, a contradiction.

\end{enumerate}
By combining these two cases together, it is the case that
\[
\forall \; \{ A_i \}_{i = 1}^n \in T_{\Pi^I}(g), \; \{ A_i \}_{i = 1}^n \in I_1.
\]
which is a contradiction. In particular, it must be the case that
\[
\forall \; \{ A_i \}_{i = 1}^n \in T_{\Pi^I}(g) = T_{\Pi^I} \uparrow 1, \{ A_i \}_{i = 1}^n \in T'_\Pi(I_1) = I_1.
\]
A contradiction.

\item Induction hypothesis: For all $\delta \leq \xi$, it is the case that
\[
\forall \{ A_i \}_{i = 1}^n \in T_{\Pi^I} \uparrow \delta, \; \{ A_i \}_{i = 1}^n \in I_1.
\]
A contradiction.

\item Inductive case. There are two parts, one in which $\gamma$ is a successor ordinal and one where $\gamma$ is a limit ordinal.

\begin{itemize}

\item Successor ordinal case: $\gamma = \xi + 1$. Then there exists some non deterministic atom $\{ A_i \}_{i = 1}^n \in {\cal N_L}$ such that
\[
\{ A_i \}_{i = 1}^n \in T_{\Pi^I} \uparrow (\xi + 1),  \{ A_i \}_{i = 1}^n \notin I_1.
\]
Let $g = T_{\Pi^I} \uparrow \xi$. Then
\[
T_{\Pi^I}(g) = \{ \; \{ A_i \}_{i = 1}^{n} \; | \; \{ A_i \}_{i = 1}^{n} \leftarrow \{ B_{i_1} \}_{i_1 = 1}^{n_1} , \ldots, \{ B_{i_l} \}_{i_l = 1}^{n_l} \in \Pi^I
\]
and $\forall(1 \leq j \leq l), \{ B_{i_j} \}_{i_j = 1}^{n_j} \in I \; \}$.

\begin{enumerate}

\item Case 1: $T_{\Pi^I}(g) \neq \emptyset$.  Each non deterministic logic rule, $r$, of the form
\[
\{ A_i \}_{i = 1}^{n} \leftarrow \{ B_{i_1} \}_{i_1 = 1}^{n_1} , \ldots, \{ B_{i_l} \}_{i_l = 1}^{n_l}
\]
belongs to $\Pi^I$ {\em if and only if} there exists a normal non deterministic logic rule, $r'$, of the form
\[
\{ A_i \}_{i = 1}^{n} \leftarrow \{ B_{i_1} \}_{i_1 = 1}^{n_1} , \ldots, \{ B_{i_l} \}_{i_l = 1}^{n_l}, not \: \{ B_{i_{l+1}} \}_{i_{l+1} = 1}^{n_{l+1}} , \ldots, not \: \{ B_{i_m} \}_{i_m = 1}^{n_m}
\]
belongs to $\Pi$ such that $\forall (l+1 \leq k \leq m$), $\{ B_{i_k} \}_{i_k = 1}^{n_k} \notin I$. Since $I_1 \subset I$, then it follows that $\forall (l+1 \leq k \leq m$), $\{ B_{i_k} \}_{i_k = 1}^{n_k} \notin I_1$. Now by induction hypothesis, for all $\delta \leq \xi$, it is the case that $\forall \{ A_i \}_{i = 1}^n \in T_{\Pi^I} \uparrow \delta$, $\{ A_i \}_{i = 1}^n \in I_1$. Therefore, it is necessary that $\forall \{ A_i \}_{i = 1}^n \in g$, $\{ A_i \}_{i = 1}^n \in I_1$. Hence, for $r' \in \Pi$, we have $\{ A_i \}_{i = 1}^n \in T'_\Pi(I_1)$. However, $T'_\Pi(I_1) = I_1$, which implies that $\{ A_i \}_{i = 1}^n \in I_1$, a contradiction.

\item Case 2: $T_{\Pi^I}(g) = \emptyset$. Then, $T_{\Pi^I}(g) = \emptyset$ {\em if and only if} $T'_\Pi(I_1) = \emptyset$, since $I_1 \subset I$.  However, $T'_\Pi(I_1) = I_1$, which implies that $\forall \; \{ A_i \}_{i = 1}^n \in T_{\Pi^I}(g)$, $\{ A_i \}_{i = 1}^n \in I_1$, a contradiction.

\end{enumerate}
By combining these two cases together, it is the case that
\[
\forall \; \{ A_i \}_{i = 1}^n \in T_{\Pi^I}(g), \; \{ A_i \}_{i = 1}^n \in I_1.
\]
which is a contradiction. In particular, it must be the case that
\[
\forall \; \{ A_i \}_{i = 1}^n \in T_{\Pi^I}(g) = T_{\Pi^I} \uparrow \gamma, \{ A_i \}_{i = 1}^n \in T'_\Pi(I_1) = I_1.
\]
A contradiction.
\\

\item Limit ordinal case. Suppose that $\gamma$ is a limit ordinal. Then there exists some non deterministic atom $\{ A_i \}_{i = 1}^n \in {\cal N_L}$ such that
\[
\{ A_i \}_{i = 1}^n \in T_{\Pi^I}\uparrow \gamma,  \{ A_i \}_{i = 1}^n \notin I_1.
\]
By the definition of the upward iterations of the immediate consequence operator of non deterministic logic programs, it is true that
\[
T_{\Pi^I}\uparrow \gamma = \bigcup_{ \delta < \gamma} \; T_{\Pi^I}\uparrow \delta.
\]
But by the induction hypothesis, it is true that for all $\delta < \gamma$,
\[
\forall \; \{ A_i \}_{i = 1}^n \in T_{\Pi^I}\uparrow \delta,  \{ A_i \}_{i = 1}^n \in I_1.
\]
Hence, it must be the case that
\[
\forall \; \{ A_i \}_{i = 1}^n \in T_{\Pi^I}\uparrow \gamma = \bigcup_{ \delta < \gamma} \; T_{\Pi^I}\uparrow \delta, \{ A_i \}_{i = 1}^n \in I_1.
\]
A contradiction. This completes the induction.
\end{itemize}
Since $I$ is a stable non deterministic  model for $\Pi$, then there exists an ordinal $\delta$ such that $I = T_{\Pi^I}\uparrow \delta$. Thus from the induction, it is the case that for all $\forall \; \{ A_i \}_{i = 1}^n \in I, \{ A_i \}_{i = 1}^n \in I_1$, i.e., $I \subseteq I_1$. A contradiction. $\blacksquare$
\end{itemize}
\textbf{Proof of Theorem (\ref{thm:non-deterministic-stable-to-deterministic-stable}).} Let $\Upsilon$ be a ground deterministic normal logic program and $I_{\Upsilon^I}$ be the least deterministic Herbrand model for $\Upsilon^I$, the deterministic reduct of $\Upsilon$ w.r.t. $I$, as defined in \cite{Gelfond_A}. In addition, let $\Pi$ be the ground normal non deterministic logic program representation of $\Upsilon$ and $J_{\Pi^J}$ be the least non deterministic model for $\Pi^I$, the non deterministic reduct of $\Pi$ w.r.t. $J$. Therefore to prove this theorem it is sufficient to show that
\[
\forall \: A  \in I_{\Upsilon^I} \leftrightarrow  \{A\} \in J_{\Pi^J}.
\]
where $(\leftrightarrow)$ means if and only if.
%
Any normal non deterministic logic rule of the form
\[
\{A\} \leftarrow \{B_1\}, \ldots, \{B_l\}, not \; \{B_{l+1}\},\ldots, not \; \{B_m\}
\]
is in $\Pi$ {\em if and only if} a deterministic normal logic rule of the form
\[
A \leftarrow B_1, \ldots, B_l, not \; B_{l+1},\ldots, not \; B_m
\]
is in $\Upsilon$. Therefore, a non deterministic logic rule of the form
\[
\{A\} \leftarrow \{B_1\}, \ldots, \{B_l\}
\]
is in the non deterministic reduct, $\Pi^J$ of $\Pi$ w.r.t. $J$ {\em if and only if} the deterministic definite logic rule
\[
A \leftarrow B_1, \ldots, B_l
\]
is in the deterministic reduct, $\Upsilon^I$ of $\Upsilon$ w.r.t. $I$. Consequently, by Theorem (\ref{thm:equiv2definite})
\[
\forall \: A  \in I_{\Upsilon^I} \leftrightarrow  \{A\} \in J_{\Pi^J}.
\]
$\blacksquare$
\\
\\
\textbf{Proof of Lemma (\ref{lemma:interpretations-equality}).} Follows directly from the definition of partial and total non deterministic interpretations and the definition of the set inclusion $\subseteq$. $\blacksquare$
\\
\\
\textbf{Proof of Lemma (\ref{lemma:lub-and-glb-for-partial}).} Let $lub\{I_1,I_2\}$ exists. First we want to show that $I_1 \cup I_2$ is an upper bound of $\{I_1, I_2\}$. From the definition of partial non deterministic interpretations, we have that $I_1 \subseteq I_1 \cup I_2$ and $I_2 \subseteq I_1 \cup I_2$. Second we show that $I_1 \cup I_2$ is the least upper bound of $\{I_1, I_2\}$. Let $I_u \in {\cal I}_\Pi$ be an upper bound of $\{I_1, I_2\}$. Then, $I_1 \subseteq I_u$ and $I_2 \subseteq I_u$. However, $I_1 \subseteq I_1 \cup I_2$ and $I_2 \subseteq I_1 \cup I_2$. Hence, $I_1 \cup I_2 \subseteq I_u$.

Similarly, let $glb \{ I_1, I_2 \}$ exists. First we want to show that $I_1 \cap I_2$ is a lower bound of $\{I_1, I_2\}$. From the definition of partial non deterministic interpretations, we have that $I_1 \cap I_2 \subseteq I_1$ and $I_1 \cap I_2 \subseteq I_2 $. Second we show that $I_1 \cap I_2$ is the greatest lower bound of $\{I_1, I_2\}$. Let $I_l \in {\cal I}_\Pi$ be a lower bound of $\{I_1, I_2\}$. Then, $I_l \subseteq I_1$ and $I_l \subseteq I_2$. However, $I_1 \cap I_2 \subseteq I_1$ and $I_1 \cap I_2 \subseteq I_2$. Hence, $I_l \subseteq I_1 \cap I_2$. $\blacksquare$
\\
\\
\textbf{Proof of Lemma (\ref{lemma:cpo-for-partial}).} To show that $\langle {\cal I}_\Pi, \subseteq \rangle$ is a complete partial order, we show that every chain in ${\cal I}_\Pi$ has a least upper bound $I_{lub}$. This means that for a chain $X \in {\cal I}_\Pi$, there is $I_{lub} \in {\cal I}_\Pi$ such that $I_{lub} = \cup \{I | I \in X\}$. Clearly, $\langle {\cal I}_\Pi, \subseteq \rangle$ is a partial order. This is because, given $I_1, I_2$, and $I_3$ are in ${\cal I}_\Pi$ we have

\begin{enumerate}

\item $I \subseteq I$ for any $I \in {\cal I}_\Pi$,

\item if $I_1 \subseteq I_2$ and $I_2 \subseteq I_1$, then $I_1 = I_2$, and

\item if $I_1 \subseteq I_2$ and $I_2 \subseteq I_3$, then $I_1 \subseteq I_3$.
\end{enumerate}
Let the chain $X$ contains $I_1, I_2, I_3, \ldots$ such that $I_1 \subseteq I_2 \subseteq I_3 \subseteq \ldots$. Since $I_1 \subseteq I_2 \subseteq I_3 \subseteq \ldots$, this implies that $I_1 \subseteq I_2 \subseteq I_3 \subseteq \ldots \subseteq \cup \{I | I \in X\} = I_{lub}$. Hence, $I_{lub} = \cup \{I | I \in X\}$ is an upper bound of $X$. Clearly, $I_{lub} \in {\cal I}_\Pi$ since $I_{lub} = \cup \{I | I \in X\} \in X$.

Now we show that $I_{lub}$ is the least upper bound of $X$. Let $I_u \in {\cal I}_\Pi$ be an upper bound of $X$. Then $I_1 \subseteq I_2 \subseteq I_3 \subseteq \ldots \subseteq I_u$. However, $I_1 \subseteq I_2 \subseteq I_3 \subseteq \ldots \subseteq \cup \{I | I \in X\}$. Hence $I_{lub} = \cup \{I | I \in X\} \subseteq I_u$. $\blacksquare$
\\
\\
\textbf{Proof of Lemma (\ref{lemma:Wp-Tp-Up-monotonic}).} Let $I_1, I_2$ be two partial non deterministic interpretations for the ground normal non deterministic logic program $\Pi$. Then the proof follows directly from the definitions of $W_\Pi, T_\Pi$, and $U_\Pi$ as follows.

\begin{enumerate}

\item  To prove that $T_\Pi$ is monotonic, we show that if $I_1 \subseteq I_2$, then $T_\Pi(I_1) \subseteq T_\Pi(I_2)$.
By the definition of $T_\Pi$, we have

\[
\begin{array}{lcl}
T_\Pi(I_1)= \{ \qquad \{ A_i \}_{i = 1}^{n} \; | \;  \{ A_i \}_{i = 1}^{n} & \leftarrow &\{ B_{i_1} \}_{i_1 = 1}^{n_1} , \ldots, \{ B_{i_l} \}_{i_l = 1}^{n_l}, \\
&& not \: \{ B_{i_{l+1}} \}_{i_{l+1} = 1}^{n_{l+1}} , \ldots, not \: \{ B_{i_m} \}_{i_m = 1}^{n_m} \; \in \Pi
\end{array}
\]
and $\forall (1 \leq j \leq l)$ $\{ B_{i_j} \}_{i_j = 1}^{n_j} \in I_1$ and $\forall (l+1 \leq k \leq m$) $not \; \{ B_{i_k} \}_{i_k = 1}^{n_k} \in I_1 \; \; \}$.
In addition, we have
\[
\begin{array}{lcl}
T_\Pi(I_2)= \{ \qquad \{ A_i \}_{i = 1}^{n} \; | \;  \{ A_i \}_{i = 1}^{n} & \leftarrow &\{ B_{i_1} \}_{i_1 = 1}^{n_1} , \ldots, \{ B_{i_l} \}_{i_l = 1}^{n_l}, \\
&& not \: \{ B_{i_{l+1}} \}_{i_{l+1} = 1}^{n_{l+1}} , \ldots, not \: \{ B_{i_m} \}_{i_m = 1}^{n_m} \; \in \Pi
\end{array}
\]
and $\forall (1 \leq j \leq l)$ $\{ B_{i_j} \}_{i_j = 1}^{n_j} \in I_2$ and $\forall (l+1 \leq k \leq m$) $not \; \{ B_{i_k} \}_{i_k = 1}^{n_k} \in I_2 \; \; \}$.
\\
\\
However, since $I_1 \subseteq I_2$, we get, for each $\{ A_i \}_{i = 1}^{n}$ belongs to $T_\Pi(I_1)$, we also have $ \{ A_i \}_{i = 1}^{n}$ belongs to $T_\Pi(I_2)$, which implies that $T_\Pi(I_1) \subseteq T_\Pi(I_2)$. This means that $T_\Pi$ is monotonic.
\\

\item Second we prove that $U_\Pi$ is monotonic. From the definition of $U_\Pi$, for any $\{ A_i \}_{i = 1}^{n} \in U_\Pi(I_1)$ implies $\{ A_i \}_{i = 1}^{n}$ must satisfy at least one condition of the Definition (\ref{def:unfounded}). Since $I_1 \subseteq I_2$, then for any $\{ A_i \}_{i = 1}^{n} \in U_\Pi(I_1)$ it is also the case that $\{ A_i \}_{i = 1}^{n} \in U_\Pi(I_2)$. This implies that $U_\Pi(I_1) \subseteq U_\Pi(I_2)$.
\\

\item Finally, we prove that $W_\Pi$ is monotonic. The proof follows directly from the construction of $W_\Pi$ in Definition (\ref{def:W-P}). Since the construction of $W_\Pi$ is based on the operators $T_\Pi$ and $U_\Pi$ and because of both the operators $T_\Pi$ and $U_\Pi$ are monotonic then it immediately follows that $W_\Pi$ is monotonic. $\blacksquare$
\end{enumerate}
\textbf{Proof of Lemma (\ref{lemma:I0-I1-I2-monotonic}).} The proof is by induction on $\alpha$ where $\alpha$ is an ordinal.

\begin{itemize}
\item Base case: When $\alpha = 0$, then $I_0 = \emptyset$ which immediately follows that it is a monotonic sequence of partial non deterministic interpretations.

\item Induction hypothesis: Assume that the lemma holds for all $\beta < \alpha$.

\item Inductive step: let $\alpha = \gamma + 1$ be a successor ordinal. We want to show that $I_\gamma \subseteq I_{\gamma+1}$. Let $\{ A_i \}_{i = 1}^{n} \in I_\gamma$, then there exists a smallest $\beta < \gamma$ such that $\{ A_i \}_{i = 1}^{n} \in W_\Pi(I_\beta)$ (even if $\gamma$ is a limit ordinal). This is true for every $\{ A_i \}_{i = 1}^{n} \in I_\gamma$. But $W_\Pi$ is monotonic, so that by the induction hypothesis $I_\beta \subseteq I_{\beta+1} = W_\Pi(I_\beta)$. We have for every $\{ A_i \}_{i = 1}^{n}  \in I_\gamma$ it is also the case that $\{ A_i \}_{i = 1}^{n} \in W_\Pi(I_\gamma)$. Thus $I_\gamma \subseteq W_\Pi(I_\gamma) = I_{\gamma+1}$. Hence, $I_\gamma \subseteq I_{\gamma+1}$. Monotonicity of the limit ordinal $\alpha$ follows directly from the definition of $I_\alpha$. $\blacksquare$
\end{itemize}
\textbf{Proof of Lemma (\ref{lemma:I0-I1-I2-chain}).} The proof follows directly from Lemma (\ref{lemma:I0-I1-I2-monotonic}), since $I_0, I_1, I_2, \ldots$ is a monotonic sequence of well-founded partial non deterministic interpretations, and hence form a chain in $\langle {\cal I}_\Pi, \subseteq \rangle$. $\blacksquare$
\\
\\
\textbf{Proof of Theorem (\ref{thm:only_fix_Wp})}. The proof follows directly from the definition of the well-founded partial or total non deterministic model of normal non deterministic logic program, $\Pi$, the definition of the operator $W_\Pi$, Lemma (\ref{lemma:I0-I1-I2-monotonic}) and Lemma (\ref{lemma:I0-I1-I2-chain}). $\blacksquare$
\\
\\
\textbf{Proof of Lemma (\ref{lemma:posI1-and-posI2})}. Since $I_1$ and $I_2$ are partial or total non deterministic interpretations and $I_1 \subseteq I_2$, then it follows directly that $pos(I_1) \subseteq pos(I_2)$ and $neg(I_1) \subseteq neg(I_2)$. Hence, $pos(I_1) \subseteq pos(I_2)$. $\blacksquare$
\\
\\
\textbf{Proof of Lemma (\ref{lemma:total111})}. We have $I$ is a total non deterministic model for $\Pi$ and it is also a non deterministic model for $\Pi^I$. On the other hand, $lfp(T_{\Pi^I})$ is the least non deterministic model of $\Pi^I$. Thus, $lfp(T_{\Pi^I}) \subseteq I$ and, hence, it is obvious that $pos(lfp(T_{\Pi^I})) \subseteq pos(I)$. $\blacksquare$
\\
\\
\textbf{Proof of Lemma (\ref{lemma:total2})}. Let $I' = lfp(T_{\Pi^I})$ be the least non deterministic model of $\Pi^I$. First we show that $U_\Pi(I) \subseteq neg(I')$. Since $I'$ is total non deterministic model for $\Pi^I$, it suffices to show that for any $\{A_i\}_{i = 1}^n \in pos(I'), \{A_i\}_{i = 1}^n \notin U_\Pi(I)$. We proceed by induction on the fixpoint iterations $i$ in $T_{\Pi^I} \uparrow i$. We will show that $\forall \; i, \{A_i\}_{i = 1}^n \in pos(T_{\Pi^I}
\uparrow i) \Longrightarrow \{A_i\}_{i = 1}^n \notin U_\Pi(I)$.

\begin{itemize}

\item Base case: $i = 0$. Then $T_{\Pi^I} \uparrow 0 = \emptyset$. Since $pos(T_{\Pi^I} \uparrow 0) =
\emptyset$, then the result is obviously true.

\item inductive hypnosis: Assume that for all $i \leq k$, we have
\[
\{A_i\}_{i = 1}^n \in pos(T_{\Pi^I} \uparrow i) \Longrightarrow \{A_i\}_{i = 1}^n \notin U_\Pi(I).
\]

\item Inductive step:  Let $\{A_i\}_{i = 1}^n \in pos(T_{\Pi^I} \uparrow k+1)$. This means that there is a non deterministic logic rule in $\Pi^I$ of the form
\[
\{ A_i \}_{i = 1}^{n} \leftarrow \{ B_{i_1} \}_{i_1 = 1}^{n_1} , \ldots, \{ B_{i_l} \}_{i_l = 1}^{n_l}
\]
such that for each $\forall (1 \leq j \leq l)$ $\{B_{i_j} \}_{i_j = 1}^{n_j}$ is satisfied by the kth fixpoint
iteration, $T_{\Pi^I} \uparrow k$, of $T_{\Pi^I}$. I.e., $\forall (1 \leq j \leq l)$ $\{B_{i_j} \}_{i_j = 1}^{n_j} \in pos(T_{\Pi^I} \uparrow k)$. This non deterministic logic rule corresponds to the normal non deterministic logic rule, $r \in \Pi$, of the form
\[
\{ A_i \}_{i = 1}^{n}  \leftarrow \{ B_{i_1} \}_{i_1 = 1}^{n_1} , \ldots, \{ B_{i_l} \}_{i_l = 1}^{n_l},
not \: \{ B_{i_{l+1}} \}_{i_{l+1} = 1}^{n_{l+1}} , \ldots, not \: \{ B_{i_m} \}_{i_m = 1}^{n_m}
\]

such that for each $\forall (l+1 \leq k \leq m$) $not \; \{ B_{i_k} \}_{i_k = 1}^{n_k}$ is satisfied by $I$. By Lemma (\ref{lemma:total111}), each $(1 \leq j \leq l)$ $\{ B_{i_j} \}_{i_j = 1}^{n_j} \in pos(I)$. Since $I$ is a total non deterministic model for $\Pi$, the body of $r$ is satisfied by $I$ and its head is also satisfied by $I$. This means that, by the inductive hypothesis, each ($1 \leq k \leq m$), $\{ B_{i_k} \}_{i_k = 1}^{n_k} \notin U_\Pi(I)$. Hence, it is also that $\{ A_i \}_{i = 1}^{n} \notin U_\Pi(I)$ since the body of $r$ is satisfied by $I$, however, $\{ A_i \}_{i = 1}^{n} \in T_\Pi \uparrow k+1$. This implies that $\{A_i\}_{i = 1}^n \in pos(T_{\Pi^I} \uparrow k+1) \Longrightarrow \{A_i\}_{i = 1}^n \notin U_\Pi(I)$. Consequently, this shows that $U_\Pi(I) \subseteq neg(I')$.
\end{itemize}
Second we prove that $neg(I') \subseteq U_\Pi(I)$ by contradiction as follows. Suppose that $\{ A_i \}_{i = 1}^{n} \in neg(I')$ and $neg(I')$ fails to satisfy any of the unfoundedness conditions described in Definition (\ref{def:unfounded}). Then there is a normal non deterministic logic rule, $r \in \Pi$, of the form
\[
\{ A_i \}_{i = 1}^{n}  \leftarrow \{ B_{i_1} \}_{i_1 = 1}^{n_1} , \ldots, \{ B_{i_l} \}_{i_l = 1}^{n_l},
not \: \{ B_{i_{l+1}} \}_{i_{l+1} = 1}^{n_{l+1}} , \ldots, not \: \{ B_{i_m} \}_{i_m = 1}^{n_m}
\]
such that the following facts hold:

\begin{enumerate}

\item Each non deterministic atom, $(1 \leq j \leq l)$, $\{ B_{i_j} \}_{i_j = 1}^{n_j}$, or the negation of a non deterministic atom, $(1 \leq k \leq m)$, $not\; \{ B_{i_k} \}_{i_k = 1}^{n_k}$ in the body of $r$ is satisfied by $I$.

\item No non deterministic atom, $(1 \leq j \leq l)$, $\{ B_{i_j} \}_{i_j = 1}^{n_j}$, in the body of $r$ belongs to $neg(I')$.
\end{enumerate}
Hence,
\[
\{ A_i \}_{i = 1}^{n}  \leftarrow \{ B_{i_1} \}_{i_1 = 1}^{n_1} , \ldots, \{ B_{i_l} \}_{i_l = 1}^{n_l}
\]
is a non deterministic logic rule in $\Pi^I$. Since $I'$ is total non deterministic model for $\Pi$, it follows that that each $(1 \leq j \leq l)$, $\{ B_{i_j} \}_{i_j = 1}^{n_j} \in pos(I')$. Hence, $\{ A_i \}_{i = 1}^{n} \in pos(I')$ and it must be that $\{ A_i \}_{i = 1}^{n} \notin neg(I')$, which leads to a contradiction. $\blacksquare$
\\
\\
\textbf{Proof of Lemma (\ref{lemma:total3}).} Let $I' = lfp(T_{\Pi^I})$ be the least total non deterministic model of $\Pi^I$. By Lemma (\ref{lemma:total111}) $pos(I') \subseteq pos(I)$, hence we have
\[
pos(I')= pos(T_{\Pi^I}(I')) \subseteq pos(T_{\Pi^I}(I))
\]
by monotonicity of $T_{\Pi^I}$. But we have by construction
\[
pos(T_{\Pi^I}(I)) = pos(T_\Pi(I)) \subseteq T_\Pi(I).
\]
Therefore,
\[pos(lfp(T_{\Pi^I})) \subseteq T_\Pi(I).
\]
$\blacksquare$
\\
\\
\textbf{Proof of Theorem (\ref{thm:sp-wp1}).} Let $I' = lfp(T_{\Pi^I})$ be the least total non deterministic model for $\Pi^I$.

\begin{enumerate}

\item Let $I$ be a fixpoint of $W_\Pi$, then we prove that $I$ is a stable non deterministic model for $\Pi$. Since $I$ is a fixpoint of $W_\Pi$, we have $neg(I) = U_\Pi(I)$. But by Lemma (\ref{lemma:total2}) we also have $neg(I') = U_\Pi(I)$. Therefore, $I' = I$.

\item  Let $I$ be a stable non deterministic model for $\Pi$, then we prove that $I$ is a fixpoint of $W_\Pi$. Since $I' = I$, by Lemma (\ref{lemma:total3}), we have $pos(I) = pos(I') \subseteq T_\Pi(I)$. But, $I$ is a total non deterministic model for $\Pi$, where $I$ satisfies each normal non deterministic logic rule in $\Pi$. Moreover, by the construction of $T_\Pi(I)$, we have $T_\Pi(I) \subseteq pos(I)$. Therefore, $T_\Pi(I) = pos(I)$. By Lemma (\ref{lemma:total2}), we have $neg(I) = U_\Pi(I)$ since $I = I'$. Therefore $I$ is a fixpoint of $W_\Pi$. $\blacksquare$
\end{enumerate}
\textbf{Proof of Corollary (\ref{coro:sp-wp1}).} It is easy to show that if $I$ is a stable non deterministic model for $\Pi$ or a fixpoint of $W_\Pi$, then $I$ satisfies $\Pi$. Hence, $I$ is a total non deterministic model for $\Pi$ and Theorem (\ref{thm:sp-wp1}) applies. $\blacksquare$
\\
\\
\textbf{Proof of Corollary (\ref{coro:sp-partial}).} Every stable non deterministic model, $I'$, for $\Pi$ is a fixpoint of $W_\Pi$, by Corollary (\ref{coro:sp-wp1}), and the well-founded partial non deterministic model, $I$, is the only fixpoint of $W_\Pi$ by Theorem (\ref{thm:only_fix_Wp}). Hence the corollary immediately follows. $\blacksquare$
\\
\\
\textbf{Proof of Corollary (\ref{coro:sp-wp2}).} Every stable non deterministic model, $I'$, for $\Pi$ is a fixpoint of $W_\Pi$, by Corollary (\ref{coro:sp-wp1}), and the well-founded total non deterministic model, $I$, is the only fixpoint of $W_\Pi$ by Theorem (\ref{thm:only_fix_Wp}). Hence the corollary immediately follows. $\blacksquare$
\\
\\
\textbf{Proof of Theorem (\ref{thm:wf-model-for-nd-program}).} The proof follows directly from Corollary (\ref{coro:sp-wp2}), Theorem (\ref{thm:sp-non-deterministic-program}), and Theorem (\ref{thm:sp-wp1}), and from the fact that a non deterministic logic program is a normal non deterministic logic program without negated non deterministic atoms.

\begin{enumerate}

\item Let $I$ be the well-founded total non deterministic model of $\Pi$. Then $I$ is the unique stable non deterministic model of $\Pi$ by Corollary (\ref{coro:sp-wp2}). Since $I$ is a unique stable non deterministic model of $\Pi$ then it is least non deterministic model of $\Pi$ by Theorem (\ref{thm:sp-non-deterministic-program}).

\item Let $I$ be the least non deterministic model of $\Pi$. Then $I$ is the unique stable non deterministic model of $\Pi$ by Theorem (\ref{thm:sp-non-deterministic-program}), which in turn is a fixpoint of $W_\Pi$ by Theorem (\ref{thm:sp-wp1}). Since every stable non deterministic model of $\Pi$ is a fixpoint of $W_\Pi$, then for any total non deterministic model $I'$ such that $I' \subseteq I$, $I'$ is not a stable non deterministic model of $\Pi$ and hence is not a fixpoint of $W_\Pi$. Therefore $I$ is the only fixpoint of $W_\Pi$ and hence a well-founded total non deterministic model of $\Pi$. $\blacksquare$
\end{enumerate}
\textbf{Proof of Theorem (\ref{thm:wf-model-for-normal-program}).} Let $\Upsilon$ be a ground deterministic normal logic program and $I$ be the deterministic well-founded partial or total model of $\Upsilon$ as defined in \cite{Gelder}. In addition, let $\Pi$ be the ground normal non deterministic logic program representation of $\Upsilon$ and $J$ be the well-founded partial or total non deterministic model of $\Pi$. Therefore to prove this theorem it is sufficient to show that
\[
\forall \: A \; or \; not \; A \in I \leftrightarrow  \{A\} \; or \; not \; \{A\} \in J.
\]
where $(\leftrightarrow)$ means if and only if. Let $W^d_\Upsilon(I)$, $T^d_\Upsilon(I)$, and $U^d_\Upsilon(I)$ be the well-founded model construction operator, the immediate consequence operator, and the greatest unfounded set operator for the ground deterministic normal logic program, $\Upsilon$, with respect to $I$ respectively, as defined in \cite{Gelder}. Since $I$ is the deterministic well-founded partial or total model of $\Upsilon$, then
\[
I = W^d_\Upsilon(I) =  T^d_\Upsilon(I) \cup not\; U^d_\Upsilon(I)
\]
In addition, since $J$ is the well-founded partial or total non deterministic model of $\Pi$, then
\[
J = W_\Pi(J) =  T_\Pi(J) \cup not \; U_\Pi(J)
\]
It is easy to see that
\[
\forall \: A \in U^d_\Upsilon(I) \leftrightarrow  \{A\}  \in U_\Pi(J).
\]
and
\[
\forall \: A \in T^d_\Upsilon(I) \leftrightarrow  \{A\}  \in T_\Pi(J).
\]
Consequently,
\[
\forall \: X \in T^d_\Upsilon(I)  \cup not\; U^d_\Upsilon(I) \leftrightarrow  \{X \}  \in T_\Pi(J) \cup not \; U_\Pi(J).
\]
This implies that
\[
\forall \: A \; or \; not \; A \in W^d_\Upsilon(I)  \leftrightarrow  \{A\} \; or \; not \; \{A\}  \in W_\Pi(J).
\]
Consequently,
\[
\forall \: A \; or \; not \; A \in I \leftrightarrow  \{A\} \; or \; not \; \{A\} \in J.
\] $\blacksquare$

\end{document}